\documentclass{article} %
\ifdefined\pdfminorversion\pdfminorversion=7\fi %
\usepackage{iclr2027_conference,times}
\usepackage{iclr2027_arxiv}

\usepackage[utf8]{inputenc}
\usepackage[T1]{fontenc}
\usepackage{hyperref}
\usepackage{url}
\usepackage{booktabs}
\usepackage{tabularx}
\usepackage{longtable}
\usepackage{amsfonts,amsmath,amssymb}
\usepackage{graphicx}
\usepackage{xcolor}
\usepackage{colortbl}
\usepackage{multirow}
\usepackage{makecell}
\usepackage{array}
\usepackage{microtype}
\usepackage{wrapfig}
\usepackage{needspace}
\usepackage{capt-of}
\usepackage{placeins}
\usepackage{subcaption}
\DeclareCaptionFont{figurefour}{\fontsize{9.2}{10.5}\selectfont\bfseries}
\usepackage[capitalise,nameinlink]{cleveref}
\crefname{figure}{Fig.}{Figs.}
\Crefname{figure}{Fig.}{Figs.}
\crefname{table}{Tab.}{Tabs.}
\Crefname{table}{Tab.}{Tabs.}
\crefname{section}{Sec.}{Secs.}
\Crefname{section}{Sec.}{Secs.}
\crefname{subsection}{Sec.}{Secs.}
\Crefname{subsection}{Sec.}{Secs.}
\crefname{subsubsection}{Sec.}{Secs.}
\Crefname{subsubsection}{Sec.}{Secs.}
\hypersetup{colorlinks=true,linkcolor=black,citecolor=black,urlcolor=blue}
\newcolumntype{L}[1]{>{\raggedright\arraybackslash}p{#1}}
\graphicspath{{figs/}}
\definecolor{ivgsuccess}{HTML}{F8EDF0}
\definecolor{ivgrecall}{HTML}{EDF5EF}
\definecolor{ivgrubric}{HTML}{EFF3FA}
\definecolor{ivgmuted}{HTML}{5F6C76}
\newcolumntype{K}[2]{>{\columncolor{#1}\centering\arraybackslash}p{#2}}

\newcommand{\ivgnote}[1]{}

\newsavebox{\ivgqatablebox}

\usepackage{amsmath,amsfonts,bm}

\def\eqref#1{equation~\ref{#1}}

\def\1{\bm{1}}

\DeclareMathAlphabet{\mathsfit}{\encodingdefault}{\sfdefault}{m}{sl}
\SetMathAlphabet{\mathsfit}{bold}{\encodingdefault}{\sfdefault}{bx}{n}

\title{Interactive Visual Gym: 
Evaluating Visual Interaction Models with Simulated Users in Interactive Worlds
}

\author{Chenhao Zheng, Zixian Ma, Kai Li, Kanchana Ranasinghe,
  Jieyu Zhang, Artemis Panagopoulou, Weikai Huang, Jianxu Shangguan,
  Yuanyuan Tan, Silvio Savarese, Juan Carlos Niebles, Ranjay Krishna,
  Ran Xu, Honglu Zhou}

\newcommand{\ivgauthor}[2]{%
  \mbox{#1\textsuperscript{#2}}}
\seticlrarxivauthors{%
  \ivgauthor{Chenhao Zheng}{1,2}\quad
  \ivgauthor{Zixian Ma}{1}\quad
  \ivgauthor{Kai Li}{2}\quad
  \ivgauthor{Kanchana Ranasinghe}{2}\\[3pt]
  \ivgauthor{Jieyu Zhang}{1}\quad
  \ivgauthor{Artemis Panagopoulou}{2}\quad
  \ivgauthor{Weikai Huang}{1,2}\quad
  \ivgauthor{Jianxu Shangguan}{1}\\[3pt]
  \ivgauthor{Yuanyuan Tan}{}\quad
  \ivgauthor{Silvio Savarese}{2}\quad
  \ivgauthor{Juan Carlos Niebles}{2}\quad
  \ivgauthor{Ranjay Krishna}{1}\\[3pt]
  \ivgauthor{Ran Xu}{2}\quad
  \ivgauthor{Honglu Zhou}{2}%
}
\seticlrarxivaffiliation{%
  \textsuperscript{1}University of Washington\qquad
  \textsuperscript{2}Salesforce AI Research}
\seticlrarxivresources{%
  \textbf{Project page:} \url{https://interactive-visual-gym.salesforceairesearch.com}\\[2pt]
  \textbf{Code:} \url{https://github.com/SalesforceAIResearch/Interactive-Visual-Gym}}
\hypersetup{%
  pdftitle={Interactive Visual Gym: Evaluating Visual Interaction Models with Simulated Users in Interactive Worlds},
  pdfauthor={Chenhao Zheng, Zixian Ma, Kai Li, Kanchana Ranasinghe, Jieyu Zhang, Artemis Panagopoulou, Weikai Huang, Jianxu Shangguan, Yuanyuan Tan, Silvio Savarese, Juan Carlos Niebles, Ranjay Krishna, Ran Xu, Honglu Zhou}}

\iclrarxivcopy
\begin{document}

\maketitle

\begin{abstract}

Visual interaction models that support users during ongoing physical tasks require evaluation in closed loop, where assistant responses can change subsequent user behavior and task outcomes. Existing benchmarks largely rely on offline replay and therefore cannot capture these effects. We introduce \textbf{Interactive Visual Gym}, a framework for closed-loop evaluation and training of visual interaction models through joint simulation of executable task environments and human users. We contribute three simulators that recreate everyday tasks and human-aligned action spaces across cooking, household manipulation, and mobile-interface domains, together with a persona-conditioned user simulation harness that collaborates with the assistant while incorporating controlled execution errors and other out-of-plan behaviors. We instantiate the framework as \textbf{IVG-Bench}, a benchmark with 315 interaction scenarios. Evaluating state-of-the-art VLMs as visual interaction models reveals substantial limitations: the best-performing model, Claude Opus 5.5, completes only 53.5\% of tasks within the time limit, exhibits significant weaknesses in factual grounding and actionable guidance, and is often too slow to prevent errors when response latency is considered.
 We verify our evaluation protocol with a human study which finds 78.5\% agreement between our interaction-quality metric and participant preferences. 
 We further show that the environments provide useful learning signals: in a proof-of-concept GRPO experiment, closed-loop training improves both task success and interaction quality for Qwen3.5-4B model in our cooking environment.

\end{abstract}

\section{Introduction}
\label{sec:intro}

Building systems that collaborate effectively with people is a longstanding goal of interactive machine learning \citep{amershi2014power}. Recent advances in LLM agents have enabled collaboration on digital tasks, including web research and software development, through dialogue, preference adaptation, and coordinated action \citep{shao2024collaborative,sun2025proactive}. More recently, extending such collaboration to everyday physical tasks has drawn growing interest, motivating \textit{visual interaction models} that observe a video stream of a user's ongoing activities and provide guidance \citep{minicpmo2026,thinkingmachines2026interaction,joyai2026interaction}. For example, an assistant monitoring a procedural physical task like cooking can help plan the next step, detect execution errors, and guide recovery if a mistake happens.

Developing and evaluating interactive agents requires closed-loop rollouts that capture how each interaction changes subsequent user behavior and environment state, enabling assessment of its contribution to task completion. For digital agents, existing websites and software provide readily available tasks and executable environments, which can be paired with simulated users to support such rollouts \citep{webarena2023,bench2024,shao2024collaborative}. Building an equivalent testbed for visual assistance is substantially more difficult. First, physical tasks and their environments are not readily available in a digital form; they must be reconstructed with visual observations, underlying world state, and an action space that supports human task execution. Second, visual assistance places substantially greater demands on user simulation. In most digital settings, the model executes actions while the user primarily clarifies goals and preferences. In physical activities, the user is the acting agent, and the model influences execution through communication. The simulated user must plan, act, make plausible mistakes, while interpreting guidance and deciding how to respond. These simulation challenges have limited visual interaction model training and evaluation to replaying recorded interactions and matching reference responses \citep{guideme2026,plan2026,egopro2026}, where the user's actions and responses remain unchanged regardless of the assistant's guidance.

We present \textbf{Interactive Visual Gym},
a framework for closed-loop evaluation of visual interaction models through joint simulation of task environments and human users. Our framework has three components:

First, we recreate everyday tasks across cooking, household manipulation, and mobile interface use in executable visual environments. These environments render visual observations from the user's perspective. They use scene graphs to represent world state and track task progress independently of variations in action execution. They expose a human-aligned semantic action space for simulated users to operate in, translating instructions such as ``put the pan on the stove'' into executable operations.

Second, we design a harness that turns a state-of-the-art VLM into simulated users that plan and act within these environments while interacting with the assistant and revising their actions in response to its guidance. The harness supports controlled insertion of common human out-of-plan events at specified task states, including repeating, forgetting, or incorrectly executing a step, changing plans, and asking questions. We further model diverse user personas through attributes such as skill level, trust in AI, and interruption preference. These attributes shape how users perform tasks and respond to guidance differently, enabling evaluation of the assistant's adaptation to different users.

Third, simulation provides execution traces that support comprehensive evaluation of an assistant, covering task completion, error detection, and interaction quality. This captures how assistance affects task execution and user experience, which offline replay cannot assess.

\begin{figure}[!t]
\centering
\includegraphics[width=\linewidth]{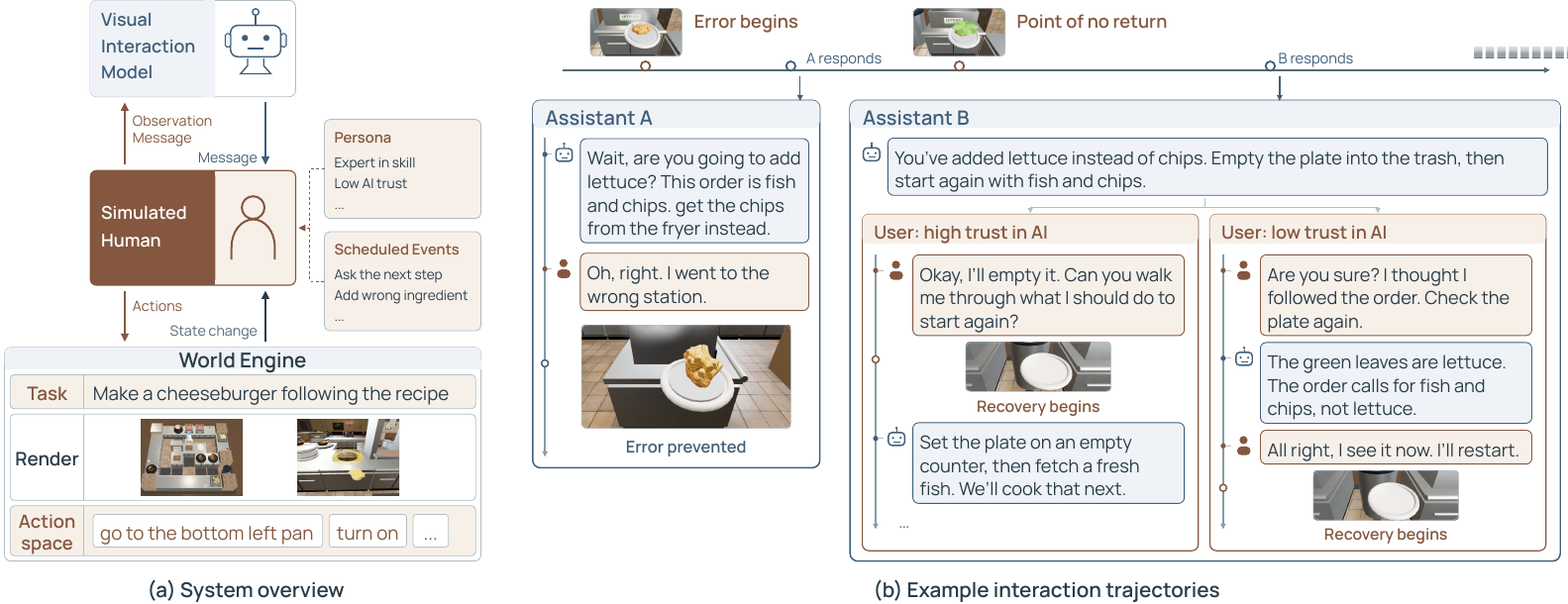}
\caption{\textbf{Interactive Visual Gym.} (a) System overview: The visual interaction model under evaluation is tasked with
  guiding a simulated user to complete a physical task in a visual environment.
(b) Examples showing how different task execution can be simulated based on different assistant response.}
\vspace{-15pt}
\label{fig:teaser}
\end{figure}

As an initial instantiation of our framework, we construct \textbf{IVG-Bench}, a benchmark with 150 cooking, 75 household, and 90 mobile-interface interaction scenarios to evaluate state-of-the-art VLMs as visual interaction models.
In each episode, the assistant is tasked to support a simulated user with a sampled persona through task execution, and needs to deal with a set of scheduled out-of-plan events. 
The dynamic rollout allows us to measure task completion under assistance and assess individual
responses against the underlying world state. 
The best-performing model, Claude Opus 5.5, completes only 53.5\% of tasks within the time limit, averaged across the three environments, and also shows severe deficits in factual grounding and actionable guidance, revealing a substantial gap toward reliable interactive assistance.

Finally, we demonstrate that CookSim can support end-to-end reinforcement learning of multimodal assistants through a proof-of-concept GRPO experiment with Qwen3.5-4B~\citep{qwen2026qwen354b} as the assistant and Gemini 3.7 Flash~\citep{google2026gemini37flash} as the simulated novice user. After 12 RL updates, the assistant's mean in-time task success rate improves by 8.9 percentage points on held-out development-validation scenarios with a Gemini 3.7 Flash user, and by 3.3 percentage points in a separate evaluation with a Gemini 3.8 Flash user (\Cref{tab:rl_cooksim_results}). Despite limited tasks and training updates, these results show that CookSim can supply a learning signal and induce policy improvements in interactive task performance.

\section{Related Work}
\label{sec:related}
\vspace{-5pt}
\noindent\textbf{Streaming visual assistance.}
Recent interaction models continuously process multimodal streams and support low-latency, proactive, or full-duplex responses \citep{minicpmo2026,joyai2026interaction,gander2026,zheng2025trajvit,zheng2026trajtok}.
Live guidance has also been studied through user-in-the-loop cooking assistance~\citep{verghese2024userintheloopevaluationmultimodalllms} and workspace-conditioned visual instructions~\citep{wu2026generativetutoriallivecontextualized, zheng2023exif}.
Streaming motion forecasting extends this focus from understanding observed activity to predicting goal-conditioned object trajectories~\citep{zhang2026molmomotion}.
Complementary datasets and benchmarks study procedural guidance, mistake recovery, correct silence, and response timing in egocentric or general video streams \citep{proactive2025,guideme2026,plan2026,streaming2026,spotbench2026,vsasbench2026,zheng2024iterated}.
In these benchmarks  a candidate model is scored against a prerecorded trajectory.
Ours instead evaluates response-contingent rollouts.

\noindent\textbf{Interactive agent environments.}
Embodied benchmarks place agents in simulated worlds for navigation, manipulation, and collaboration \citep{alfred2020,virtualhome2018,behavior2022,teach2022,habitat2024,partnr2025}, and recent multimodal gyms unify diverse visual tasks for evaluation and reinforcement learning \citep{visgym2026,gymv2026} or study which environment mixtures best train visual agents \citep{envdistributions2026}; in all of these, the evaluated model selects the actions itself.
Digital interaction benchmarks instead pair tool-using agents with simulated users and verify outcomes from application state \citep{bench2024,bench2025,shao2024collaborative}, which $\tau$-Voice extends to voice interaction \citep{ray2026tauvoice}.
BASIS trains and evaluates visual interaction models through simulated user interactions in Alexa Arena~\citep{zhang2025basis}.
User simulation has long enabled scalable dialogue evaluation \citep{schatzmann2005quantitative}, and language models now support persona-conditioned users \citep{aher2023turing}, but simulated users can diverge systematically from people, requesting validation \citep{lostinsimulation2026}.

\begin{figure}[!tb]
\centering
\includegraphics[width=0.93\linewidth]{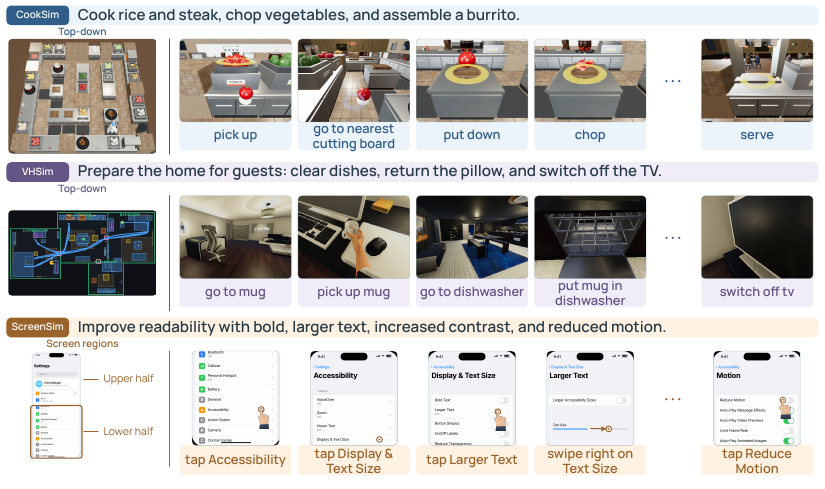}
\vspace{-10pt}
  \caption{\textbf{Example tasks in our environments.} We show the simulated tasks across three domains, with representative
  action commands and corresponding visual observations during task execution. Our semantic action spaces express actions at the level of
  everyday human communication.}
\label{fig:engines}
\vspace{-15pt}
\end{figure}

\section{Method}
\label{sec:method}
\vspace{-5pt}
Our framework combines executable physical task environments with simulated human users to evaluate visual interaction models. We first describe how these environments reconstruct everyday tasks, expose human-communicable actions, and track task progress. We then introduce the human-agent harness and its interaction with the assistant, followed by the evaluation metrics.

\subsection{Executable Physical Task Environments}
\label{sec:engines}

\noindent\textbf{Simulating everyday tasks.}
We reconstruct procedural physical tasks in three domains (\Cref{fig:engines}):

\emph{Cooking.} Inspired by Overcooked~\citep{ghosttown2016overcooked}, we build CookSim from scratch, with recipes specifying ingredient quantities, preparation requirements, and plating order where applicable. Users navigate kitchens with distributed ingredients and tools while accomplishing the cooking task. For example, preparing a cheeseburger requires cooking a patty and plating the bun, patty, and cheese in that order. The simulator is implemented in Three.js/WebGL and supports top-down and egocentric 3D rendering.

\emph{Household activities.} We adapt VirtualHome~\citep{virtualhome2018} for everyday apartment tasks combining navigation and object manipulation. For example, cleaning up after a meal requires loading plates and a mug into the dishwasher, returning an apple to the fridge, and closing both appliances. The simulator uses VirtualHome's Unity backend for egocentric 3D rendering, while providing a top-down trajectory map.

\emph{Mobile interface use.} ScreenSim replicates the iPhone 17 Settings interface, with multi-step configuration tasks derived from Apple's user guides~\citep{apple2026iphoneguide}. For example, restricting an app's access requires disabling its location permission, cross-app tracking, and background refresh while leaving other apps unchanged. The simulator is implemented in HTML/CSS rendered in Chromium and captures mobile-screen observations through Playwright.

\noindent\textbf{Human-communicable action space.}
Existing physical-task simulators expose simulator-specific movement and interaction primitives, such as movement increments or object identifiers~\citep{behavior2022,virtualhome2018,fan2022minedojo}. People typically communicate at a more abstract semantic level, referring to objects and intended operations. We therefore build a semantic action space for each environment (\Cref{fig:engines}), with commands such as ``go to the nearest pan'' in CookSim, ``place the book on the table'' in VHSim, and ``select Background App Refresh'' in ScreenSim. During execution, an LLM-based grounding module resolves contextual references, such as ``the pan on your left,'' into object identifiers recognized by the simulator. The simulator then expands each high-level action into a sequence of low-level controls that update the world state and generate visual observations.

\noindent\textbf{Progress tracking via scene-graph state.}
Each simulator represents world state as a scene graph with nodes encoding objects and attributes, such as a dishwasher's open or closed state, and edges encoding containment and support relations. We decompose each task into sub-goals defined by conditions over partial graph state, such as the designated dishes being inside the dishwasher. Tracking these conditions measures task progress independently of different valid execution procedures.

\subsection{User--Assistant Interaction}
\label{sec:harness}
The visual interaction model under evaluation is tasked with guiding a simulated user through tasks in the environments described above. We implement the
  user as a VLM agent with access to structured world-state information, execution feedback, and conversation history. The VLM is prompted to role-play a human user to test the assistant's interaction capabilities, planning and acting independently while responding to guidance according
  to its persona. A user-simulation harness conditions its behavior through persona instructions and temporarily takes control of its actions
  to enact scheduled out-of-plan events. Below, we describe the action and communication interfaces, account for assistant response delays, and
  detail these mechanisms for controlling user behavior.

\noindent\textbf{Action interface: user--environment.}
The simulated user maintains an action queue that the engine executes at each simulation tick and can revise in response to environment changes or assistant suggestions. The engine updates the world state and renders the resulting observation:
\begin{equation}
s_{t+1}=T(s_t,a_t), \qquad o_{t+1}=R(s_{t+1}),
\label{eq:world_transition}
\end{equation}
where $a_t$ is the action executed from the user's queue, $T$ is the state-transition, and $R$ is the renderer.

\noindent\textbf{Communication interface: user--assistant.}
The assistant monitors sampled frames from the user's observation stream alongside the conversation history. At an assistant invocation at tick $t$, it generates a response
\begin{equation}
m_t=A(o_{\leq t},c_{\leq t}),
\label{eq:assistant_response}
\end{equation}
where $o_{\leq t}$ denotes the sampled visual history and $c_{\leq t}$ is the conversation history. The response $m_t$ contains an utterance or silence in text format and an optional event-type flag (\Cref{tab:user_simulation}(A)), which is enabled when the assistant detects a user execution error. On each user turn, the simulated user likewise produces an utterance or remains silent. Its conversational response types include asking for an explanation, questioning the correctness of guidance, requesting more guidance or fewer interruptions, accepting or rejecting a suggestion, and answering a question.

 \noindent\textbf{Accounting for assistant delay.}
  Assistant processing delay can substantially affect interaction quality: by the time a response arrives, the user and environment may have
  progressed beyond the observed state, rendering the guidance outdated or invalid. Following the separation of simulation time and wall-clock
  time in $\tau$-Voice~\citep{ray2026tauvoice}, we assign each simulation tick a duration $\Delta$ based on approximate real-world action
  speeds. For example, VirtualHome is configured to move the user one meter per simulation tick. Assuming this distance takes 1.2 seconds to
  walk in the real world, we assign each tick a duration of $\Delta=1.2$ seconds. 
  We convert the measured assistant inference latency $\ell_t$ into a delay of $d_t=\lceil \ell_t/\Delta\rceil$ simulation ticks. The response
  $m_t$, generated from observations through tick $t$, is delivered at tick $t+d_t$.
During this interval, the engine continues
  executing the user's queued actions and updating the world state.

\noindent\textbf{Scheduled out-of-plan events.}
We introduce scheduled out-of-plan events in all three simulators that require the assistant to respond to execution errors, questions and intended plan change. For example, the event types in CookSim are shown in \Cref{tab:user_simulation}(A): we simulate common cooking errors of incorrect preparation, omitted or improperly sequenced steps, and ill-timed actions~\citep{captaincook4d2024}; we generate temporal questions adapted from streaming-video benchmarks~\citep{Niu_2025_CVPR,Huang_2025_CVPR,streamingbench2024} and proactive requests for timely assistance~\citep{proactive2025}; we also introduce intended plan changes, where users may announce revised goals or a side task during execution.
Each event is triggered by a specified
  condition on task progress.

When an event is triggered, the harness temporarily takes control of the user's action queue to enforce its execution.
It privately informs the user agent of the ongoing event and explains that the event is intended to test the assistant, instructing the user agent to adapt its conversation accordingly. For mistake events, the user agent judges whether the assistant has correctly identified the mistake and decides whether to accept its guidance. Once the guidance is accepted, the harness returns control of the action queue to the user.

\noindent\textbf{Persona-conditioned users.}
Real-world users differ in how they perform tasks and respond to assistance. Drawing on HCI and related behavioral research, we operationalize this diversity through four representative axes: skill level~\citep{shneiderman2000universal}, trust in AI~\citep{lee2004trust}, execution speed~\citep{wickelgren1977speed},  and interruption preference~\citep{mcfarlane2002interruption} (\Cref{tab:user_simulation}(B)). The harness instantiates each persona through axis-specific instructions in the user model's context. For example, high-skill users are prompted to plan ahead independently, while low-skill users are prompted to ask assistant frequently to determine the next steps. Users who prefer minimal interruption are prompted to find unnecessary commentary disruptive, whereas those who benefit from step-by-step support prefer more frequent guidance. The execution speed and interruption preference are shared to assistant as part of announced user preference, while the skill level and trust in AI are treated as internal property of a user for assistants to infer and adapt.

\begin{table}[!t]
\scalebox{0.97}{%
\centering\scriptsize
\begin{tabularx}{\linewidth}{@{}X@{\hspace{0.025\linewidth}}X@{}}
\toprule
\begin{minipage}[t]{\linewidth}
\centering

\setlength{\tabcolsep}{3pt}
\renewcommand{\arraystretch}{1.14}
\begin{tabularx}{\linewidth}[t]{@{}l >{\raggedright\arraybackslash}X@{}}
\multicolumn{2}{@{}l@{}}{\textbf{A\quad Scheduled out-of-plan events} \normalfont(CookSim)}\\
\midrule
\textbf{Category} & \textbf{Event types}\\
\midrule
Execution errors & Repeat, forget, or skip a step; wrong plating order; wrong ingredient; add unwanted ingredients; plate raw food; unattended cooking.\\
\addlinespace[4pt]
Questions & Action order; counts; elapsed time; time duration; event recall\\
\addlinespace[4pt]
Proactive requests & Notify when a specified condition holds.\\
\addlinespace[4pt]
Plan changes & Modify the recipe; perform a side task.\\
\end{tabularx}

\end{minipage} &
\begin{minipage}[t]{\linewidth}
\centering
\setlength{\tabcolsep}{3pt}
\renewcommand{\arraystretch}{1.14}
\begin{tabularx}{\linewidth}[t]{@{}l >{\raggedright\arraybackslash}X@{}}
\multicolumn{2}{@{}l@{}}{\textbf{B\quad User personas}}\\
\midrule
\textbf{Axis} & \textbf{Levels / behavior}\\
\midrule
Skill & Low / Medium / High\newline \textit{Planning and anticipation.}\\
\addlinespace[4pt]
Trust in AI & Low / Medium / High\newline \textit{Evidence needed to accept advice.}\\
\addlinespace[4pt]
Execution speed & Cautious / Balanced / Quick\newline \textit{Careful versus rapid execution.}\\
\addlinespace[4pt]
Interruption & Step-by-step / Moderate / Necessary only\newline \textit{Receptiveness to frequent guidance.}\\
\end{tabularx}

\end{minipage}\\
\bottomrule
\end{tabularx}

}
\caption{\textbf{Controllable user simulation.} (A) Scheduled out-of-plan events, illustrated with CookSim, test assistant responses to different types of out-of-plan events. Events are triggered by conditions on task progress. (B) Four persona
 axes, each with three levels, simulate how different simulated users perform tasks and respond to guidance differently .}
    \vspace{-20pt}
\label{tab:user_simulation}

\end{table}

\subsection{Evaluation Metrics}
\label{sec:evaluation_metrics}

Our simulation rollouts allow us to measure task completion under assistance and assess individual responses against the underlying world state. We evaluate the assistants with the following metrics:

\noindent\textbf{In-time success.}
\emph{In-time success} measures the proportion of episodes whose engine-verified goal is satisfied within a completion budget of
  $2(T_i^{\mathrm{ref}}+\alpha_i)$, where $T_i^{\mathrm{ref}}$ is the duration of an engine-generated, error-free reference rollout and $
  \alpha_i$ is the prescribed event allowance. Environment-specific budgets are detailed in \Cref{app:evaluation_metrics}.

\noindent\textbf{Error detection.}
\emph{Detection recall} is the proportion of triggered errors correctly identified within their eligible detection windows, credited when
  the assistant raises a correct event flag verified by the engine or when the simulated user judges its diagnosis correct. For the latency
  ablation, we also report \emph{error prevention}, the proportion of triggered errors prevented before the point of no return. Prevention
  requires both timely identification and guidance that persuades the user to change course.

\noindent\textbf{Interaction quality.}
    We evaluate each assistant response against the underlying world state, rather than against a reference response as in prior works. We
  define five dimensions of evaluation informed by dialogue evaluation~\citep{mehri-eskenazi-2020-unsupervised} and HCI guidance on context-sensitive, user-controllable assistance~\citep{amershi2019guidelines}: factual grounding, situational relevance,
  actionable guidance, user intent uptake, and guidance
    conciseness. We operationalize these dimensions as binary rubric items evaluated by two LLM judges~\citep{liu-etal-2023-g,checkeval2025}.
  For
    each conversational round, each judge receives engine state, user preferences, conversation history, and subsequent user actions,
  returning evidence-supported pass, fail, or not-applicable judgments of each rubrics. A category passes only when all applicable items
    pass; each judge's overall score averages applicable categories, gated by its factual-grounding judgment. We average the two judges' round scores before aggregating across rounds and runs. Judge prompts and score aggregation are detailed in
    \Cref{app:rubric}.

\noindent\textbf{Question answering and proactive requests.}
The engine pairs each generated question with a grading procedure that derives reference facts from world state and execution history. A semantic judge assigns correctness against these facts. For proactive requests, the engine derives an acceptable notification window from the relevant state transitions.

\section{Experiments}
\label{sec:experiments}

We use our engine to instantiate \textbf{IVG-Bench} benchmark for evaluating state-of-the-art VLMs as visual interaction models (\Cref{sec:benchmark}), and validate our evaluation paradigm through a human study (\Cref{sec:human_study}). We further present a preliminary reinforcement learning experiment showing that our engine can support training visual interaction models and improve their interaction performance (\Cref{sec:rl}).

\subsection{IVG-Bench: Benchmarking Visual Interaction Models in Interactive Visual Gym}
\label{sec:benchmark}

IVG-Bench comprises 315 interaction scenarios: 150 in CookSim, 75 in VHSim, and 90 in ScreenSim. Each case specifies a task, a user persona, and scheduled out-of-plan events triggered by task-progress conditions. Below we detail the setting for evaluated visual interaction models and show evaluation results.

\subsubsection{Visual Interaction Model Setting}
\label{sec:assistance_setting}

\noindent\textbf{Models.}
We evaluate four classes of assistants: (1) frontier API models: GPT-6 Astra~\citep{openai2026astra}, GPT-6.1 Sol~\citep{openai2026sol}, Gemini 3.8 Flash~\citep{google2026gemini38flash}, Gemini 4 Argon~\citep{google2026gemini4argon}, and Claude Opus 5.5~\citep{anthropic2026opus55}; (2) real-time streaming API models: Gemini 3.8 Live~\citep{google2026gemini38live} and GPT-Realtime 2.1~\citep{openai2026realtime21}; (3) open-weight non-streaming VLMs: Qwen3.8-27B~\citep{qwen2026qwen38}, Molmo2-8B~\citep{clark2026molmo2}, InternVideo3-8B~\citep{yan2026internvideo3}, Kimi-VL-A3B-Thinking~\citep{kimiteam2025kimivl}, Muse Glimmer~\citep{meta2026museglimmer}, and Inkling~\citep{thinkingmachines2026inkling}; and (4) open-weight streaming VLMs: MiniCPM-o 4.5~\citep{minicpmo2026}, Qwen3-Omni-30B~\citep{xu2025qwen3omni}, Proact-VL~\citep{yan2026proactvl}, and StreamingVLM~\citep{xu2025streamingvlm}. We use Gemini 3.7 Flash as the simulated user in CookSim and ScreenSim, and Gemini 3.1 Pro in VHSim. We examine sensitivity to the user model in \Cref{app:user_model_ablation}.

\noindent\textbf{Inference.}
Assistants observe the user's ongoing activity through visual observations and conversation, without access to the scheduled events or the user's private action plan. In CookSim and VHSim, we query assistants every two simulation ticks, including real-time API models in a polled setting. Streaming models retain context within their sessions. We adapt non-streaming models with a sliding window of the eight most recent observations and a recurrent, text-based Plan--Act memory. This follows the use of persistent procedural plans for visual assistance in \citet{plan2026}. Each locally hosted open-weight model is served on an H100 GPU in CookSim and VHSim. ScreenSim settings are detailed in \Cref{app:assistance}.

\noindent\textbf{Response timing.}
We use fixed response timing for the main comparison: two simulation ticks for CookSim and VHSim's streaming interactions, and zero ticks for ScreenSim's action--response interactions. We separately evaluate the effect of measured assistant latency in CookSim and VHSim. Each setting uses three independent rollouts per benchmark scenario.

\subsubsection{Quantitative Results}
\label{sec:results}
\begin{wrapfigure}{r}{0.43\textwidth}
\vspace{-8pt}
\centering
\includegraphics[width=\linewidth]{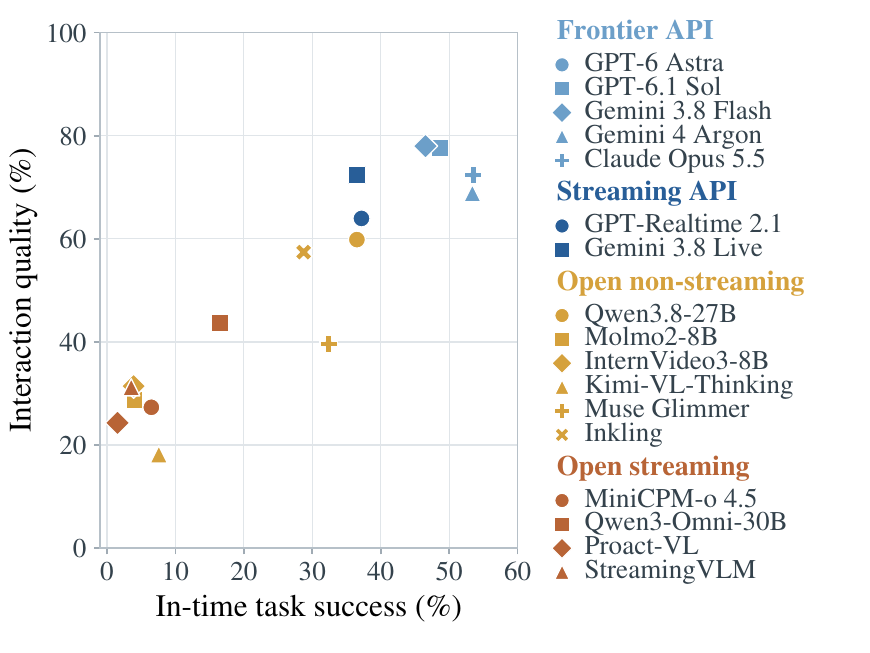}
\vspace{-15pt}
\caption{\textbf{Task success vs. interaction quality.} Scores are averaged over environments.}
\label{fig:success_quality}
\vspace{-10pt}
\end{wrapfigure}

We report task success, error detection, and interaction quality across the three environments in \Cref{tab:assistance} and \Cref{fig:success_quality}. \Cref{tab:qa_proactive,tab:assistant_latency} report question answering and the effect of assistant latency, respectively. We discuss the main findings below.

\begin{table}[!t]
\centering
\begingroup
\setlength{\tabcolsep}{0pt}
\renewcommand{\arraystretch}{1.06}
\rmfamily\fontsize{6.0}{8.7}\selectfont
\begin{tabular}{@{}L{.128\linewidth}K{ivgsuccess}{0.050\linewidth}K{ivgrecall}{0.035\linewidth}K{white}{0.032\linewidth}K{white}{0.032\linewidth}K{white}{0.032\linewidth}K{white}{0.035\linewidth}K{white}{0.032\linewidth}K{ivgrubric}{0.038\linewidth}@{\hspace{2pt}}K{ivgsuccess}{0.050\linewidth}K{ivgrecall}{0.035\linewidth}K{white}{0.032\linewidth}K{white}{0.032\linewidth}K{white}{0.032\linewidth}K{white}{0.035\linewidth}K{white}{0.032\linewidth}K{ivgrubric}{0.038\linewidth}@{\hspace{2pt}}K{ivgsuccess}{0.050\linewidth}K{ivgrecall}{0.035\linewidth}K{white}{0.032\linewidth}K{white}{0.032\linewidth}K{white}{0.032\linewidth}K{white}{0.035\linewidth}K{white}{0.032\linewidth}K{ivgrubric}{0.038\linewidth}@{}}
\toprule
 & \multicolumn{8}{c}{\bfseries CookSim} & \multicolumn{8}{c}{\bfseries VHSim} & \multicolumn{8}{c}{\bfseries ScreenSim}\\[2pt]
\cmidrule(lr){2-9}\cmidrule(lr){10-17}\cmidrule(lr){18-25}
 & \multicolumn{1}{c}{\multirow{2}{*}{\shortstack{In-time\\success}}} & \multicolumn{1}{c}{\multirow{2}{*}{\shortstack{Det.\\recall}}} & \multicolumn{6}{c}{\bfseries Interaction quality} & \multicolumn{1}{c}{\multirow{2}{*}{\shortstack{In-time\\success}}} & \multicolumn{1}{c}{\multirow{2}{*}{\shortstack{Det.\\recall}}} & \multicolumn{6}{c}{\bfseries Interaction quality} & \multicolumn{1}{c}{\multirow{2}{*}{\shortstack{In-time\\success}}} & \multicolumn{1}{c}{\multirow{2}{*}{\shortstack{Det.\\recall}}} & \multicolumn{6}{c}{\bfseries Interaction quality}\\
\cmidrule(lr){4-9}\cmidrule(lr){12-17}\cmidrule(lr){20-25}
\textbf{Assistant} &  & & {\fontsize{5.5}{7.2}\selectfont\shortstack{Fac.\\grnd.}} & {\fontsize{5.5}{7.2}\selectfont\shortstack{Sit.\\rel.}} & {\fontsize{5.5}{7.2}\selectfont\shortstack{Act.\\guid.}} & {\fontsize{5.5}{7.2}\selectfont\shortstack{User\\intent}} & {\fontsize{5.5}{7.2}\selectfont\shortstack{Guid.\\conc.}} & {\fontsize{5.5}{7.2}\selectfont\shortstack{Overall}} &  & & {\fontsize{5.5}{7.2}\selectfont\shortstack{Fac.\\grnd.}} & {\fontsize{5.5}{7.2}\selectfont\shortstack{Sit.\\rel.}} & {\fontsize{5.5}{7.2}\selectfont\shortstack{Act.\\guid.}} & {\fontsize{5.5}{7.2}\selectfont\shortstack{User\\intent}} & {\fontsize{5.5}{7.2}\selectfont\shortstack{Guid.\\conc.}} & {\fontsize{5.5}{7.2}\selectfont\shortstack{Overall}} &  & & {\fontsize{5.5}{7.2}\selectfont\shortstack{Fac.\\grnd.}} & {\fontsize{5.5}{7.2}\selectfont\shortstack{Sit.\\rel.}} & {\fontsize{5.5}{7.2}\selectfont\shortstack{Act.\\guid.}} & {\fontsize{5.5}{7.2}\selectfont\shortstack{User\\intent}} & {\fontsize{5.5}{7.2}\selectfont\shortstack{Guid.\\conc.}} & {\fontsize{5.5}{7.2}\selectfont\shortstack{Overall}}\\[3pt]
\midrule\rowcolor{white}\multicolumn{25}{@{}l@{}}{\bfseries Frontier API models}\\[2pt]
Gemini 4 Argon & \textbf{60.9} & 80.2 & 70.2 & 90.1 & 71.4 & 82.2 & 76.9 & 63.8 & \textbf{15.6} & \textbf{81.2} & 60.3 & 90.7 & 60.8 & 83.2 & 84.8 & 55.6 & 83.7 & \textbf{100.0} & 91.0 & 95.5 & 86.5 & 87.1 & 95.1 & 87.1\\
Claude Opus 5.5 & \underline{57.1} & \textbf{86.0} & 69.3 & \underline{92.4} & 75.3 & 83.8 & 90.7 & 66.0 & 14.2 & 62.6 & 71.8 & \textbf{93.8} & \textbf{65.1} & 88.4 & 91.4 & 67.0 & \underline{89.3} & \textbf{100.0} & 87.1 & 96.5 & 90.0 & 85.9 & 95.3 & 84.0\\
Gemini 3.8 Flash & 46.4 & 77.9 & \textbf{83.5} & 83.5 & 65.5 & 66.2 & \underline{90.9} & 73.4 & \underline{14.7} & 72.2 & \underline{79.2} & 88.4 & 63.4 & 75.3 & \textbf{92.7} & \textbf{71.2} & 78.5 & 97.8 & \underline{92.4} & 95.7 & 89.4 & 80.2 & \textbf{99.0} & \textbf{89.4}\\
GPT-6.1 Sol & 42.9 & 79.9 & 80.4 & \underline{92.4} & \underline{78.1} & \underline{91.1} & \textbf{91.7} & \underline{75.8} & \textbf{15.6} & 60.3 & 77.8 & \underline{92.9} & 57.2 & \underline{89.1} & \underline{92.2} & 70.5 & 87.4 & \textbf{100.0} & 89.0 & \underline{96.6} & \textbf{91.7} & \textbf{93.4} & 97.3 & 86.5\\
GPT-6 Astra & 40.2 & \underline{80.4} & \underline{80.5} & \textbf{92.9} & \textbf{78.9} & \textbf{93.6} & 90.5 & \textbf{76.0} & 10.7 & 62.1 & 77.7 & 92.8 & 59.5 & \textbf{90.7} & 91.4 & \underline{71.1} & \textbf{90.4} & \textbf{100.0} & 89.0 & \textbf{96.8} & \underline{91.0} & \underline{91.5} & 96.5 & 86.3\\
\midrule\rowcolor{white}\multicolumn{25}{@{}l@{}}{\bfseries Real-time streaming API models}\\[2pt]
GPT-Realtime 2.1 & 33.3 & 68.0 & 65.9 & 69.4 & 46.2 & 44.8 & 37.3 & 45.1 & 11.6 & 60.8 & \underline{79.2} & 86.9 & \underline{64.3} & 67.4 & 56.5 & 62.3 & 66.7 & \underline{98.6} & 90.5 & 96.2 & 85.3 & 88.4 & 86.3 & 84.5\\
Gemini 3.8 Live & 26.2 & 58.5 & 76.8 & 71.4 & 44.6 & 55.1 & 81.8 & 60.6 & 11.6 & 24.6 & \textbf{83.5} & 81.5 & 49.8 & 58.5 & 83.8 & 67.4 & 71.9 & 96.1 & \textbf{92.5} & 95.1 & 86.8 & 72.2 & \underline{98.2} & \underline{89.1}\\
\midrule\rowcolor{white}\multicolumn{25}{@{}l@{}}{\bfseries Open-weight non-streaming VLMs}\\[2pt]
Muse Glimmer & 40.7 & 60.5 & 28.9 & 68.6 & 33.7 & 36.5 & 27.6 & 21.6 & 12.4 & 14.3 & 53.8 & 83.0 & 50.0 & 61.7 & 50.4 & 43.6 & 44.1 & 96.1 & 61.0 & 83.2 & 61.7 & 55.0 & 66.4 & 53.8\\
Inkling & 30.4 & 79.5 & 52.7 & 72.5 & 39.2 & 44.6 & 68.2 & 41.3 & 8.0 & \underline{79.0} & 62.8 & 86.2 & 48.4 & 63.7 & 83.5 & 53.9 & 47.8 & 98.4 & 83.9 & 89.6 & 73.7 & 51.5 & 92.0 & 77.0\\
Qwen3.8-27B & 29.8 & 60.6 & 51.9 & 83.8 & 53.7 & 70.1 & 68.1 & 44.6 & 12.0 & 31.8 & 54.3 & 88.0 & 56.3 & 79.8 & 74.9 & 48.7 & 67.8 & 96.2 & 88.7 & 95.7 & 89.0 & 72.7 & 97.0 & 86.2\\
InternVideo3-8B & 7.3 & 25.6 & 46.5 & 36.3 & 16.7 & 17.1 & 48.3 & 27.7 & 4.0 & 0.0 & 48.7 & 54.7 & 31.2 & 26.0 & 45.1 & 32.6 & 0.4 & 41.1 & 51.0 & 41.9 & 22.5 & 14.7 & 62.6 & 33.9\\
Kimi-VL-Thinking & 7.3 & 56.6 & 7.3 & 27.4 & 5.5 & 14.5 & 11.2 & 4.3 & 5.8 & 74.0 & 26.7 & 50.6 & 17.4 & 25.2 & 31.1 & 17.2 & 9.6 & 80.0 & 50.2 & 48.7 & 25.4 & 20.9 & 50.4 & 33.0\\
Molmo2-8B & 5.6 & 5.1 & 41.9 & 18.5 & 10.0 & 14.6 & 12.3 & 16.9 & 5.8 & 1.1 & 54.4 & 41.3 & 23.0 & 13.0 & 32.1 & 30.3 & 0.7 & 21.5 & 74.0 & 27.1 & 17.6 & 12.3 & 59.7 & 39.2\\
\midrule\rowcolor{white}\multicolumn{25}{@{}l@{}}{\bfseries Open-weight streaming VLMs}\\[2pt]
Qwen3-Omni-30B & 23.6 & 39.0 & 40.7 & 52.9 & 28.8 & 24.7 & 47.4 & 28.3 & 10.7 & 2.7 & 56.9 & 73.4 & 41.8 & 39.8 & 67.3 & 44.4 & 15.2 & 58.3 & 71.1 & 68.8 & 51.7 & 28.5 & 80.0 & 58.2\\
MiniCPM-o 4.5 & 11.1 & 39.6 & 29.0 & 31.9 & 9.8 & 22.1 & 34.5 & 15.8 & 5.8 & 9.8 & 45.3 & 53.8 & 18.7 & 34.2 & 35.3 & 27.7 & 2.6 & 45.2 & 56.8 & 52.2 & 27.6 & 43.2 & 59.3 & 38.5\\
StreamingVLM & 5.3 & 0.7 & 73.9 & 6.6 & 16.8 & 30.7 & 14.9 & 27.2 & 5.3 & 0.0 & 68.4 & 14.0 & 7.7 & 22.8 & 15.8 & 29.3 & 0.0 & 0.9 & 89.8 & 9.6 & 12.7 & 16.7 & 22.7 & 37.2\\
Proact-VL & 2.4 & 0.8 & 43.5 & 10.5 & 5.1 & 13.8 & 23.7 & 17.8 & 2.2 & 0.0 & 58.5 & 23.1 & 7.0 & 16.3 & 30.5 & 27.1 & 0.0 & 1.9 & 72.8 & 6.5 & 3.0 & 3.4 & 31.6 & 27.9\\
\bottomrule
\end{tabular}\par
\endgroup

\caption{\textbf{Quantitative results across three environments.} Results are averaged across three independent runs. Det.\ recall: error detection recall; Fac.\ grnd.: factual grounding; Sit.\ rel.: situational relevance; Act.\ guid.: actionable guidance; User intent: user intent uptake; Guid.\ conc.: guidance conciseness. Scores are percentages; bold and underline indicate the best and second-best scores in each column.}
\label{tab:assistance}
\vspace{-14pt}
\end{table}

\noindent\textbf{Even the strongest VLMs fall short of reliable visual assistance.}
Claude Opus 5.5 achieves the highest task success, but completes only 53.5\% of tasks within the time limit, averaged across the three environments. Moreover, Claude Opus 5.5 trails Astra, Sol, and Gemini Flash in interaction quality, with notable limitations in factual grounding and actionable guidance. \textit{Open-weight assistants generally fall short of proprietary models on both task success and interaction quality.}

\noindent\textbf{The strongest streaming models remain competitive in interaction quality but lag in task success.}
Gemini Live achieves high average interaction quality, yet completes fewer tasks than Astra and Gemini Flash. This pattern may reflect a trade-off between low-latency inference and high planning and reasoning ability.

\noindent\textbf{Higher task success does not necessarily imply better interaction quality.}
Although the two measures are positively associated (\Cref{fig:success_quality}), task outcomes alone can hide weak assistance. For example, Muse Glimmer completes more tasks than Astra in CookSim while receiving much lower interaction-quality scores. Completing the task is therefore not sufficient evidence of high quality guidance.

\noindent\textbf{Temporal question answering is surprisingly difficult during task-directed interaction.}
We present the result of temporal question answering from the measured-delay CookSim runs in \Cref{tab:qa_proactive}.
Unlike questions about fixed videos, the questions are asked by human agent at specific moments during multi-turn interactions in which the assistant must also help the user complete the task, closely reflecting real world scenario.
We find that open-weight models often lose track of the question or request during multi-turn conversations (\Cref{tab:qa_proactive}).
Proprietary models generally perform better, with Gemini 4 Argon and Astra leading the pooled evaluation but still exhibit weakness in temporal action counting.

\begin{table}[!t]
\centering
\begingroup\small
\setlength{\tabcolsep}{3pt}\renewcommand{\arraystretch}{1.08}
\begin{tabular*}{\linewidth}{@{\extracolsep{\fill}}lrrrrrr@{}}
\toprule
 & \multicolumn{4}{c}{Question answering} & \multirow{2}{*}{\shortstack{Proactive\\request}} & \multirow{2}{*}{Overall}\\
\cmidrule(lr){2-5}
Assistant & \shortstack{Temp.\\order} & \shortstack{Act.\\count} & \shortstack{Time\\dur.} & \shortstack{Event\\recall} & & \\
\midrule
\multicolumn{7}{@{}l}{\itshape Frontier API models}\\
Gemini 4 Argon & 66.8 & 52.8 & 82.3 & 85.8 & 85.3 & 79.0\\
GPT-6 Astra & 51.1 & 34.8 & 89.5 & 88.9 & 86.8 & 78.6\\
GPT-6.1 Sol & 58.9 & 22.9 & 81.7 & 76.5 & 88.2 & 74.3\\
Claude Opus 5.5 & 62.5 & 28.3 & 75.8 & 69.1 & 62.7 & 63.3\\
Gemini 3.8 Flash & 67.3 & 30.0 & 59.1 & 55.8 & 69.7 & 59.4\\
\addlinespace[2pt]
\multicolumn{7}{@{}l}{\itshape Real-time streaming API models}\\
Gemini 3.8 Live & 46.9 & 23.5 & 51.5 & 73.5 & 43.4 & 47.6\\
GPT-Realtime 2.1 & 39.1 & 13.3 & 58.2 & 62.5 & 43.7 & 46.5\\
\addlinespace[2pt]
\multicolumn{7}{@{}l}{\itshape Open-weight non-streaming VLMs}\\
Qwen3.8-27B & 47.1 & 18.2 & 49.8 & 64.9 & 25.8 & 38.7\\
Muse Glimmer & 9.9 & 1.8 & 38.8 & 31.2 & 28.5 & 27.5\\
Inkling & 9.2 & 9.2 & 36.4 & 8.8 & 36.4 & 27.4\\
Kimi-VL-Thinking & 3.0 & 0.0 & 29.7 & 0.0 & 26.7 & 19.2\\
Molmo2-8B & 9.7 & 1.9 & 15.3 & 38.1 & 14.3 & 15.8\\
InternVideo3-8B & 0.0 & 0.0 & 17.4 & 8.3 & 21.8 & 14.3\\
\addlinespace[2pt]
\multicolumn{7}{@{}l}{\itshape Open-weight streaming VLMs}\\
Qwen3-Omni-30B & 7.5 & 3.2 & 30.1 & 10.2 & 30.9 & 22.9\\
MiniCPM-o 4.5 & 5.6 & 0.0 & 9.3 & 19.2 & 9.6 & 9.6\\
StreamingVLM & 2.2 & 1.8 & 2.3 & 2.8 & 5.3 & 3.4\\
Proact-VL & 0.0 & 1.9 & 0.8 & 0.0 & 2.3 & 1.3\\
\addlinespace[2pt]
\bottomrule\end{tabular*}\endgroup

\caption{\textbf{Question answering and proactive requests in CookSim.} Results are averaged across three runs with measured latency. Temp.\ order: temporal order; Act.\ count: action count; Time dur.: duration.}
\label{tab:qa_proactive}
\vspace{-6pt}
\end{table}

\subsubsection{Effect of Assistant Latency}
\label{sec:latency_results}
\begin{figure}[!t]
\centering
\includegraphics[width=\linewidth]{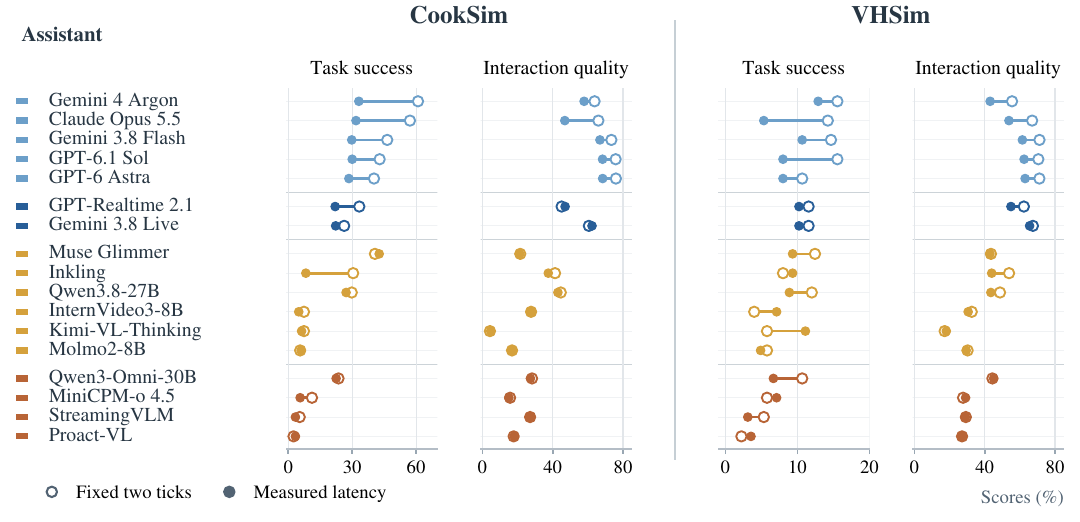}
\caption{\textbf{Effect of assistant latency on task success and interaction quality.} Open markers show the fixed two-tick setting; filled markers show measured latency. Lines connect the same model under the two settings. Scores are averaged across three independent runs.}
\label{fig:latency_shift}
\vspace{-6pt}
\end{figure}

We assess the effect of assistant response latency in CookSim and VHSim by comparing rollouts with measured latency against the fixed two-tick setting (\Cref{tab:assistant_latency,fig:latency_shift}). We use the same tasks, personas, and scheduled events in both settings, each averaged across three independent runs.

\begin{table}[!t]
\centering
\begingroup
\setlength{\tabcolsep}{0pt}
\renewcommand{\arraystretch}{1.06}
\rmfamily\fontsize{6.8}{8.7}\selectfont
\begin{tabular}{@{}L{.190\linewidth}K{ivgsuccess}{0.065\linewidth}K{ivgrecall}{0.052\linewidth}K{white}{0.047\linewidth}K{white}{0.047\linewidth}K{white}{0.047\linewidth}K{white}{0.047\linewidth}K{white}{0.047\linewidth}K{ivgrubric}{0.049\linewidth}@{\hspace{2pt}}K{ivgsuccess}{0.065\linewidth}K{ivgrecall}{0.052\linewidth}K{white}{0.047\linewidth}K{white}{0.047\linewidth}K{white}{0.047\linewidth}K{white}{0.047\linewidth}K{white}{0.047\linewidth}K{ivgrubric}{0.049\linewidth}@{}}
\toprule
 & \multicolumn{8}{c}{\bfseries CookSim} & \multicolumn{8}{c}{\bfseries VHSim}\\[2pt]
\cmidrule(lr){2-9}\cmidrule(lr){10-17}
 & \multicolumn{1}{c}{\multirow{2}{*}{\shortstack{In-time\\success}}} & \multicolumn{1}{c}{\multirow{2}{*}{\shortstack{Det.\\recall}}} & \multicolumn{6}{c}{\bfseries Interaction quality} & \multicolumn{1}{c}{\multirow{2}{*}{\shortstack{In-time\\success}}} & \multicolumn{1}{c}{\multirow{2}{*}{\shortstack{Det.\\recall}}} & \multicolumn{6}{c}{\bfseries Interaction quality}\\
\cmidrule(lr){4-9}\cmidrule(lr){12-17}
\textbf{Assistant} &  & & {\fontsize{6.2}{7.2}\selectfont\shortstack{Fac.\\grnd.}} & {\fontsize{6.2}{7.2}\selectfont\shortstack{Sit.\\rel.}} & {\fontsize{6.2}{7.2}\selectfont\shortstack{Act.\\guid.}} & {\fontsize{6.2}{7.2}\selectfont\shortstack{User\\intent}} & {\fontsize{6.2}{7.2}\selectfont\shortstack{Guid.\\conc.}} & {\fontsize{6.2}{7.2}\selectfont\shortstack{Overall}} &  & & {\fontsize{6.2}{7.2}\selectfont\shortstack{Fac.\\grnd.}} & {\fontsize{6.2}{7.2}\selectfont\shortstack{Sit.\\rel.}} & {\fontsize{6.2}{7.2}\selectfont\shortstack{Act.\\guid.}} & {\fontsize{6.2}{7.2}\selectfont\shortstack{User\\intent}} & {\fontsize{6.2}{7.2}\selectfont\shortstack{Guid.\\conc.}} & {\fontsize{6.2}{7.2}\selectfont\shortstack{Overall}}\\[3pt]
\midrule\rowcolor{white}\multicolumn{17}{@{}l@{}}{\bfseries Frontier API models}\\[2pt]
Gemini 4 Argon & \shortstack{\underline{33.1}\\[-1pt]{\fontsize{5.7}{6.2}\selectfont\textcolor{black!65}{(-27.8)}}} & \shortstack{\underline{66.9}\\[-1pt]{\fontsize{5.7}{6.2}\selectfont\textcolor{black!65}{(-13.3)}}} & \shortstack{65.3\\[-1pt]{\fontsize{5.7}{6.2}\selectfont\textcolor{black!65}{(-4.9)}}} & \shortstack{\underline{84.9}\\[-1pt]{\fontsize{5.7}{6.2}\selectfont\textcolor{black!65}{(-5.2)}}} & \shortstack{63.0\\[-1pt]{\fontsize{5.7}{6.2}\selectfont\textcolor{black!65}{(-8.4)}}} & \shortstack{86.0\\[-1pt]{\fontsize{5.7}{6.2}\selectfont\textcolor{black!65}{(+3.7)}}} & \shortstack{71.1\\[-1pt]{\fontsize{5.7}{6.2}\selectfont\textcolor{black!65}{(-5.8)}}} & \shortstack{57.9\\[-1pt]{\fontsize{5.7}{6.2}\selectfont\textcolor{black!65}{(-6.0)}}} & \shortstack{\textbf{12.9}\\[-1pt]{\fontsize{5.7}{6.2}\selectfont\textcolor{black!65}{(-2.7)}}} & \shortstack{\textbf{71.4}\\[-1pt]{\fontsize{5.7}{6.2}\selectfont\textcolor{black!65}{(-9.7)}}} & \shortstack{49.5\\[-1pt]{\fontsize{5.7}{6.2}\selectfont\textcolor{black!65}{(-10.7)}}} & \shortstack{79.1\\[-1pt]{\fontsize{5.7}{6.2}\selectfont\textcolor{black!65}{(-11.6)}}} & \shortstack{43.7\\[-1pt]{\fontsize{5.7}{6.2}\selectfont\textcolor{black!65}{(-17.1)}}} & \shortstack{77.0\\[-1pt]{\fontsize{5.7}{6.2}\selectfont\textcolor{black!65}{(-6.2)}}} & \shortstack{76.0\\[-1pt]{\fontsize{5.7}{6.2}\selectfont\textcolor{black!65}{(-8.8)}}} & \shortstack{43.1\\[-1pt]{\fontsize{5.7}{6.2}\selectfont\textcolor{black!65}{(-12.4)}}}\\[1pt]
Claude Opus 5.5 & \shortstack{31.8\\[-1pt]{\fontsize{5.7}{6.2}\selectfont\textcolor{black!65}{(-25.3)}}} & \shortstack{64.2\\[-1pt]{\fontsize{5.7}{6.2}\selectfont\textcolor{black!65}{(-21.7)}}} & \shortstack{52.3\\[-1pt]{\fontsize{5.7}{6.2}\selectfont\textcolor{black!65}{(-17.0)}}} & \shortstack{81.2\\[-1pt]{\fontsize{5.7}{6.2}\selectfont\textcolor{black!65}{(-11.2)}}} & \shortstack{52.5\\[-1pt]{\fontsize{5.7}{6.2}\selectfont\textcolor{black!65}{(-22.8)}}} & \shortstack{85.7\\[-1pt]{\fontsize{5.7}{6.2}\selectfont\textcolor{black!65}{(+1.9)}}} & \shortstack{79.2\\[-1pt]{\fontsize{5.7}{6.2}\selectfont\textcolor{black!65}{(-11.5)}}} & \shortstack{46.9\\[-1pt]{\fontsize{5.7}{6.2}\selectfont\textcolor{black!65}{(-19.1)}}} & \shortstack{5.3\\[-1pt]{\fontsize{5.7}{6.2}\selectfont\textcolor{black!65}{(-8.9)}}} & \shortstack{54.9\\[-1pt]{\fontsize{5.7}{6.2}\selectfont\textcolor{black!65}{(-7.7)}}} & \shortstack{60.0\\[-1pt]{\fontsize{5.7}{6.2}\selectfont\textcolor{black!65}{(-11.8)}}} & \shortstack{85.0\\[-1pt]{\fontsize{5.7}{6.2}\selectfont\textcolor{black!65}{(-8.8)}}} & \shortstack{47.5\\[-1pt]{\fontsize{5.7}{6.2}\selectfont\textcolor{black!65}{(-17.6)}}} & \shortstack{83.8\\[-1pt]{\fontsize{5.7}{6.2}\selectfont\textcolor{black!65}{(-4.6)}}} & \shortstack{85.3\\[-1pt]{\fontsize{5.7}{6.2}\selectfont\textcolor{black!65}{(-6.1)}}} & \shortstack{53.8\\[-1pt]{\fontsize{5.7}{6.2}\selectfont\textcolor{black!65}{(-13.2)}}}\\[1pt]
Gemini 3.8 Flash & \shortstack{29.8\\[-1pt]{\fontsize{5.7}{6.2}\selectfont\textcolor{black!65}{(-16.7)}}} & \shortstack{61.3\\[-1pt]{\fontsize{5.7}{6.2}\selectfont\textcolor{black!65}{(-16.6)}}} & \shortstack{\underline{77.6}\\[-1pt]{\fontsize{5.7}{6.2}\selectfont\textcolor{black!65}{(-5.9)}}} & \shortstack{77.8\\[-1pt]{\fontsize{5.7}{6.2}\selectfont\textcolor{black!65}{(-5.7)}}} & \shortstack{58.5\\[-1pt]{\fontsize{5.7}{6.2}\selectfont\textcolor{black!65}{(-7.0)}}} & \shortstack{66.9\\[-1pt]{\fontsize{5.7}{6.2}\selectfont\textcolor{black!65}{(+0.7)}}} & \shortstack{\underline{88.4}\\[-1pt]{\fontsize{5.7}{6.2}\selectfont\textcolor{black!65}{(-2.5)}}} & \shortstack{\underline{66.9}\\[-1pt]{\fontsize{5.7}{6.2}\selectfont\textcolor{black!65}{(-6.5)}}} & \shortstack{10.7\\[-1pt]{\fontsize{5.7}{6.2}\selectfont\textcolor{black!65}{(-4.0)}}} & \shortstack{57.8\\[-1pt]{\fontsize{5.7}{6.2}\selectfont\textcolor{black!65}{(-14.4)}}} & \shortstack{70.0\\[-1pt]{\fontsize{5.7}{6.2}\selectfont\textcolor{black!65}{(-9.2)}}} & \shortstack{82.2\\[-1pt]{\fontsize{5.7}{6.2}\selectfont\textcolor{black!65}{(-6.2)}}} & \shortstack{\textbf{53.2}\\[-1pt]{\fontsize{5.7}{6.2}\selectfont\textcolor{black!65}{(-10.2)}}} & \shortstack{72.2\\[-1pt]{\fontsize{5.7}{6.2}\selectfont\textcolor{black!65}{(-3.2)}}} & \shortstack{\textbf{90.2}\\[-1pt]{\fontsize{5.7}{6.2}\selectfont\textcolor{black!65}{(-2.5)}}} & \shortstack{61.4\\[-1pt]{\fontsize{5.7}{6.2}\selectfont\textcolor{black!65}{(-9.7)}}}\\[1pt]
GPT-6.1 Sol & \shortstack{30.0\\[-1pt]{\fontsize{5.7}{6.2}\selectfont\textcolor{black!65}{(-12.9)}}} & \shortstack{\textbf{67.5}\\[-1pt]{\fontsize{5.7}{6.2}\selectfont\textcolor{black!65}{(-12.5)}}} & \shortstack{74.8\\[-1pt]{\fontsize{5.7}{6.2}\selectfont\textcolor{black!65}{(-5.5)}}} & \shortstack{\textbf{87.6}\\[-1pt]{\fontsize{5.7}{6.2}\selectfont\textcolor{black!65}{(-4.8)}}} & \shortstack{\underline{69.2}\\[-1pt]{\fontsize{5.7}{6.2}\selectfont\textcolor{black!65}{(-8.9)}}} & \shortstack{\underline{90.6}\\[-1pt]{\fontsize{5.7}{6.2}\selectfont\textcolor{black!65}{(-0.6)}}} & \shortstack{86.3\\[-1pt]{\fontsize{5.7}{6.2}\selectfont\textcolor{black!65}{(-5.4)}}} & \shortstack{\textbf{68.4}\\[-1pt]{\fontsize{5.7}{6.2}\selectfont\textcolor{black!65}{(-7.5)}}} & \shortstack{8.0\\[-1pt]{\fontsize{5.7}{6.2}\selectfont\textcolor{black!65}{(-7.6)}}} & \shortstack{59.0\\[-1pt]{\fontsize{5.7}{6.2}\selectfont\textcolor{black!65}{(-1.2)}}} & \shortstack{71.7\\[-1pt]{\fontsize{5.7}{6.2}\selectfont\textcolor{black!65}{(-6.1)}}} & \shortstack{\textbf{86.2}\\[-1pt]{\fontsize{5.7}{6.2}\selectfont\textcolor{black!65}{(-6.7)}}} & \shortstack{44.3\\[-1pt]{\fontsize{5.7}{6.2}\selectfont\textcolor{black!65}{(-13.0)}}} & \shortstack{\underline{84.0}\\[-1pt]{\fontsize{5.7}{6.2}\selectfont\textcolor{black!65}{(-5.1)}}} & \shortstack{88.3\\[-1pt]{\fontsize{5.7}{6.2}\selectfont\textcolor{black!65}{(-3.9)}}} & \shortstack{62.4\\[-1pt]{\fontsize{5.7}{6.2}\selectfont\textcolor{black!65}{(-8.1)}}}\\[1pt]
GPT-6 Astra & \shortstack{28.4\\[-1pt]{\fontsize{5.7}{6.2}\selectfont\textcolor{black!65}{(-11.8)}}} & \shortstack{63.6\\[-1pt]{\fontsize{5.7}{6.2}\selectfont\textcolor{black!65}{(-16.7)}}} & \shortstack{74.0\\[-1pt]{\fontsize{5.7}{6.2}\selectfont\textcolor{black!65}{(-6.5)}}} & \shortstack{\textbf{87.6}\\[-1pt]{\fontsize{5.7}{6.2}\selectfont\textcolor{black!65}{(-5.3)}}} & \shortstack{\textbf{69.6}\\[-1pt]{\fontsize{5.7}{6.2}\selectfont\textcolor{black!65}{(-9.3)}}} & \shortstack{\textbf{92.4}\\[-1pt]{\fontsize{5.7}{6.2}\selectfont\textcolor{black!65}{(-1.2)}}} & \shortstack{\textbf{88.5}\\[-1pt]{\fontsize{5.7}{6.2}\selectfont\textcolor{black!65}{(-2.0)}}} & \shortstack{\textbf{68.4}\\[-1pt]{\fontsize{5.7}{6.2}\selectfont\textcolor{black!65}{(-7.5)}}} & \shortstack{8.0\\[-1pt]{\fontsize{5.7}{6.2}\selectfont\textcolor{black!65}{(-2.7)}}} & \shortstack{47.9\\[-1pt]{\fontsize{5.7}{6.2}\selectfont\textcolor{black!65}{(-14.2)}}} & \shortstack{71.0\\[-1pt]{\fontsize{5.7}{6.2}\selectfont\textcolor{black!65}{(-6.6)}}} & \shortstack{\underline{85.7}\\[-1pt]{\fontsize{5.7}{6.2}\selectfont\textcolor{black!65}{(-7.1)}}} & \shortstack{47.2\\[-1pt]{\fontsize{5.7}{6.2}\selectfont\textcolor{black!65}{(-12.3)}}} & \shortstack{\textbf{85.4}\\[-1pt]{\fontsize{5.7}{6.2}\selectfont\textcolor{black!65}{(-5.3)}}} & \shortstack{\underline{89.3}\\[-1pt]{\fontsize{5.7}{6.2}\selectfont\textcolor{black!65}{(-2.2)}}} & \shortstack{\underline{63.0}\\[-1pt]{\fontsize{5.7}{6.2}\selectfont\textcolor{black!65}{(-8.1)}}}\\[1pt]
\midrule\rowcolor{white}\multicolumn{17}{@{}l@{}}{\bfseries Real-time streaming API models}\\[2pt]
GPT-Realtime 2.1 & \shortstack{22.0\\[-1pt]{\fontsize{5.7}{6.2}\selectfont\textcolor{black!65}{(-11.3)}}} & \shortstack{60.6\\[-1pt]{\fontsize{5.7}{6.2}\selectfont\textcolor{black!65}{(-7.4)}}} & \shortstack{66.1\\[-1pt]{\fontsize{5.7}{6.2}\selectfont\textcolor{black!65}{(+0.1)}}} & \shortstack{70.7\\[-1pt]{\fontsize{5.7}{6.2}\selectfont\textcolor{black!65}{(+1.3)}}} & \shortstack{46.9\\[-1pt]{\fontsize{5.7}{6.2}\selectfont\textcolor{black!65}{(+0.7)}}} & \shortstack{54.2\\[-1pt]{\fontsize{5.7}{6.2}\selectfont\textcolor{black!65}{(+9.4)}}} & \shortstack{44.0\\[-1pt]{\fontsize{5.7}{6.2}\selectfont\textcolor{black!65}{(+6.8)}}} & \shortstack{47.1\\[-1pt]{\fontsize{5.7}{6.2}\selectfont\textcolor{black!65}{(+2.0)}}} & \shortstack{10.2\\[-1pt]{\fontsize{5.7}{6.2}\selectfont\textcolor{black!65}{(-1.3)}}} & \shortstack{35.5\\[-1pt]{\fontsize{5.7}{6.2}\selectfont\textcolor{black!65}{(-25.3)}}} & \shortstack{\underline{72.7}\\[-1pt]{\fontsize{5.7}{6.2}\selectfont\textcolor{black!65}{(-6.6)}}} & \shortstack{80.4\\[-1pt]{\fontsize{5.7}{6.2}\selectfont\textcolor{black!65}{(-6.5)}}} & \shortstack{\underline{49.9}\\[-1pt]{\fontsize{5.7}{6.2}\selectfont\textcolor{black!65}{(-14.4)}}} & \shortstack{71.8\\[-1pt]{\fontsize{5.7}{6.2}\selectfont\textcolor{black!65}{(+4.4)}}} & \shortstack{56.9\\[-1pt]{\fontsize{5.7}{6.2}\selectfont\textcolor{black!65}{(+0.4)}}} & \shortstack{54.9\\[-1pt]{\fontsize{5.7}{6.2}\selectfont\textcolor{black!65}{(-7.3)}}}\\[1pt]
Gemini 3.8 Live & \shortstack{22.2\\[-1pt]{\fontsize{5.7}{6.2}\selectfont\textcolor{black!65}{(-4.0)}}} & \shortstack{53.0\\[-1pt]{\fontsize{5.7}{6.2}\selectfont\textcolor{black!65}{(-5.4)}}} & \shortstack{\textbf{77.7}\\[-1pt]{\fontsize{5.7}{6.2}\selectfont\textcolor{black!65}{(+0.9)}}} & \shortstack{74.0\\[-1pt]{\fontsize{5.7}{6.2}\selectfont\textcolor{black!65}{(+2.6)}}} & \shortstack{46.3\\[-1pt]{\fontsize{5.7}{6.2}\selectfont\textcolor{black!65}{(+1.7)}}} & \shortstack{61.0\\[-1pt]{\fontsize{5.7}{6.2}\selectfont\textcolor{black!65}{(+5.9)}}} & \shortstack{84.9\\[-1pt]{\fontsize{5.7}{6.2}\selectfont\textcolor{black!65}{(+3.2)}}} & \shortstack{62.4\\[-1pt]{\fontsize{5.7}{6.2}\selectfont\textcolor{black!65}{(+1.8)}}} & \shortstack{10.2\\[-1pt]{\fontsize{5.7}{6.2}\selectfont\textcolor{black!65}{(-1.3)}}} & \shortstack{19.8\\[-1pt]{\fontsize{5.7}{6.2}\selectfont\textcolor{black!65}{(-4.8)}}} & \shortstack{\textbf{81.7}\\[-1pt]{\fontsize{5.7}{6.2}\selectfont\textcolor{black!65}{(-1.8)}}} & \shortstack{78.8\\[-1pt]{\fontsize{5.7}{6.2}\selectfont\textcolor{black!65}{(-2.7)}}} & \shortstack{46.4\\[-1pt]{\fontsize{5.7}{6.2}\selectfont\textcolor{black!65}{(-3.3)}}} & \shortstack{58.7\\[-1pt]{\fontsize{5.7}{6.2}\selectfont\textcolor{black!65}{(+0.2)}}} & \shortstack{85.5\\[-1pt]{\fontsize{5.7}{6.2}\selectfont\textcolor{black!65}{(+1.7)}}} & \shortstack{\textbf{65.6}\\[-1pt]{\fontsize{5.7}{6.2}\selectfont\textcolor{black!65}{(-1.9)}}}\\[1pt]
\midrule\rowcolor{white}\multicolumn{17}{@{}l@{}}{\bfseries Open-weight non-streaming VLMs}\\[2pt]
Muse Glimmer & \shortstack{\textbf{42.7}\\[-1pt]{\fontsize{5.7}{6.2}\selectfont\textcolor{black!65}{(+2.0)}}} & \shortstack{60.7\\[-1pt]{\fontsize{5.7}{6.2}\selectfont\textcolor{black!65}{(+0.2)}}} & \shortstack{28.5\\[-1pt]{\fontsize{5.7}{6.2}\selectfont\textcolor{black!65}{(-0.4)}}} & \shortstack{67.5\\[-1pt]{\fontsize{5.7}{6.2}\selectfont\textcolor{black!65}{(-1.1)}}} & \shortstack{32.6\\[-1pt]{\fontsize{5.7}{6.2}\selectfont\textcolor{black!65}{(-1.1)}}} & \shortstack{36.0\\[-1pt]{\fontsize{5.7}{6.2}\selectfont\textcolor{black!65}{(-0.5)}}} & \shortstack{27.5\\[-1pt]{\fontsize{5.7}{6.2}\selectfont\textcolor{black!65}{(-0.1)}}} & \shortstack{21.2\\[-1pt]{\fontsize{5.7}{6.2}\selectfont\textcolor{black!65}{(-0.3)}}} & \shortstack{9.3\\[-1pt]{\fontsize{5.7}{6.2}\selectfont\textcolor{black!65}{(-3.1)}}} & \shortstack{13.9\\[-1pt]{\fontsize{5.7}{6.2}\selectfont\textcolor{black!65}{(-0.4)}}} & \shortstack{52.8\\[-1pt]{\fontsize{5.7}{6.2}\selectfont\textcolor{black!65}{(-1.0)}}} & \shortstack{82.1\\[-1pt]{\fontsize{5.7}{6.2}\selectfont\textcolor{black!65}{(-1.0)}}} & \shortstack{48.2\\[-1pt]{\fontsize{5.7}{6.2}\selectfont\textcolor{black!65}{(-1.8)}}} & \shortstack{65.4\\[-1pt]{\fontsize{5.7}{6.2}\selectfont\textcolor{black!65}{(+3.7)}}} & \shortstack{50.8\\[-1pt]{\fontsize{5.7}{6.2}\selectfont\textcolor{black!65}{(+0.5)}}} & \shortstack{43.0\\[-1pt]{\fontsize{5.7}{6.2}\selectfont\textcolor{black!65}{(-0.6)}}}\\[1pt]
Inkling & \shortstack{8.2\\[-1pt]{\fontsize{5.7}{6.2}\selectfont\textcolor{black!65}{(-22.2)}}} & \shortstack{38.1\\[-1pt]{\fontsize{5.7}{6.2}\selectfont\textcolor{black!65}{(-41.4)}}} & \shortstack{50.4\\[-1pt]{\fontsize{5.7}{6.2}\selectfont\textcolor{black!65}{(-2.3)}}} & \shortstack{60.9\\[-1pt]{\fontsize{5.7}{6.2}\selectfont\textcolor{black!65}{(-11.6)}}} & \shortstack{29.5\\[-1pt]{\fontsize{5.7}{6.2}\selectfont\textcolor{black!65}{(-9.6)}}} & \shortstack{52.2\\[-1pt]{\fontsize{5.7}{6.2}\selectfont\textcolor{black!65}{(+7.7)}}} & \shortstack{67.6\\[-1pt]{\fontsize{5.7}{6.2}\selectfont\textcolor{black!65}{(-0.6)}}} & \shortstack{37.5\\[-1pt]{\fontsize{5.7}{6.2}\selectfont\textcolor{black!65}{(-3.8)}}} & \shortstack{9.3\\[-1pt]{\fontsize{5.7}{6.2}\selectfont\textcolor{black!65}{(+1.3)}}} & \shortstack{49.8\\[-1pt]{\fontsize{5.7}{6.2}\selectfont\textcolor{black!65}{(-29.2)}}} & \shortstack{55.4\\[-1pt]{\fontsize{5.7}{6.2}\selectfont\textcolor{black!65}{(-7.4)}}} & \shortstack{75.9\\[-1pt]{\fontsize{5.7}{6.2}\selectfont\textcolor{black!65}{(-10.2)}}} & \shortstack{33.4\\[-1pt]{\fontsize{5.7}{6.2}\selectfont\textcolor{black!65}{(-14.9)}}} & \shortstack{60.2\\[-1pt]{\fontsize{5.7}{6.2}\selectfont\textcolor{black!65}{(-3.6)}}} & \shortstack{77.8\\[-1pt]{\fontsize{5.7}{6.2}\selectfont\textcolor{black!65}{(-5.7)}}} & \shortstack{44.0\\[-1pt]{\fontsize{5.7}{6.2}\selectfont\textcolor{black!65}{(-9.9)}}}\\[1pt]
Qwen3.8-27B & \shortstack{27.1\\[-1pt]{\fontsize{5.7}{6.2}\selectfont\textcolor{black!65}{(-2.7)}}} & \shortstack{57.1\\[-1pt]{\fontsize{5.7}{6.2}\selectfont\textcolor{black!65}{(-3.6)}}} & \shortstack{50.4\\[-1pt]{\fontsize{5.7}{6.2}\selectfont\textcolor{black!65}{(-1.4)}}} & \shortstack{81.4\\[-1pt]{\fontsize{5.7}{6.2}\selectfont\textcolor{black!65}{(-2.4)}}} & \shortstack{50.9\\[-1pt]{\fontsize{5.7}{6.2}\selectfont\textcolor{black!65}{(-2.8)}}} & \shortstack{70.2\\[-1pt]{\fontsize{5.7}{6.2}\selectfont\textcolor{black!65}{(+0.1)}}} & \shortstack{66.4\\[-1pt]{\fontsize{5.7}{6.2}\selectfont\textcolor{black!65}{(-1.7)}}} & \shortstack{43.0\\[-1pt]{\fontsize{5.7}{6.2}\selectfont\textcolor{black!65}{(-1.6)}}} & \shortstack{8.9\\[-1pt]{\fontsize{5.7}{6.2}\selectfont\textcolor{black!65}{(-3.1)}}} & \shortstack{30.4\\[-1pt]{\fontsize{5.7}{6.2}\selectfont\textcolor{black!65}{(-1.4)}}} & \shortstack{49.2\\[-1pt]{\fontsize{5.7}{6.2}\selectfont\textcolor{black!65}{(-5.1)}}} & \shortstack{83.2\\[-1pt]{\fontsize{5.7}{6.2}\selectfont\textcolor{black!65}{(-4.9)}}} & \shortstack{48.3\\[-1pt]{\fontsize{5.7}{6.2}\selectfont\textcolor{black!65}{(-8.0)}}} & \shortstack{77.9\\[-1pt]{\fontsize{5.7}{6.2}\selectfont\textcolor{black!65}{(-1.9)}}} & \shortstack{72.3\\[-1pt]{\fontsize{5.7}{6.2}\selectfont\textcolor{black!65}{(-2.6)}}} & \shortstack{43.5\\[-1pt]{\fontsize{5.7}{6.2}\selectfont\textcolor{black!65}{(-5.2)}}}\\[1pt]
InternVideo3-8B & \shortstack{4.9\\[-1pt]{\fontsize{5.7}{6.2}\selectfont\textcolor{black!65}{(-2.4)}}} & \shortstack{14.7\\[-1pt]{\fontsize{5.7}{6.2}\selectfont\textcolor{black!65}{(-10.8)}}} & \shortstack{44.9\\[-1pt]{\fontsize{5.7}{6.2}\selectfont\textcolor{black!65}{(-1.6)}}} & \shortstack{33.9\\[-1pt]{\fontsize{5.7}{6.2}\selectfont\textcolor{black!65}{(-2.4)}}} & \shortstack{16.3\\[-1pt]{\fontsize{5.7}{6.2}\selectfont\textcolor{black!65}{(-0.5)}}} & \shortstack{18.1\\[-1pt]{\fontsize{5.7}{6.2}\selectfont\textcolor{black!65}{(+1.0)}}} & \shortstack{52.7\\[-1pt]{\fontsize{5.7}{6.2}\selectfont\textcolor{black!65}{(+4.4)}}} & \shortstack{27.7\\[-1pt]{\fontsize{5.7}{6.2}\selectfont\textcolor{black!65}{(0.0)}}} & \shortstack{7.1\\[-1pt]{\fontsize{5.7}{6.2}\selectfont\textcolor{black!65}{(+3.1)}}} & \shortstack{0.0\\[-1pt]{\fontsize{5.7}{6.2}\selectfont\textcolor{black!65}{(0.0)}}} & \shortstack{44.5\\[-1pt]{\fontsize{5.7}{6.2}\selectfont\textcolor{black!65}{(-4.2)}}} & \shortstack{52.5\\[-1pt]{\fontsize{5.7}{6.2}\selectfont\textcolor{black!65}{(-2.1)}}} & \shortstack{28.9\\[-1pt]{\fontsize{5.7}{6.2}\selectfont\textcolor{black!65}{(-2.3)}}} & \shortstack{29.1\\[-1pt]{\fontsize{5.7}{6.2}\selectfont\textcolor{black!65}{(+3.1)}}} & \shortstack{47.9\\[-1pt]{\fontsize{5.7}{6.2}\selectfont\textcolor{black!65}{(+2.8)}}} & \shortstack{30.6\\[-1pt]{\fontsize{5.7}{6.2}\selectfont\textcolor{black!65}{(-2.0)}}}\\[1pt]
Kimi-VL-Thinking & \shortstack{6.2\\[-1pt]{\fontsize{5.7}{6.2}\selectfont\textcolor{black!65}{(-1.1)}}} & \shortstack{46.1\\[-1pt]{\fontsize{5.7}{6.2}\selectfont\textcolor{black!65}{(-10.4)}}} & \shortstack{8.2\\[-1pt]{\fontsize{5.7}{6.2}\selectfont\textcolor{black!65}{(+0.9)}}} & \shortstack{26.7\\[-1pt]{\fontsize{5.7}{6.2}\selectfont\textcolor{black!65}{(-0.7)}}} & \shortstack{5.4\\[-1pt]{\fontsize{5.7}{6.2}\selectfont\textcolor{black!65}{(-0.2)}}} & \shortstack{14.0\\[-1pt]{\fontsize{5.7}{6.2}\selectfont\textcolor{black!65}{(-0.5)}}} & \shortstack{12.2\\[-1pt]{\fontsize{5.7}{6.2}\selectfont\textcolor{black!65}{(+0.9)}}} & \shortstack{4.8\\[-1pt]{\fontsize{5.7}{6.2}\selectfont\textcolor{black!65}{(+0.4)}}} & \shortstack{\underline{11.1}\\[-1pt]{\fontsize{5.7}{6.2}\selectfont\textcolor{black!65}{(+5.3)}}} & \shortstack{\underline{68.3}\\[-1pt]{\fontsize{5.7}{6.2}\selectfont\textcolor{black!65}{(-5.7)}}} & \shortstack{27.9\\[-1pt]{\fontsize{5.7}{6.2}\selectfont\textcolor{black!65}{(+1.2)}}} & \shortstack{52.4\\[-1pt]{\fontsize{5.7}{6.2}\selectfont\textcolor{black!65}{(+1.8)}}} & \shortstack{17.9\\[-1pt]{\fontsize{5.7}{6.2}\selectfont\textcolor{black!65}{(+0.5)}}} & \shortstack{29.0\\[-1pt]{\fontsize{5.7}{6.2}\selectfont\textcolor{black!65}{(+3.8)}}} & \shortstack{32.0\\[-1pt]{\fontsize{5.7}{6.2}\selectfont\textcolor{black!65}{(+0.8)}}} & \shortstack{18.2\\[-1pt]{\fontsize{5.7}{6.2}\selectfont\textcolor{black!65}{(+0.9)}}}\\[1pt]
Molmo2-8B & \shortstack{6.0\\[-1pt]{\fontsize{5.7}{6.2}\selectfont\textcolor{black!65}{(+0.4)}}} & \shortstack{4.5\\[-1pt]{\fontsize{5.7}{6.2}\selectfont\textcolor{black!65}{(-0.6)}}} & \shortstack{40.1\\[-1pt]{\fontsize{5.7}{6.2}\selectfont\textcolor{black!65}{(-1.8)}}} & \shortstack{18.7\\[-1pt]{\fontsize{5.7}{6.2}\selectfont\textcolor{black!65}{(+0.2)}}} & \shortstack{9.9\\[-1pt]{\fontsize{5.7}{6.2}\selectfont\textcolor{black!65}{(-0.1)}}} & \shortstack{15.6\\[-1pt]{\fontsize{5.7}{6.2}\selectfont\textcolor{black!65}{(+1.0)}}} & \shortstack{14.2\\[-1pt]{\fontsize{5.7}{6.2}\selectfont\textcolor{black!65}{(+1.9)}}} & \shortstack{16.8\\[-1pt]{\fontsize{5.7}{6.2}\selectfont\textcolor{black!65}{(-0.1)}}} & \shortstack{4.9\\[-1pt]{\fontsize{5.7}{6.2}\selectfont\textcolor{black!65}{(-0.9)}}} & \shortstack{0.0\\[-1pt]{\fontsize{5.7}{6.2}\selectfont\textcolor{black!65}{(-1.1)}}} & \shortstack{52.7\\[-1pt]{\fontsize{5.7}{6.2}\selectfont\textcolor{black!65}{(-1.8)}}} & \shortstack{39.9\\[-1pt]{\fontsize{5.7}{6.2}\selectfont\textcolor{black!65}{(-1.4)}}} & \shortstack{21.4\\[-1pt]{\fontsize{5.7}{6.2}\selectfont\textcolor{black!65}{(-1.5)}}} & \shortstack{12.9\\[-1pt]{\fontsize{5.7}{6.2}\selectfont\textcolor{black!65}{(-0.2)}}} & \shortstack{32.5\\[-1pt]{\fontsize{5.7}{6.2}\selectfont\textcolor{black!65}{(+0.4)}}} & \shortstack{29.5\\[-1pt]{\fontsize{5.7}{6.2}\selectfont\textcolor{black!65}{(-0.8)}}}\\[1pt]
\midrule\rowcolor{white}\multicolumn{17}{@{}l@{}}{\bfseries Open-weight streaming VLMs}\\[2pt]
Qwen3-Omni-30B & \shortstack{22.4\\[-1pt]{\fontsize{5.7}{6.2}\selectfont\textcolor{black!65}{(-1.1)}}} & \shortstack{42.5\\[-1pt]{\fontsize{5.7}{6.2}\selectfont\textcolor{black!65}{(+3.5)}}} & \shortstack{39.4\\[-1pt]{\fontsize{5.7}{6.2}\selectfont\textcolor{black!65}{(-1.3)}}} & \shortstack{51.3\\[-1pt]{\fontsize{5.7}{6.2}\selectfont\textcolor{black!65}{(-1.7)}}} & \shortstack{28.0\\[-1pt]{\fontsize{5.7}{6.2}\selectfont\textcolor{black!65}{(-0.8)}}} & \shortstack{23.7\\[-1pt]{\fontsize{5.7}{6.2}\selectfont\textcolor{black!65}{(-1.1)}}} & \shortstack{47.0\\[-1pt]{\fontsize{5.7}{6.2}\selectfont\textcolor{black!65}{(-0.4)}}} & \shortstack{27.3\\[-1pt]{\fontsize{5.7}{6.2}\selectfont\textcolor{black!65}{(-0.9)}}} & \shortstack{6.7\\[-1pt]{\fontsize{5.7}{6.2}\selectfont\textcolor{black!65}{(-4.0)}}} & \shortstack{4.9\\[-1pt]{\fontsize{5.7}{6.2}\selectfont\textcolor{black!65}{(+2.2)}}} & \shortstack{57.5\\[-1pt]{\fontsize{5.7}{6.2}\selectfont\textcolor{black!65}{(+0.6)}}} & \shortstack{75.1\\[-1pt]{\fontsize{5.7}{6.2}\selectfont\textcolor{black!65}{(+1.6)}}} & \shortstack{42.2\\[-1pt]{\fontsize{5.7}{6.2}\selectfont\textcolor{black!65}{(+0.4)}}} & \shortstack{41.1\\[-1pt]{\fontsize{5.7}{6.2}\selectfont\textcolor{black!65}{(+1.3)}}} & \shortstack{69.7\\[-1pt]{\fontsize{5.7}{6.2}\selectfont\textcolor{black!65}{(+2.4)}}} & \shortstack{45.1\\[-1pt]{\fontsize{5.7}{6.2}\selectfont\textcolor{black!65}{(+0.7)}}}\\[1pt]
MiniCPM-o 4.5 & \shortstack{5.6\\[-1pt]{\fontsize{5.7}{6.2}\selectfont\textcolor{black!65}{(-5.6)}}} & \shortstack{36.3\\[-1pt]{\fontsize{5.7}{6.2}\selectfont\textcolor{black!65}{(-3.3)}}} & \shortstack{28.0\\[-1pt]{\fontsize{5.7}{6.2}\selectfont\textcolor{black!65}{(-1.1)}}} & \shortstack{29.6\\[-1pt]{\fontsize{5.7}{6.2}\selectfont\textcolor{black!65}{(-2.3)}}} & \shortstack{9.2\\[-1pt]{\fontsize{5.7}{6.2}\selectfont\textcolor{black!65}{(-0.6)}}} & \shortstack{21.1\\[-1pt]{\fontsize{5.7}{6.2}\selectfont\textcolor{black!65}{(-1.0)}}} & \shortstack{36.6\\[-1pt]{\fontsize{5.7}{6.2}\selectfont\textcolor{black!65}{(+2.1)}}} & \shortstack{15.1\\[-1pt]{\fontsize{5.7}{6.2}\selectfont\textcolor{black!65}{(-0.7)}}} & \shortstack{7.1\\[-1pt]{\fontsize{5.7}{6.2}\selectfont\textcolor{black!65}{(+1.3)}}} & \shortstack{11.2\\[-1pt]{\fontsize{5.7}{6.2}\selectfont\textcolor{black!65}{(+1.4)}}} & \shortstack{47.5\\[-1pt]{\fontsize{5.7}{6.2}\selectfont\textcolor{black!65}{(+2.1)}}} & \shortstack{55.1\\[-1pt]{\fontsize{5.7}{6.2}\selectfont\textcolor{black!65}{(+1.3)}}} & \shortstack{20.7\\[-1pt]{\fontsize{5.7}{6.2}\selectfont\textcolor{black!65}{(+2.0)}}} & \shortstack{37.0\\[-1pt]{\fontsize{5.7}{6.2}\selectfont\textcolor{black!65}{(+2.8)}}} & \shortstack{34.3\\[-1pt]{\fontsize{5.7}{6.2}\selectfont\textcolor{black!65}{(-0.9)}}} & \shortstack{29.1\\[-1pt]{\fontsize{5.7}{6.2}\selectfont\textcolor{black!65}{(+1.4)}}}\\[1pt]
StreamingVLM & \shortstack{3.3\\[-1pt]{\fontsize{5.7}{6.2}\selectfont\textcolor{black!65}{(-2.0)}}} & \shortstack{1.0\\[-1pt]{\fontsize{5.7}{6.2}\selectfont\textcolor{black!65}{(+0.4)}}} & \shortstack{73.3\\[-1pt]{\fontsize{5.7}{6.2}\selectfont\textcolor{black!65}{(-0.6)}}} & \shortstack{7.1\\[-1pt]{\fontsize{5.7}{6.2}\selectfont\textcolor{black!65}{(+0.4)}}} & \shortstack{16.7\\[-1pt]{\fontsize{5.7}{6.2}\selectfont\textcolor{black!65}{(-0.2)}}} & \shortstack{29.1\\[-1pt]{\fontsize{5.7}{6.2}\selectfont\textcolor{black!65}{(-1.6)}}} & \shortstack{15.5\\[-1pt]{\fontsize{5.7}{6.2}\selectfont\textcolor{black!65}{(+0.6)}}} & \shortstack{27.2\\[-1pt]{\fontsize{5.7}{6.2}\selectfont\textcolor{black!65}{(0.0)}}} & \shortstack{3.1\\[-1pt]{\fontsize{5.7}{6.2}\selectfont\textcolor{black!65}{(-2.2)}}} & \shortstack{0.0\\[-1pt]{\fontsize{5.7}{6.2}\selectfont\textcolor{black!65}{(0.0)}}} & \shortstack{68.1\\[-1pt]{\fontsize{5.7}{6.2}\selectfont\textcolor{black!65}{(-0.3)}}} & \shortstack{14.4\\[-1pt]{\fontsize{5.7}{6.2}\selectfont\textcolor{black!65}{(+0.3)}}} & \shortstack{9.8\\[-1pt]{\fontsize{5.7}{6.2}\selectfont\textcolor{black!65}{(+2.1)}}} & \shortstack{24.4\\[-1pt]{\fontsize{5.7}{6.2}\selectfont\textcolor{black!65}{(+1.6)}}} & \shortstack{15.1\\[-1pt]{\fontsize{5.7}{6.2}\selectfont\textcolor{black!65}{(-0.7)}}} & \shortstack{29.0\\[-1pt]{\fontsize{5.7}{6.2}\selectfont\textcolor{black!65}{(-0.3)}}}\\[1pt]
Proact-VL & \shortstack{3.3\\[-1pt]{\fontsize{5.7}{6.2}\selectfont\textcolor{black!65}{(+0.9)}}} & \shortstack{1.0\\[-1pt]{\fontsize{5.7}{6.2}\selectfont\textcolor{black!65}{(+0.2)}}} & \shortstack{45.0\\[-1pt]{\fontsize{5.7}{6.2}\selectfont\textcolor{black!65}{(+1.5)}}} & \shortstack{10.2\\[-1pt]{\fontsize{5.7}{6.2}\selectfont\textcolor{black!65}{(-0.3)}}} & \shortstack{5.5\\[-1pt]{\fontsize{5.7}{6.2}\selectfont\textcolor{black!65}{(+0.4)}}} & \shortstack{10.9\\[-1pt]{\fontsize{5.7}{6.2}\selectfont\textcolor{black!65}{(-3.0)}}} & \shortstack{23.6\\[-1pt]{\fontsize{5.7}{6.2}\selectfont\textcolor{black!65}{(-0.2)}}} & \shortstack{18.1\\[-1pt]{\fontsize{5.7}{6.2}\selectfont\textcolor{black!65}{(+0.3)}}} & \shortstack{3.6\\[-1pt]{\fontsize{5.7}{6.2}\selectfont\textcolor{black!65}{(+1.3)}}} & \shortstack{0.0\\[-1pt]{\fontsize{5.7}{6.2}\selectfont\textcolor{black!65}{(0.0)}}} & \shortstack{58.5\\[-1pt]{\fontsize{5.7}{6.2}\selectfont\textcolor{black!65}{(+0.1)}}} & \shortstack{22.7\\[-1pt]{\fontsize{5.7}{6.2}\selectfont\textcolor{black!65}{(-0.4)}}} & \shortstack{7.0\\[-1pt]{\fontsize{5.7}{6.2}\selectfont\textcolor{black!65}{(0.0)}}} & \shortstack{15.9\\[-1pt]{\fontsize{5.7}{6.2}\selectfont\textcolor{black!65}{(-0.4)}}} & \shortstack{29.3\\[-1pt]{\fontsize{5.7}{6.2}\selectfont\textcolor{black!65}{(-1.2)}}} & \shortstack{27.0\\[-1pt]{\fontsize{5.7}{6.2}\selectfont\textcolor{black!65}{(-0.1)}}}\\[1pt]
\bottomrule
\end{tabular}\par
\endgroup

\caption{\textbf{Results with measured assistant latency.} Parentheses show changes from the fixed two-tick setting in percentage points. Results are averaged across three independent runs. }
\label{tab:assistant_latency}
\vspace{-6pt}
\end{table}

\noindent\textbf{Response latency substantially reduces in-time task success for frontier API models.}
Introducing measured latency lowers success for all five frontier API models in both environments. The accompanying decline in error prevention helps explain this loss (\Cref{tab:latency_prevention}): even when an assistant recognizes an impending mistake, its warning may arrive after the user has committed the action, incurring either task failure or additional steps to recover from errors.

\noindent\textbf{Accounting for response latency changes model rankings.}
Gemini 4 Argon achieves the highest task success in CookSim under fixed response timing, while Muse Glimmer leads when measured latency is considered. In CookSim, Gemini 4 Argon's success falls from 60.9\% to 33.1\%, while Muse Glimmer changes little. Measured latency also lowers interaction quality for the frontier API models in both environments, particularly in situational relevance and actionable guidance.

\begingroup
\setlength{\intextsep}{0pt}
\setlength{\columnsep}{10pt}
\Needspace{235pt}
\subsection{Validating Interactive Visual Gym via Human Study}
\label{sec:human_study}

\begin{wrapfigure}{r}{0.43\textwidth}
\centering
\begingroup
\setlength{\tabcolsep}{2pt}
\renewcommand{\arraystretch}{1.06}
\fontsize{7.2}{8.5}\selectfont
\begin{tabularx}{\linewidth}{@{}l>{\raggedright\arraybackslash}Xr@{}}
\toprule
Axis & Measure & Similarity\\
\midrule
\multirow{3}{*}{\itshape Communication}
 & Utterance length & 0.89\\
 & Speech rate & 0.05\\
 & Responsiveness & 0.05\\
\addlinespace[2pt]
\itshape Skill & Task success & 0.56\\
\multirow{2}{*}{\itshape Trust}
 & Suggestion uptake & 0.77\\
 & Disagreement & 0.88\\
\multirow{2}{*}{\itshape Interruption}
 & Guidance requests & 0.98\\
 & Ignored interruptions & 0.82\\
\multirow{2}{*}{\itshape Speed}
 & Action efficiency & 0.82\\
 & Parallelism & 0.00\\
\bottomrule
\end{tabularx}
\includegraphics[width=\linewidth]{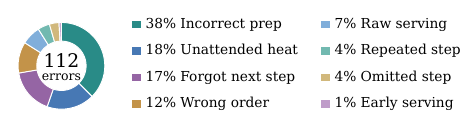}
\captionsetup{font=footnotesize,skip=5pt,justification=raggedright,singlelinecheck=false}
\caption{\textbf{Behavior comparison of real human and simulated human agent.} See \Cref{tab:behavior_definitions} in the appendix for measure definitions.}
\label{fig:human_comparison}
\endgroup
\vspace{-8pt}
\end{wrapfigure}%

To examine both evaluation validity and user-simulation fidelity, we conduct a human study in CookSim simulator.
45 participants interact with 3 anonymously labeled assistants and rank the interaction experience after completing for 3 rounds. Full study protocol provided in \Cref{app:human_study}.

\noindent\textbf{Interaction quality reflects human preferences.}
To test the validity of our interaction-quality metric, we compare each assistant's mean score with participants' rankings. The metric agrees with human preferences in \textbf{78.5\%} of participant--assistant-pair comparisons, corresponding to a pooled Kendall's $\tau$ of \textbf{+0.57}. This correspondence supports its use as a measure of the assistance experienced by users.

\noindent\textbf{Simulated users reproduce most of human behaviors.}
We compare human and simulated episodes along three dimensions: communication behavior, persona-dependent variation, and action error types (\Cref{fig:human_comparison}). We use binary rubrics to measure behavioral dimensions and compute the similarity between human and simulated users on each dimension (detailed in appendix). Simulated users reproduce human behavioral differences across most persona axes, with particularly close agreement for trust and interruption preferences. The engine also covers the error types observed in participants' action behaviors, including incorrect preparation, omitted or repeated steps, and assembly-order errors. Action parallelism match less well, possibly because the human speed groups show almost no difference in parallel cooking. We also found a clear deviation between the communication hebavior between human agent and real human: although their utterance lengths are similar, simulated users speak to assistants much more often than human participants. One possible reason is that the simulated user is invoked very frequently to reconsider whether to speak, repeatedly making conversation salient.

\endgroup

\begin{figure}[!ht]
  \centering
  \begingroup
    \phantomsubcaption\label{fig:qual_delay}
    \phantomsubcaption\label{fig:qual_repetition}
  \endgroup
  \includegraphics[width=\textwidth]{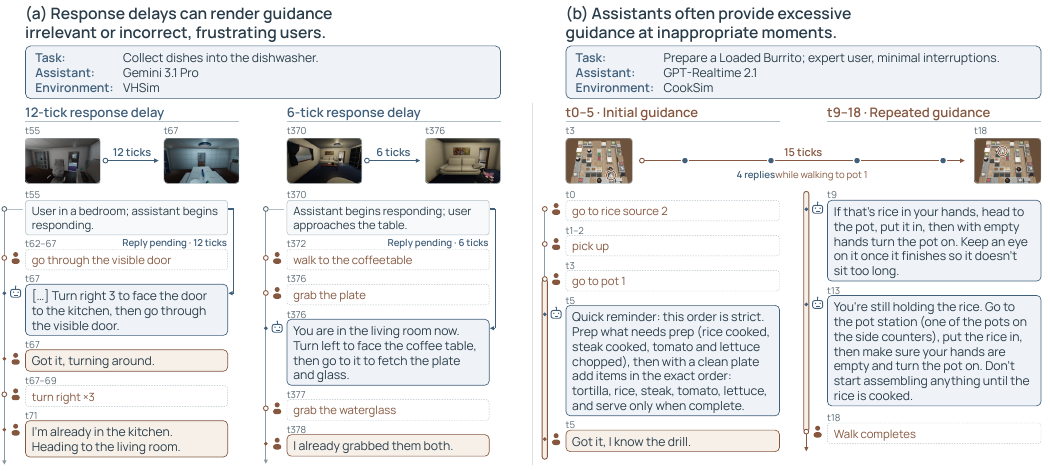}
    \vspace{-5pt}
  \caption{\textbf{Qualitative examples of assistant interaction trajectories.} It illustrates recurring problems across assistant types that our dynamic evaluation expose.}
  \label{fig:qualitative_trajectories}
  \vspace{-7pt}
\end{figure}

\Needspace{8\baselineskip}
\subsection{Interactive Visual Gym as an RL environment}
\label{sec:rl}
\begin{figure}[!ht]
    \centering
    \includegraphics[width=\linewidth]
    {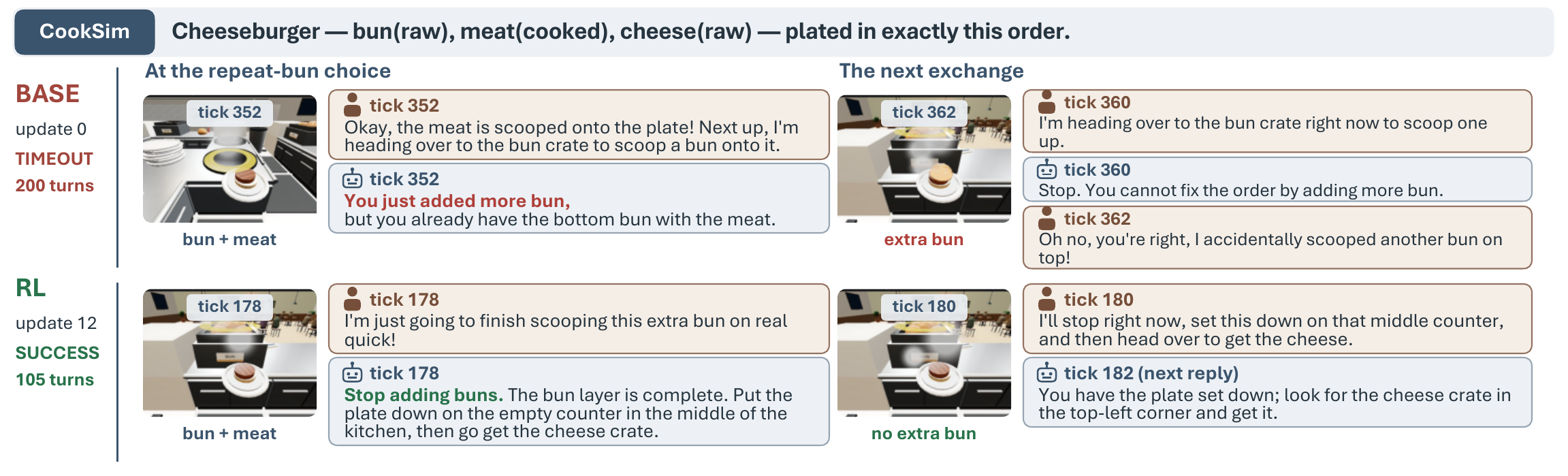}
      \vspace{-5pt}
    \caption{\textbf{Qualitative example from RL.}
    The base assistant misperceives the task state and fails to prevent the error. After RL, the assistant gives timely guidance that prevents the error.}
    
    \label{fig:rl_cooksim}
  \vspace{-3pt}
\end{figure}
We train Qwen3.5-4B with GRPO on CookSim, using Gemini 3.7/3.8 Flash as a classic-novice user. We split CookSim's 150 scenarios into 120 for training and 30 for validation. Each update uses 24 episodes: 6 scenarios with 4 episodes each. Because binary in-time success is sparse, we use the shaped reward $
r = \mathbb{I}[\text{success}] + 0.30\,\mathbb{I}[\text{error prevented}]
-0.02\min(N_{\text{false flags}},15)
-0.05\min(T/200,1).
$
Here, $T$ counts interaction turns. Task success dominates ($r\geq0.65$ for successes; $r\leq0.30$ for completed failures); shaping encourages timely intervention and discourages false flags and excess turns (\Cref{app:rl}).

\begingroup
\setlength{\columnsep}{10pt}
\Needspace{150pt}
\begin{wraptable}{r}{0.58\textwidth}
\vspace{-8pt}
\centering
\begingroup
\fontsize{8}{9.5}\selectfont
\setlength{\tabcolsep}{2pt}\renewcommand{\arraystretch}{1.13}
\begin{tabular*}{\linewidth}{@{\extracolsep{\fill}}llrrr@{}}
\toprule
User & Assistant & \shortstack{In-time\\success} & \shortstack{Det.\\recall} & \shortstack{Interaction\\quality}\\
\midrule
\multirow{2}{*}{\shortstack[l]{Gemini 3.7\\Flash}}
& Qwen3.5-4B & $33.3 \pm 3.3$ & $\mathbf{49.0} \pm 11.9$ & $29.6 \pm 4.8$\\
& \quad + RL & $\mathbf{42.2} \pm 6.9$ & $34.9 \pm 6.7$ & $\mathbf{31.1} \pm 0.4$\\
\addlinespace[3pt]
\multirow{2}{*}{\shortstack[l]{Gemini 3.8\\Flash}}
& Qwen3.5-4B & $64.4 \pm 6.9$ & $\mathbf{43.0} \pm 14.5$ & $24.3 \pm 2.1$\\
& \quad + RL & $\mathbf{67.8} \pm 7.7$ & $28.7 \pm 3.4$ & $\mathbf{30.2} \pm 0.3$\\
\bottomrule
\end{tabular*}
\endgroup

\captionsetup{font=footnotesize,skip=5pt,justification=raggedright,singlelinecheck=false}
\caption{\textbf{CookSim validation results before and after RL, by simulated user model.} + RL denotes 12 updates. Scores are percentages, reported as mean $\pm$ SD. Bold marks higher means within pairs.}
\label{tab:rl_cooksim_results}
\vspace{-6pt}
\end{wraptable}

Across three evaluation passes on CookSim's held-out development-validation scenarios with a Gemini 3.7 Flash user, mean in-time task success rate rises from 33.3\% to 42.2\% after 12 updates, an increase of 8.9 percentage points (\Cref{tab:rl_cooksim_results}). A separate experiment with a Gemini 3.8 Flash user improves mean in-time task success rate by 3.3 percentage points (64.4\% to 67.8\%) and mean interaction quality by 5.9 points (24.3\% to 30.2\%). Mean detection recall declines in both settings, possibly because the reward favors selective flagging that can miss uncertain errors, consistent with false flags falling from 517 to 391 across 90 episodes with a Gemini 3.8 Flash user. \Cref{fig:rl_cooksim} illustrates one case where the trained assistant gives timely guidance that prevents an error. These fixed-scenario results show that Interactive Visual Gym supplies a useful learning signal for interactive assistance, paving the foundation for future work to demonstrate generalization across tasks, user models, or domains.
\par\endgroup

\section{Conclusion}
We present Interactive Visual Gym, a framework for closed-loop evaluation and training of visual interaction models through joint simulation of task environments and human users. Across three
  domains, we find substantial limitations in current VLM to serve as reliable assistants. We also validated our interaction paradigm with human study. Finally, a preliminary reinforcement-learning experiment shows that the environments can improve assistant performance.

\subsection*{Acknowledgments}

This work was supported in part by Thinking Machines Lab's \href{https://thinkingmachines.ai/news/tinker-research-and-teaching-grants/}{Tinker Research Grant}.

\subsection*{AI use statement}

We used generative AI tools, including coding agents, to assist with implementing our simulators and evaluation pipeline, generating synthetic benchmark cases, refining experiments, and interpreting results. We also used them for literature search, figure creation, code editing, and drafting and revising manuscript text. We reviewed AI-assisted outputs and take responsibility for the final code, data, figures, and claims.

\subsection*{Ethics statement}

Our human study took place in a browser-based cooking simulator; participants did not perform physical cooking tasks. The study recorded action traces, text exchanges, and preferences, which we analyze and report in aggregate. Because free-text exchanges may contain personal information, any release of raw traces must follow the study's consent and privacy procedures. Performance in our simulators should not be taken as evidence that an assistant is safe for real-world use.

\subsection*{Reproducibility statement}

We describe environment and benchmark construction in \Cref{app:environments}, assistant settings in \Cref{app:assistance}, user prompts in \Cref{app:prompts}, and event placement and scoring in \Cref{app:questions,app:evaluation_metrics,app:rubric}. The appendix also reports the human-study protocol (\Cref{app:human_study}), variation across independent runs (\Cref{app:run_variation}), and the reinforcement-learning configuration (\Cref{app:rl}).

\bibliography{refs}
\bibliographystyle{iclr2027_conference}

\clearpage
\appendix
\section*{Appendix Contents}
\suppressfloats[t]
\begingroup
\setlength{\tabcolsep}{0pt}
\renewcommand{\arraystretch}{1.12}
\noindent\begin{tabularx}{\linewidth}{@{}Xr@{}}
\hyperref[app:user_model_ablation]{\ref*{app:user_model_ablation}\quad Sensitivity to the Simulated-User Model} & \pageref*{app:user_model_ablation}\\
\hyperref[app:run_variation]{\ref*{app:run_variation}\quad Variation Across Independent Runs} & \pageref*{app:run_variation}\\
\hyperref[app:rl]{\ref*{app:rl}\quad Reinforcement-Learning Setup and Results} & \pageref*{app:rl}\\
\hyperref[app:evaluation_metrics]{\ref*{app:evaluation_metrics}\quad Task Success and Error Detection} & \pageref*{app:evaluation_metrics}\\
\hyperref[app:questions]{\ref*{app:questions}\quad Temporal Questions and Proactive Requests} & \pageref*{app:questions}\\
\hyperref[app:human_study]{\ref*{app:human_study}\quad Human Study: Protocol and Behavioral Comparisons} & \pageref*{app:human_study}\\
\hyperref[app:environments]{\ref*{app:environments}\quad Environment and Benchmark Construction} & \pageref*{app:environments}\\
\hyperref[app:assistance]{\ref*{app:assistance}\quad Assistant Settings} & \pageref*{app:assistance}\\
\hyperref[app:rubric]{\ref*{app:rubric}\quad Interaction-Quality Rubric and Judging} & \pageref*{app:rubric}\\
\hyperref[app:prompts]{\ref*{app:prompts}\quad User Harness and Prompt Design} & \pageref*{app:prompts}\\
\hyperref[app:qualitative]{\ref*{app:qualitative}\quad Qualitative Interaction Trajectories} & \pageref*{app:qualitative}\\
\end{tabularx}
\endgroup
\medskip
\FloatBarrier

\section{Sensitivity to the Simulated-User Model}
\label{app:results}
\label{app:user_model_ablation}
We examine how the choice of base model for the simulated human agent affects evaluation results. We fix the visual interaction model to Gemini 3.7 Flash and vary the model controlling the simulated user in CookSim, evaluating on the same subset of 90 benchmark episodes (\Cref{tab:user_model_ablation}).

\begin{table}[ht]\centering\small
\setlength{\tabcolsep}{4pt}\renewcommand{\arraystretch}{1.1}
\begin{tabularx}{\linewidth}{@{}Xrrr@{}}
\toprule
User model & In-time success (\%) & Any success (\%) & Words/episode\\\midrule
Gemini 3.1 Pro & 28.9 & 56.7 & 228.6\\
GPT-6 Astra & 23.3 & 48.9 & 215.2\\
Gemini 3.7 Flash & 21.1 & 52.2 & 147.1\\
Qwen3.8-27B & 2.2 & 27.8 & 423.7\\
\bottomrule\end{tabularx}
\caption{\textbf{Sensitivity to the simulated-user model.} Any success ignores the completion deadline. Word counts refer to the simulated user, not the assistant.}
\label{tab:user_model_ablation}
\end{table}

We find that GPT-6 Astra and Gemini 3.7 Flash yield similar task-success rates when used to evaluate the same assistant, whereas Gemini 3.1 Pro differ a bit. Local open-weight models yield much lower task-success rates, potentially reflecting smaller models' difficulty following the complex instructions required to elicit the intended user behavior.

\FloatBarrier

\section{Variation Across Independent Runs}
\label{app:run_variation}

We examine the variability of closed-loop evaluation by running each assistant three times with the same tasks, personas, and scheduled events. \Cref{tab:runs_cooksim,tab:runs_vhhome,tab:runs_screensim} report in-time success, detection recall, and interaction quality for each run, together with their mean and population standard deviation (SD). We evaluate CookSim and VHSim under both fixed two-tick and measured assistant delay. All scores are percentages.

\begingroup
\scriptsize
\setlength{\tabcolsep}{7.8pt}\renewcommand{\arraystretch}{1.10}
\setlength{\LTleft}{\fill}
\setlength{\LTright}{\fill}
\setlength{\LTcapwidth}{\linewidth}
\begin{longtable}{@{}lrrrrrrrrr@{}}
\toprule
& \multicolumn{3}{c}{In-time success (\%)} & \multicolumn{3}{c}{Detection recall (\%)} & \multicolumn{3}{c}{Interaction quality (\%)}\\
\cmidrule(lr){2-4}\cmidrule(lr){5-7}\cmidrule(lr){8-10}
Assistant & R1 / R2 / R3 & & Mean $\pm$ SD & R1 / R2 / R3 & & Mean $\pm$ SD & R1 / R2 / R3 & & Mean $\pm$ SD\\
\endfirsthead
\multicolumn{10}{@{}l}{\tablename~\ref{tab:runs_cooksim}\ (continued)}\\[3pt]
\toprule
& \multicolumn{3}{c}{In-time success (\%)} & \multicolumn{3}{c}{Detection recall (\%)} & \multicolumn{3}{c}{Interaction quality (\%)}\\
\cmidrule(lr){2-4}\cmidrule(lr){5-7}\cmidrule(lr){8-10}
Assistant & R1 / R2 / R3 & & Mean $\pm$ SD & R1 / R2 / R3 & & Mean $\pm$ SD & R1 / R2 / R3 & & Mean $\pm$ SD\\
\endhead
\midrule\multicolumn{10}{@{}l}{\bfseries Fixed two-tick delay}\\[2pt]
Gemini 4 Argon & \multicolumn{2}{c}{64.0 / 58.7 / 60.0} & 60.9\,$\pm$\,2.3 & \multicolumn{2}{c}{81.1 / 80.6 / 79.0} & 80.2\,$\pm$\,0.9 & \multicolumn{2}{c}{63.5 / 64.5 / 63.5} & 63.8\,$\pm$\,0.5\\
Claude Opus 5.5 & \multicolumn{2}{c}{59.3 / 60.0 / 52.0} & 57.1\,$\pm$\,3.6 & \multicolumn{2}{c}{87.0 / 87.0 / 83.9} & 86.0\,$\pm$\,1.5 & \multicolumn{2}{c}{65.1 / 66.8 / 66.0} & 66.0\,$\pm$\,0.7\\
Gemini 3.1 Pro & \multicolumn{2}{c}{46.0 / 51.3 / 47.3} & 48.2\,$\pm$\,2.3 & \multicolumn{2}{c}{83.4 / 87.0 / 84.8} & 85.1\,$\pm$\,1.5 & \multicolumn{2}{c}{36.0 / 37.1 / 38.2} & 37.1\,$\pm$\,0.9\\
Gemini 3.8 Flash & \multicolumn{2}{c}{48.0 / 46.0 / 45.3} & 46.4\,$\pm$\,1.1 & \multicolumn{2}{c}{77.7 / 77.7 / 78.3} & 77.9\,$\pm$\,0.3 & \multicolumn{2}{c}{73.7 / 72.8 / 73.6} & 73.4\,$\pm$\,0.4\\
GPT-6.1 Sol & \multicolumn{2}{c}{45.3 / 42.7 / 40.7} & 42.9\,$\pm$\,1.9 & \multicolumn{2}{c}{79.6 / 80.6 / 79.6} & 79.9\,$\pm$\,0.5 & \multicolumn{2}{c}{76.2 / 75.1 / 76.1} & 75.8\,$\pm$\,0.5\\
GPT-6 Astra & \multicolumn{2}{c}{42.0 / 38.7 / 40.0} & 40.2\,$\pm$\,1.4 & \multicolumn{2}{c}{80.6 / 80.4 / 80.1} & 80.4\,$\pm$\,0.2 & \multicolumn{2}{c}{77.2 / 75.5 / 75.2} & 76.0\,$\pm$\,0.9\\
\addlinespace[2pt]
GPT-Realtime 2.1 & \multicolumn{2}{c}{34.0 / 32.0 / 34.0} & 33.3\,$\pm$\,0.9 & \multicolumn{2}{c}{70.0 / 68.1 / 66.0} & 68.0\,$\pm$\,1.6 & \multicolumn{2}{c}{43.8 / 45.8 / 45.6} & 45.1\,$\pm$\,0.9\\
Gemini 3.8 Live & \multicolumn{2}{c}{22.7 / 28.0 / 28.0} & 26.2\,$\pm$\,2.5 & \multicolumn{2}{c}{56.1 / 60.3 / 59.0} & 58.5\,$\pm$\,1.7 & \multicolumn{2}{c}{60.2 / 62.1 / 59.4} & 60.6\,$\pm$\,1.2\\
\addlinespace[2pt]
Muse Glimmer & \multicolumn{2}{c}{44.7 / 42.0 / 35.3} & 40.7\,$\pm$\,3.9 & \multicolumn{2}{c}{61.3 / 63.2 / 57.0} & 60.5\,$\pm$\,2.6 & \multicolumn{2}{c}{21.2 / 22.1 / 21.3} & 21.6\,$\pm$\,0.4\\
Inkling & \multicolumn{2}{c}{28.7 / 31.3 / 31.3} & 30.4\,$\pm$\,1.3 & \multicolumn{2}{c}{79.8 / 78.2 / 80.5} & 79.5\,$\pm$\,0.9 & \multicolumn{2}{c}{41.6 / 41.1 / 41.3} & 41.3\,$\pm$\,0.2\\
Qwen3.8-27B & \multicolumn{2}{c}{36.7 / 26.0 / 26.7} & 29.8\,$\pm$\,4.9 & \multicolumn{2}{c}{62.7 / 61.4 / 57.7} & 60.6\,$\pm$\,2.1 & \multicolumn{2}{c}{44.3 / 44.2 / 45.4} & 44.6\,$\pm$\,0.5\\
InternVideo3-8B & \multicolumn{2}{c}{9.3 / 5.3 / 7.3} & 7.3\,$\pm$\,1.6 & \multicolumn{2}{c}{23.0 / 25.8 / 28.0} & 25.6\,$\pm$\,2.1 & \multicolumn{2}{c}{26.3 / 28.3 / 28.5} & 27.7\,$\pm$\,1.0\\
Kimi-VL-Thinking & \multicolumn{2}{c}{9.3 / 6.7 / 6.0} & 7.3\,$\pm$\,1.4 & \multicolumn{2}{c}{55.6 / 56.1 / 58.1} & 56.6\,$\pm$\,1.1 & \multicolumn{2}{c}{4.4 / 4.3 / 4.2} & 4.3\,$\pm$\,0.0\\
Molmo2-8B & \multicolumn{2}{c}{6.0 / 4.0 / 6.7} & 5.6\,$\pm$\,1.1 & \multicolumn{2}{c}{5.4 / 3.9 / 5.9} & 5.1\,$\pm$\,0.9 & \multicolumn{2}{c}{17.0 / 16.8 / 16.8} & 16.9\,$\pm$\,0.1\\
\addlinespace[2pt]
Qwen3-Omni-30B & \multicolumn{2}{c}{27.3 / 25.3 / 18.0} & 23.6\,$\pm$\,4.0 & \multicolumn{2}{c}{39.9 / 40.2 / 36.8} & 39.0\,$\pm$\,1.5 & \multicolumn{2}{c}{27.6 / 28.8 / 28.4} & 28.3\,$\pm$\,0.5\\
MiniCPM-o 4.5 & \multicolumn{2}{c}{10.0 / 10.7 / 12.7} & 11.1\,$\pm$\,1.1 & \multicolumn{2}{c}{39.0 / 37.6 / 42.2} & 39.6\,$\pm$\,1.9 & \multicolumn{2}{c}{15.6 / 15.7 / 16.0} & 15.8\,$\pm$\,0.1\\
StreamingVLM & \multicolumn{2}{c}{5.3 / 6.0 / 4.7} & 5.3\,$\pm$\,0.5 & \multicolumn{2}{c}{1.1 / 0.3 / 0.6} & 0.7\,$\pm$\,0.3 & \multicolumn{2}{c}{27.2 / 26.6 / 27.6} & 27.2\,$\pm$\,0.4\\
Proact-VL & \multicolumn{2}{c}{2.7 / 2.0 / 2.7} & 2.4\,$\pm$\,0.3 & \multicolumn{2}{c}{0.5 / 0.6 / 1.2} & 0.8\,$\pm$\,0.3 & \multicolumn{2}{c}{18.1 / 17.0 / 18.2} & 17.8\,$\pm$\,0.5\\
\addlinespace[2pt]
\midrule\multicolumn{10}{@{}l}{\bfseries Measured assistant delay}\\[2pt]
Gemini 4 Argon & \multicolumn{2}{c}{35.3 / 31.3 / 32.7} & 33.1\,$\pm$\,1.7 & \multicolumn{2}{c}{69.9 / 65.2 / 65.6} & 66.9\,$\pm$\,2.1 & \multicolumn{2}{c}{57.6 / 57.6 / 58.3} & 57.9\,$\pm$\,0.3\\
Claude Opus 5.5 & \multicolumn{2}{c}{25.3 / 32.0 / 38.0} & 31.8\,$\pm$\,5.2 & \multicolumn{2}{c}{62.4 / 64.1 / 66.2} & 64.2\,$\pm$\,1.6 & \multicolumn{2}{c}{47.1 / 48.7 / 44.9} & 46.9\,$\pm$\,1.6\\
Gemini 3.1 Pro & \multicolumn{2}{c}{21.3 / 22.0 / 18.0} & 20.4\,$\pm$\,1.7 & \multicolumn{2}{c}{53.3 / 48.3 / 44.5} & 48.7\,$\pm$\,3.6 & \multicolumn{2}{c}{33.8 / 36.0 / 34.5} & 34.8\,$\pm$\,0.9\\
Gemini 3.8 Flash & \multicolumn{2}{c}{28.7 / 32.0 / 28.7} & 29.8\,$\pm$\,1.6 & \multicolumn{2}{c}{60.2 / 61.8 / 62.0} & 61.3\,$\pm$\,0.8 & \multicolumn{2}{c}{66.5 / 66.1 / 68.0} & 66.9\,$\pm$\,0.8\\
GPT-6.1 Sol & \multicolumn{2}{c}{28.7 / 32.7 / 28.7} & 30.0\,$\pm$\,1.9 & \multicolumn{2}{c}{68.2 / 68.5 / 65.6} & 67.5\,$\pm$\,1.3 & \multicolumn{2}{c}{69.5 / 67.3 / 68.3} & 68.4\,$\pm$\,0.9\\
GPT-6 Astra & \multicolumn{2}{c}{28.0 / 25.3 / 32.0} & 28.4\,$\pm$\,2.7 & \multicolumn{2}{c}{63.4 / 60.9 / 66.7} & 63.6\,$\pm$\,2.4 & \multicolumn{2}{c}{68.2 / 66.0 / 71.1} & 68.4\,$\pm$\,2.1\\
\addlinespace[2pt]
GPT-Realtime 2.1 & \multicolumn{2}{c}{17.3 / 22.7 / 26.0} & 22.0\,$\pm$\,3.6 & \multicolumn{2}{c}{56.0 / 62.4 / 63.5} & 60.6\,$\pm$\,3.3 & \multicolumn{2}{c}{46.5 / 47.2 / 47.5} & 47.1\,$\pm$\,0.4\\
Gemini 3.8 Live & \multicolumn{2}{c}{18.7 / 22.0 / 26.0} & 22.2\,$\pm$\,3.0 & \multicolumn{2}{c}{52.4 / 53.2 / 53.6} & 53.0\,$\pm$\,0.5 & \multicolumn{2}{c}{63.4 / 62.2 / 61.6} & 62.4\,$\pm$\,0.7\\
\addlinespace[2pt]
Muse Glimmer & \multicolumn{2}{c}{43.3 / 39.3 / 45.3} & 42.7\,$\pm$\,2.5 & \multicolumn{2}{c}{63.0 / 57.6 / 61.5} & 60.7\,$\pm$\,2.3 & \multicolumn{2}{c}{21.2 / 21.1 / 21.4} & 21.2\,$\pm$\,0.1\\
Inkling & \multicolumn{2}{c}{5.3 / 8.7 / 10.7} & 8.2\,$\pm$\,2.2 & \multicolumn{2}{c}{37.9 / 37.1 / 39.3} & 38.1\,$\pm$\,0.9 & \multicolumn{2}{c}{38.4 / 37.9 / 36.2} & 37.5\,$\pm$\,0.9\\
Qwen3.8-27B & \multicolumn{2}{c}{32.7 / 24.0 / 24.7} & 27.1\,$\pm$\,3.9 & \multicolumn{2}{c}{55.8 / 58.3 / 57.1} & 57.1\,$\pm$\,1.0 & \multicolumn{2}{c}{43.1 / 42.0 / 43.8} & 43.0\,$\pm$\,0.7\\
InternVideo3-8B & \multicolumn{2}{c}{4.7 / 4.0 / 6.0} & 4.9\,$\pm$\,0.8 & \multicolumn{2}{c}{15.5 / 15.7 / 13.0} & 14.7\,$\pm$\,1.2 & \multicolumn{2}{c}{27.7 / 28.1 / 27.1} & 27.7\,$\pm$\,0.4\\
Kimi-VL-Thinking & \multicolumn{2}{c}{6.0 / 6.0 / 6.7} & 6.2\,$\pm$\,0.3 & \multicolumn{2}{c}{48.2 / 45.8 / 44.4} & 46.1\,$\pm$\,1.5 & \multicolumn{2}{c}{4.6 / 4.9 / 4.8} & 4.8\,$\pm$\,0.2\\
Molmo2-8B & \multicolumn{2}{c}{6.7 / 4.0 / 7.3} & 6.0\,$\pm$\,1.4 & \multicolumn{2}{c}{3.7 / 5.4 / 4.4} & 4.5\,$\pm$\,0.7 & \multicolumn{2}{c}{17.8 / 15.4 / 17.2} & 16.8\,$\pm$\,1.0\\
\addlinespace[2pt]
Qwen3-Omni-30B & \multicolumn{2}{c}{24.0 / 24.7 / 18.7} & 22.4\,$\pm$\,2.7 & \multicolumn{2}{c}{40.4 / 45.4 / 41.6} & 42.5\,$\pm$\,2.1 & \multicolumn{2}{c}{28.9 / 26.2 / 27.0} & 27.3\,$\pm$\,1.1\\
MiniCPM-o 4.5 & \multicolumn{2}{c}{6.7 / 4.7 / 5.3} & 5.6\,$\pm$\,0.8 & \multicolumn{2}{c}{39.3 / 36.2 / 33.3} & 36.3\,$\pm$\,2.4 & \multicolumn{2}{c}{15.5 / 15.1 / 14.7} & 15.1\,$\pm$\,0.3\\
StreamingVLM & \multicolumn{2}{c}{4.7 / 2.0 / 3.3} & 3.3\,$\pm$\,1.1 & \multicolumn{2}{c}{1.4 / 0.8 / 0.8} & 1.0\,$\pm$\,0.3 & \multicolumn{2}{c}{26.9 / 27.7 / 26.9} & 27.2\,$\pm$\,0.4\\
Proact-VL & \multicolumn{2}{c}{4.0 / 3.3 / 2.7} & 3.3\,$\pm$\,0.5 & \multicolumn{2}{c}{1.4 / 0.9 / 0.8} & 1.0\,$\pm$\,0.3 & \multicolumn{2}{c}{18.1 / 17.7 / 18.5} & 18.1\,$\pm$\,0.3\\
\addlinespace[2pt]
\bottomrule
\caption{\textbf{Independent-run results in CookSim.} Scores are percentages for three independent runs (R1--R3), with their mean and population standard deviation (SD).}
\label{tab:runs_cooksim}\\
\end{longtable}
\endgroup

\begingroup
\scriptsize
\setlength{\tabcolsep}{7.8pt}\renewcommand{\arraystretch}{1.10}
\setlength{\LTleft}{\fill}
\setlength{\LTright}{\fill}
\setlength{\LTcapwidth}{\linewidth}
\begin{longtable}{@{}lrrrrrrrrr@{}}
\toprule
& \multicolumn{3}{c}{In-time success (\%)} & \multicolumn{3}{c}{Detection recall (\%)} & \multicolumn{3}{c}{Interaction quality (\%)}\\
\cmidrule(lr){2-4}\cmidrule(lr){5-7}\cmidrule(lr){8-10}
Assistant & R1 / R2 / R3 & & Mean $\pm$ SD & R1 / R2 / R3 & & Mean $\pm$ SD & R1 / R2 / R3 & & Mean $\pm$ SD\\
\endfirsthead
\multicolumn{10}{@{}l}{\tablename~\ref{tab:runs_vhhome}\ (continued)}\\[3pt]
\toprule
& \multicolumn{3}{c}{In-time success (\%)} & \multicolumn{3}{c}{Detection recall (\%)} & \multicolumn{3}{c}{Interaction quality (\%)}\\
\cmidrule(lr){2-4}\cmidrule(lr){5-7}\cmidrule(lr){8-10}
Assistant & R1 / R2 / R3 & & Mean $\pm$ SD & R1 / R2 / R3 & & Mean $\pm$ SD & R1 / R2 / R3 & & Mean $\pm$ SD\\
\endhead
\midrule\multicolumn{10}{@{}l}{\bfseries Fixed two-tick delay}\\[2pt]
Gemini 4 Argon & \multicolumn{2}{c}{20.0 / 13.3 / 13.3} & 15.6\,$\pm$\,3.1 & \multicolumn{2}{c}{81.4 / 79.8 / 82.3} & 81.2\,$\pm$\,1.0 & \multicolumn{2}{c}{57.5 / 54.8 / 54.4} & 55.6\,$\pm$\,1.4\\
Claude Opus 5.5 & \multicolumn{2}{c}{14.7 / 13.3 / 14.7} & 14.2\,$\pm$\,0.6 & \multicolumn{2}{c}{63.9 / 64.9 / 59.1} & 62.6\,$\pm$\,2.5 & \multicolumn{2}{c}{66.2 / 68.2 / 66.6} & 67.0\,$\pm$\,0.9\\
Gemini 3.1 Pro & \multicolumn{2}{c}{17.3 / 12.0 / 14.7} & 14.7\,$\pm$\,2.2 & \multicolumn{2}{c}{89.6 / 90.1 / 89.3} & 89.6\,$\pm$\,0.3 & \multicolumn{2}{c}{53.8 / 56.4 / 54.8} & 55.0\,$\pm$\,1.1\\
Gemini 3.8 Flash & \multicolumn{2}{c}{16.0 / 17.3 / 10.7} & 14.7\,$\pm$\,2.9 & \multicolumn{2}{c}{73.5 / 73.7 / 69.4} & 72.2\,$\pm$\,2.0 & \multicolumn{2}{c}{69.5 / 71.4 / 72.6} & 71.2\,$\pm$\,1.3\\
GPT-6.1 Sol & \multicolumn{2}{c}{17.3 / 16.0 / 13.3} & 15.6\,$\pm$\,1.7 & \multicolumn{2}{c}{56.7 / 64.1 / 60.0} & 60.3\,$\pm$\,3.0 & \multicolumn{2}{c}{70.0 / 70.8 / 70.8} & 70.5\,$\pm$\,0.4\\
GPT-6 Astra & \multicolumn{2}{c}{10.7 / 9.3 / 12.0} & 10.7\,$\pm$\,1.1 & \multicolumn{2}{c}{63.6 / 60.4 / 62.3} & 62.1\,$\pm$\,1.3 & \multicolumn{2}{c}{72.1 / 70.0 / 71.3} & 71.1\,$\pm$\,0.9\\
\addlinespace[2pt]
GPT-Realtime 2.1 & \multicolumn{2}{c}{9.3 / 12.0 / 13.3} & 11.6\,$\pm$\,1.7 & \multicolumn{2}{c}{58.5 / 56.6 / 67.3} & 60.8\,$\pm$\,4.7 & \multicolumn{2}{c}{63.2 / 61.2 / 62.4} & 62.3\,$\pm$\,0.8\\
Gemini 3.8 Live & \multicolumn{2}{c}{12.0 / 14.7 / 8.0} & 11.6\,$\pm$\,2.7 & \multicolumn{2}{c}{18.6 / 25.5 / 29.8} & 24.6\,$\pm$\,4.6 & \multicolumn{2}{c}{68.3 / 66.0 / 68.1} & 67.4\,$\pm$\,1.0\\
\addlinespace[2pt]
Muse Glimmer & \multicolumn{2}{c}{12.0 / 14.7 / 10.7} & 12.4\,$\pm$\,1.7 & \multicolumn{2}{c}{17.5 / 14.3 / 11.1} & 14.3\,$\pm$\,2.6 & \multicolumn{2}{c}{43.2 / 42.9 / 44.6} & 43.6\,$\pm$\,0.7\\
Inkling & \multicolumn{2}{c}{9.3 / 9.3 / 5.3} & 8.0\,$\pm$\,1.9 & \multicolumn{2}{c}{78.6 / 81.1 / 77.4} & 79.0\,$\pm$\,1.5 & \multicolumn{2}{c}{54.3 / 54.4 / 53.0} & 53.9\,$\pm$\,0.6\\
Qwen3.8-27B & \multicolumn{2}{c}{10.7 / 14.7 / 10.7} & 12.0\,$\pm$\,1.9 & \multicolumn{2}{c}{31.6 / 34.3 / 29.4} & 31.8\,$\pm$\,2.0 & \multicolumn{2}{c}{48.8 / 48.4 / 48.9} & 48.7\,$\pm$\,0.2\\
InternVideo3-8B & \multicolumn{2}{c}{2.7 / 5.3 / 4.0} & 4.0\,$\pm$\,1.1 & \multicolumn{2}{c}{0.0 / 0.0 / 0.0} & 0.0\,$\pm$\,0.0 & \multicolumn{2}{c}{30.1 / 35.2 / 32.6} & 32.6\,$\pm$\,2.1\\
Kimi-VL-Thinking & \multicolumn{2}{c}{5.3 / 8.0 / 4.0} & 5.8\,$\pm$\,1.7 & \multicolumn{2}{c}{64.9 / 83.5 / 73.6} & 74.0\,$\pm$\,7.6 & \multicolumn{2}{c}{17.6 / 16.4 / 17.7} & 17.2\,$\pm$\,0.6\\
Molmo2-8B & \multicolumn{2}{c}{6.7 / 5.3 / 5.3} & 5.8\,$\pm$\,0.6 & \multicolumn{2}{c}{2.1 / 0.0 / 1.1} & 1.1\,$\pm$\,0.9 & \multicolumn{2}{c}{31.9 / 27.6 / 31.4} & 30.3\,$\pm$\,1.9\\
\addlinespace[2pt]
Qwen3-Omni-30B & \multicolumn{2}{c}{12.0 / 10.7 / 9.3} & 10.7\,$\pm$\,1.1 & \multicolumn{2}{c}{3.9 / 3.1 / 1.1} & 2.7\,$\pm$\,1.2 & \multicolumn{2}{c}{44.5 / 44.2 / 44.4} & 44.4\,$\pm$\,0.1\\
MiniCPM-o 4.5 & \multicolumn{2}{c}{5.3 / 8.0 / 4.0} & 5.8\,$\pm$\,1.7 & \multicolumn{2}{c}{6.2 / 11.2 / 11.8} & 9.8\,$\pm$\,2.5 & \multicolumn{2}{c}{27.0 / 26.7 / 29.4} & 27.7\,$\pm$\,1.2\\
StreamingVLM & \multicolumn{2}{c}{6.7 / 5.3 / 4.0} & 5.3\,$\pm$\,1.1 & \multicolumn{2}{c}{0.0 / 0.0 / 0.0} & 0.0\,$\pm$\,0.0 & \multicolumn{2}{c}{29.5 / 28.7 / 29.7} & 29.3\,$\pm$\,0.5\\
Proact-VL & \multicolumn{2}{c}{2.7 / 2.7 / 1.3} & 2.2\,$\pm$\,0.6 & \multicolumn{2}{c}{0.0 / 0.0 / 0.0} & 0.0\,$\pm$\,0.0 & \multicolumn{2}{c}{26.4 / 27.1 / 28.0} & 27.1\,$\pm$\,0.7\\
\addlinespace[2pt]
\midrule\multicolumn{10}{@{}l}{\bfseries Measured assistant delay}\\[2pt]
Gemini 4 Argon & \multicolumn{2}{c}{13.3 / 13.3 / 12.0} & 12.9\,$\pm$\,0.6 & \multicolumn{2}{c}{70.9 / 73.4 / 70.0} & 71.4\,$\pm$\,1.4 & \multicolumn{2}{c}{41.9 / 43.0 / 44.4} & 43.1\,$\pm$\,1.0\\
Claude Opus 5.5 & \multicolumn{2}{c}{9.3 / 1.3 / 5.3} & 5.3\,$\pm$\,3.3 & \multicolumn{2}{c}{59.8 / 53.6 / 51.4} & 54.9\,$\pm$\,3.6 & \multicolumn{2}{c}{54.1 / 55.0 / 52.2} & 53.8\,$\pm$\,1.2\\
Gemini 3.1 Pro & \multicolumn{2}{c}{9.3 / 2.7 / 6.7} & 6.2\,$\pm$\,2.7 & \multicolumn{2}{c}{74.4 / 64.3 / 63.6} & 67.5\,$\pm$\,4.9 & \multicolumn{2}{c}{43.9 / 42.6 / 42.0} & 42.8\,$\pm$\,0.8\\
Gemini 3.8 Flash & \multicolumn{2}{c}{12.0 / 6.7 / 13.3} & 10.7\,$\pm$\,2.9 & \multicolumn{2}{c}{55.0 / 57.4 / 61.0} & 57.8\,$\pm$\,2.5 & \multicolumn{2}{c}{64.9 / 59.6 / 59.9} & 61.4\,$\pm$\,2.4\\
GPT-6.1 Sol & \multicolumn{2}{c}{6.7 / 9.3 / 8.0} & 8.0\,$\pm$\,1.1 & \multicolumn{2}{c}{59.0 / 59.3 / 58.8} & 59.0\,$\pm$\,0.2 & \multicolumn{2}{c}{61.8 / 62.5 / 63.0} & 62.4\,$\pm$\,0.5\\
GPT-6 Astra & \multicolumn{2}{c}{8.0 / 9.3 / 6.7} & 8.0\,$\pm$\,1.1 & \multicolumn{2}{c}{46.3 / 45.8 / 51.5} & 47.9\,$\pm$\,2.6 & \multicolumn{2}{c}{62.5 / 62.2 / 64.3} & 63.0\,$\pm$\,1.0\\
\addlinespace[2pt]
GPT-Realtime 2.1 & \multicolumn{2}{c}{9.3 / 8.0 / 13.3} & 10.2\,$\pm$\,2.3 & \multicolumn{2}{c}{34.5 / 34.1 / 37.9} & 35.5\,$\pm$\,1.7 & \multicolumn{2}{c}{55.8 / 53.2 / 55.8} & 54.9\,$\pm$\,1.2\\
Gemini 3.8 Live & \multicolumn{2}{c}{13.3 / 9.3 / 8.0} & 10.2\,$\pm$\,2.3 & \multicolumn{2}{c}{24.1 / 14.5 / 20.8} & 19.8\,$\pm$\,4.0 & \multicolumn{2}{c}{65.3 / 65.5 / 65.9} & 65.6\,$\pm$\,0.3\\
\addlinespace[2pt]
Muse Glimmer & \multicolumn{2}{c}{5.3 / 13.3 / 9.3} & 9.3\,$\pm$\,3.3 & \multicolumn{2}{c}{9.3 / 19.8 / 12.4} & 13.9\,$\pm$\,4.4 & \multicolumn{2}{c}{43.3 / 43.0 / 42.6} & 43.0\,$\pm$\,0.3\\
Inkling & \multicolumn{2}{c}{9.3 / 8.0 / 10.7} & 9.3\,$\pm$\,1.1 & \multicolumn{2}{c}{51.6 / 45.2 / 52.8} & 49.8\,$\pm$\,3.3 & \multicolumn{2}{c}{44.9 / 45.4 / 41.7} & 44.0\,$\pm$\,1.6\\
Qwen3.8-27B & \multicolumn{2}{c}{10.7 / 8.0 / 8.0} & 8.9\,$\pm$\,1.3 & \multicolumn{2}{c}{31.6 / 34.9 / 24.8} & 30.4\,$\pm$\,4.2 & \multicolumn{2}{c}{43.5 / 43.2 / 43.9} & 43.5\,$\pm$\,0.3\\
InternVideo3-8B & \multicolumn{2}{c}{6.7 / 6.7 / 8.0} & 7.1\,$\pm$\,0.6 & \multicolumn{2}{c}{0.0 / 0.0 / 0.0} & 0.0\,$\pm$\,0.0 & \multicolumn{2}{c}{32.2 / 28.4 / 31.2} & 30.6\,$\pm$\,1.6\\
Kimi-VL-Thinking & \multicolumn{2}{c}{9.3 / 17.3 / 6.7} & 11.1\,$\pm$\,4.5 & \multicolumn{2}{c}{70.7 / 67.0 / 67.3} & 68.3\,$\pm$\,1.7 & \multicolumn{2}{c}{18.6 / 18.3 / 17.7} & 18.2\,$\pm$\,0.4\\
Molmo2-8B & \multicolumn{2}{c}{6.7 / 4.0 / 4.0} & 4.9\,$\pm$\,1.3 & \multicolumn{2}{c}{0.0 / 0.0 / 0.0} & 0.0\,$\pm$\,0.0 & \multicolumn{2}{c}{31.8 / 27.7 / 28.9} & 29.5\,$\pm$\,1.7\\
\addlinespace[2pt]
Qwen3-Omni-30B & \multicolumn{2}{c}{6.7 / 6.7 / 6.7} & 6.7\,$\pm$\,0.0 & \multicolumn{2}{c}{6.4 / 6.1 / 2.2} & 4.9\,$\pm$\,1.9 & \multicolumn{2}{c}{45.9 / 44.0 / 45.3} & 45.1\,$\pm$\,0.8\\
MiniCPM-o 4.5 & \multicolumn{2}{c}{8.0 / 5.3 / 8.0} & 7.1\,$\pm$\,1.3 & \multicolumn{2}{c}{15.6 / 8.9 / 9.0} & 11.2\,$\pm$\,3.1 & \multicolumn{2}{c}{29.3 / 29.6 / 28.5} & 29.1\,$\pm$\,0.5\\
StreamingVLM & \multicolumn{2}{c}{4.0 / 4.0 / 1.3} & 3.1\,$\pm$\,1.3 & \multicolumn{2}{c}{0.0 / 0.0 / 0.0} & 0.0\,$\pm$\,0.0 & \multicolumn{2}{c}{29.4 / 29.6 / 28.0} & 29.0\,$\pm$\,0.7\\
Proact-VL & \multicolumn{2}{c}{6.7 / 4.0 / 0.0} & 3.6\,$\pm$\,2.7 & \multicolumn{2}{c}{0.0 / 0.0 / 0.0} & 0.0\,$\pm$\,0.0 & \multicolumn{2}{c}{27.7 / 27.5 / 25.8} & 27.0\,$\pm$\,0.9\\
\addlinespace[2pt]
\bottomrule
\caption{\textbf{Independent-run results in VHSim.} Scores are percentages for three independent runs (R1--R3), with their mean and population standard deviation (SD).}
\label{tab:runs_vhhome}\\
\end{longtable}
\endgroup

\begingroup
\scriptsize
\setlength{\tabcolsep}{7.8pt}\renewcommand{\arraystretch}{1.10}
\setlength{\LTleft}{\fill}
\setlength{\LTright}{\fill}
\setlength{\LTcapwidth}{\linewidth}
\begin{longtable}{@{}lrrrrrrrrr@{}}
\toprule
& \multicolumn{3}{c}{In-time success (\%)} & \multicolumn{3}{c}{Detection recall (\%)} & \multicolumn{3}{c}{Interaction quality (\%)}\\
\cmidrule(lr){2-4}\cmidrule(lr){5-7}\cmidrule(lr){8-10}
Assistant & R1 / R2 / R3 & & Mean $\pm$ SD & R1 / R2 / R3 & & Mean $\pm$ SD & R1 / R2 / R3 & & Mean $\pm$ SD\\
\endfirsthead
\multicolumn{10}{@{}l}{\tablename~\ref{tab:runs_screensim}\ (continued)}\\[3pt]
\toprule
& \multicolumn{3}{c}{In-time success (\%)} & \multicolumn{3}{c}{Detection recall (\%)} & \multicolumn{3}{c}{Interaction quality (\%)}\\
\cmidrule(lr){2-4}\cmidrule(lr){5-7}\cmidrule(lr){8-10}
Assistant & R1 / R2 / R3 & & Mean $\pm$ SD & R1 / R2 / R3 & & Mean $\pm$ SD & R1 / R2 / R3 & & Mean $\pm$ SD\\
\endhead
\midrule\multicolumn{10}{@{}l}{\bfseries ScreenSim}\\[2pt]
Gemini 4 Argon & \multicolumn{2}{c}{82.2 / 86.7 / 82.2} & 83.7\,$\pm$\,2.1 & \multicolumn{2}{c}{100.0 / 100.0 / 100.0} & 100.0\,$\pm$\,0.0 & \multicolumn{2}{c}{84.7 / 87.6 / 89.1} & 87.1\,$\pm$\,1.8\\
Claude Opus 5.5 & \multicolumn{2}{c}{86.7 / 90.0 / 91.1} & 89.3\,$\pm$\,1.9 & \multicolumn{2}{c}{100.0 / 100.0 / 100.0} & 100.0\,$\pm$\,0.0 & \multicolumn{2}{c}{82.7 / 85.4 / 84.0} & 84.0\,$\pm$\,1.1\\
Gemini 3.1 Pro & \multicolumn{2}{c}{85.6 / 84.4 / 88.9} & 86.3\,$\pm$\,1.9 & \multicolumn{2}{c}{99.4 / 100.0 / 99.4} & 99.6\,$\pm$\,0.3 & \multicolumn{2}{c}{85.7 / 83.7 / 85.2} & 84.9\,$\pm$\,0.9\\
Gemini 3.8 Flash & \multicolumn{2}{c}{78.9 / 75.6 / 81.1} & 78.5\,$\pm$\,2.3 & \multicolumn{2}{c}{97.6 / 97.6 / 98.3} & 97.8\,$\pm$\,0.3 & \multicolumn{2}{c}{89.4 / 89.5 / 89.3} & 89.4\,$\pm$\,0.1\\
GPT-6.1 Sol & \multicolumn{2}{c}{85.6 / 88.9 / 87.8} & 87.4\,$\pm$\,1.4 & \multicolumn{2}{c}{100.0 / 100.0 / 100.0} & 100.0\,$\pm$\,0.0 & \multicolumn{2}{c}{86.7 / 85.7 / 87.2} & 86.5\,$\pm$\,0.6\\
GPT-6 Astra & \multicolumn{2}{c}{91.1 / 92.2 / 87.8} & 90.4\,$\pm$\,1.9 & \multicolumn{2}{c}{100.0 / 100.0 / 100.0} & 100.0\,$\pm$\,0.0 & \multicolumn{2}{c}{86.1 / 86.0 / 86.7} & 86.3\,$\pm$\,0.3\\
\addlinespace[2pt]
GPT-Realtime 2.1 & \multicolumn{2}{c}{64.4 / 67.8 / 67.8} & 66.7\,$\pm$\,1.6 & \multicolumn{2}{c}{100.0 / 100.0 / 95.9} & 98.6\,$\pm$\,1.9 & \multicolumn{2}{c}{84.7 / 85.1 / 83.7} & 84.5\,$\pm$\,0.6\\
Gemini 3.8 Live & \multicolumn{2}{c}{68.9 / 70.0 / 76.7} & 71.9\,$\pm$\,3.4 & \multicolumn{2}{c}{95.3 / 96.4 / 96.6} & 96.1\,$\pm$\,0.6 & \multicolumn{2}{c}{88.4 / 89.4 / 89.5} & 89.1\,$\pm$\,0.5\\
\addlinespace[2pt]
Muse Glimmer & \multicolumn{2}{c}{43.3 / 46.7 / 42.2} & 44.1\,$\pm$\,1.9 & \multicolumn{2}{c}{95.9 / 94.8 / 97.6} & 96.1\,$\pm$\,1.2 & \multicolumn{2}{c}{55.3 / 50.4 / 55.8} & 53.8\,$\pm$\,2.4\\
Inkling & \multicolumn{2}{c}{47.8 / 52.2 / 43.3} & 47.8\,$\pm$\,3.6 & \multicolumn{2}{c}{97.1 / 99.4 / 98.8} & 98.4\,$\pm$\,1.0 & \multicolumn{2}{c}{76.2 / 76.6 / 78.2} & 77.0\,$\pm$\,0.8\\
Qwen3.8-27B & \multicolumn{2}{c}{65.6 / 67.8 / 70.0} & 67.8\,$\pm$\,1.8 & \multicolumn{2}{c}{93.8 / 97.7 / 97.2} & 96.2\,$\pm$\,1.7 & \multicolumn{2}{c}{86.5 / 87.1 / 85.0} & 86.2\,$\pm$\,0.9\\
InternVideo3-8B & \multicolumn{2}{c}{1.1 / 0.0 / 0.0} & 0.4\,$\pm$\,0.5 & \multicolumn{2}{c}{39.3 / 38.3 / 45.6} & 41.1\,$\pm$\,3.2 & \multicolumn{2}{c}{33.1 / 32.0 / 36.7} & 33.9\,$\pm$\,2.0\\
Kimi-VL-Thinking & \multicolumn{2}{c}{11.1 / 7.8 / 10.0} & 9.6\,$\pm$\,1.4 & \multicolumn{2}{c}{82.3 / 78.1 / 79.7} & 80.0\,$\pm$\,1.7 & \multicolumn{2}{c}{32.8 / 32.7 / 33.5} & 33.0\,$\pm$\,0.3\\
Molmo2-8B & \multicolumn{2}{c}{2.2 / 0.0 / 0.0} & 0.7\,$\pm$\,1.0 & \multicolumn{2}{c}{16.7 / 24.3 / 23.6} & 21.5\,$\pm$\,3.4 & \multicolumn{2}{c}{38.1 / 39.5 / 40.0} & 39.2\,$\pm$\,0.8\\
\addlinespace[2pt]
Qwen3-Omni-30B & \multicolumn{2}{c}{13.3 / 14.4 / 17.8} & 15.2\,$\pm$\,1.9 & \multicolumn{2}{c}{54.8 / 59.1 / 61.0} & 58.3\,$\pm$\,2.6 & \multicolumn{2}{c}{57.4 / 57.6 / 59.5} & 58.2\,$\pm$\,0.9\\
MiniCPM-o 4.5 & \multicolumn{2}{c}{2.2 / 3.3 / 2.2} & 2.6\,$\pm$\,0.5 & \multicolumn{2}{c}{42.0 / 46.4 / 47.1} & 45.2\,$\pm$\,2.2 & \multicolumn{2}{c}{40.1 / 37.5 / 37.8} & 38.5\,$\pm$\,1.1\\
StreamingVLM & \multicolumn{2}{c}{0.0 / 0.0 / 0.0} & 0.0\,$\pm$\,0.0 & \multicolumn{2}{c}{2.7 / 0.0 / 0.0} & 0.9\,$\pm$\,1.3 & \multicolumn{2}{c}{36.4 / 38.5 / 36.5} & 37.2\,$\pm$\,1.0\\
Proact-VL & \multicolumn{2}{c}{0.0 / 0.0 / 0.0} & 0.0\,$\pm$\,0.0 & \multicolumn{2}{c}{0.0 / 1.4 / 4.2} & 1.9\,$\pm$\,1.7 & \multicolumn{2}{c}{25.8 / 29.5 / 28.5} & 27.9\,$\pm$\,1.6\\
\addlinespace[2pt]
\bottomrule
\caption{\textbf{Independent-run results in ScreenSim.} Scores are percentages for three independent runs (R1--R3), with their mean and population standard deviation (SD).}
\label{tab:runs_screensim}\\
\end{longtable}
\endgroup

\FloatBarrier
\section{Reinforcement-Learning Setup and Results}
\label{app:rl}

These proof-of-concept RL experiments are separate from the multi-model benchmark and user-model ablation. The main CookSim validation comparison appears in \Cref{tab:rl_cooksim_results}; \Cref{tab:rl_cooksim_splits,tab:rl_screensim_no_manual} summarize additional CookSim and ScreenSim results.

\paragraph{CookSim.}
The assistant is Qwen3.5-4B, trained with GRPO while freezing the vision encoder. The language model uses Adam with a constant learning rate of $5\times10^{-7}$, zero weight decay, and no KL or entropy regularization. Generation uses temperature 0.8, top-$p=1.0$, a maximum of 512 output tokens, and a 16,384-token context window. Gemini 3.7 Flash simulates a classic-novice user.

The training set contains 120 scenarios specifying a recipe, layout, and scheduled user error. An update samples six scenario groups with five episodes each and retains the first four valid completions per group, yielding 24 training episodes. The reward is
\begin{equation}
\begin{aligned}
r ={}& \mathbf{1}[\mathrm{success}]
       +0.30\,\mathbf{1}[\mathrm{error\ prevented}]\\
     &{}-0.02\,\min\!\left(N_{\mathrm{false\ flags}},15\right)
       -0.05\,\min\!\left(T/200,1\right)
\end{aligned}
\end{equation}
Here, $T$ is the number of interaction turns.

\paragraph{CookSim evaluation.}
The original evaluation uses three stochastic passes over 120 training and 30 held-out development-validation scenarios. Validation variants share recipe/layout cells with training; there is no separate frozen test set. Success is CookSim's native \texttt{won} outcome. A separate three-pass evaluation covers all 30 validation scenarios with complete traces; requests for Gemini 3.7 Flash were routed to effective Gemini 3.8 Flash, as labeled in \Cref{tab:rl_cooksim_results}. Means and sample SDs reflect rollout variation on fixed scenarios. The two evaluations differ in rollouts, effective user model, and interaction-quality judging, so their absolute scores are not directly comparable.

The original interaction-quality scores use Gemini judgments on matched subsets of 95/360 training and 9/90 validation episodes per checkpoint (104/450 overall); validation spans only two passes, so none estimate full-split quality. The complete-trace evaluation judges all validation episodes with Gemini and GPT. Original detection recall is an audit-based proxy that counts a valid flag or prevention; available native reports agree with it, but full equivalence cannot be verified. Complete-trace recall comes from a post hoc rescore of saved reports under the current upstream rule, which accepts a correctly named error within its window or up to 30 ticks after commitment. The two recall definitions differ, and changing error exposure across rollouts makes both rates descriptive.

\begin{table}[!ht]
\centering\small
\setlength{\tabcolsep}{3pt}\renewcommand{\arraystretch}{1.12}
\begin{tabular*}{\linewidth}{@{\extracolsep{\fill}}llllrrr@{}}
\toprule
Assistant & User & Split & Checkpoint & \shortstack{In-time task\\success} & \shortstack{Detection\\recall} & \shortstack{Interaction\\quality}\\
\midrule
\multirow{6}{*}{Qwen3.5-4B} & \multirow{6}{*}{\shortstack[l]{Gemini 3.7\\Flash}}
& \multirow{2}{*}{Train} & Base & $32.2 \pm 4.3$ & $\mathbf{49.1} \pm 7.0$ & $23.5 \pm 0.3$\\
& & & + RL & $\mathbf{42.2} \pm 2.9$ & $43.8 \pm 6.8$ & $\mathbf{32.5} \pm 1.9$\\
& & \multirow{2}{*}{Validation} & Base & $33.3 \pm 3.3$ & $\mathbf{49.0} \pm 11.9$ & $29.6 \pm 4.8$\\
& & & + RL & $\mathbf{42.2} \pm 6.9$ & $34.9 \pm 6.7$ & $\mathbf{31.1} \pm 0.4$\\
& & \multirow{2}{*}{Overall} & Base & $32.4 \pm 4.1$ & $\mathbf{49.0} \pm 4.4$ & $23.7 \pm 0.5$\\
& & & + RL & $\mathbf{42.2} \pm 2.5$ & $42.1 \pm 4.9$ & $\mathbf{32.5} \pm 1.8$\\
\bottomrule
\end{tabular*}
\caption{\textbf{CookSim RL results by split.} Scores are percentages, reported as mean $\pm$ SD. + RL denotes 12 updates; bold marks higher means within pairs. Success and detection SDs are sample SDs over three passes. Gemini-judged interaction quality uses a matched subset, with population SD over three passes for train and overall and two for validation.}
\label{tab:rl_cooksim_splits}
\end{table}

\FloatBarrier
\paragraph{ScreenSim.}
Qwen3.5-4B receives no task manual during training or evaluation; Gemini 3.7 Flash simulates baseline and classic-novice users. The task-disjoint catalog has 136 training, 32 development, and 30 frozen-test scenarios. Each update uses 48 episodes; each split is evaluated in three passes. Strict in-time task success requires the requested phone state by the deadline without lasting unrequested changes.

\begin{table}[!ht]
\centering\small
\setlength{\tabcolsep}{3pt}\renewcommand{\arraystretch}{1.12}
\begin{tabular*}{\linewidth}{@{\extracolsep{\fill}}llllrrr@{}}
\toprule
Assistant & User & Split & Checkpoint & \shortstack{Strict in-time\\task success} & \shortstack{Native detection\\recall} & \shortstack{Interaction\\quality}\\
\midrule
\multirow{8}{*}{Qwen3.5-4B} & \multirow{4}{*}{Baseline}
& \multirow{2}{*}{Development} & Base & $24.0 \pm 11.8$ & $91.2 \pm 1.1$ & $\mathbf{30.0} \pm 0.6$\\
& & & + RL & $\mathbf{27.1} \pm 9.5$ & $\mathbf{94.9} \pm 4.7$ & $29.7 \pm 2.9$\\
& & \multirow{2}{*}{Frozen test} & Base & $\mathbf{27.8} \pm 6.9$ & $85.4 \pm 3.2$ & $\mathbf{36.4} \pm 1.7$\\
& & & + RL & $25.6 \pm 1.9$ & $\mathbf{88.7} \pm 6.1$ & $31.4 \pm 1.3$\\
\addlinespace[3pt]
& \multirow{4}{*}{Classic novice}
& \multirow{2}{*}{Development} & Base & $2.1 \pm 1.8$ & \textemdash & $\mathbf{27.0} \pm 3.2$\\
& & & + RL & $\mathbf{4.2} \pm 1.8$ & \textemdash & $22.1 \pm 1.5$\\
& & \multirow{2}{*}{Frozen test} & Base & $3.3 \pm 3.3$ & \textemdash & $\mathbf{27.9} \pm 3.1$\\
& & & + RL & $\mathbf{4.4} \pm 1.9$ & \textemdash & $26.9 \pm 0.6$\\
\bottomrule
\end{tabular*}
\caption{\textbf{ScreenSim no-manual RL results.} Scores are percentages, reported as mean $\pm$ SD over three passes. + RL denotes 12 updates; bold marks the higher mean within each user, split, and metric. Success and recall SDs are sample SDs; interaction-quality SDs are population SDs. Interaction quality uses the primary GPT-5.4-mini judge; dashes indicate unavailable recall.}
\label{tab:rl_screensim_no_manual}
\end{table}

Only classic-novice frozen-test success rises slightly; GPT-judged interaction quality falls for both users. Baseline native detection recall rises with more false flags, but it combines user calls with engine flags; classic-novice recall is unavailable. These fixed-scenario changes do not establish a reliable frozen-test improvement.
\FloatBarrier

\section{Task Success and Error Detection}
\label{app:evaluation_metrics}

We use the simulator's execution records to determine task completion and whether an assistant detects or prevents scheduled errors. These metrics complement response-level interaction quality by measuring the effect of assistance on task execution.

\subsection{In-time success}
For episode $i$, let $T_i$ be its duration and $T_i^{\mathrm{ref}}$ the duration of an engine-generated reference rollout that completes the task without errors. CookSim and VHSim define the completion budget and success indicator as
\begin{equation}
D_i=2(T_i^{\mathrm{ref}}+\alpha_i),\qquad
S_i=\mathbf{1}[G_i(s_i^{\mathrm{final}})=1\ \land\ T_i\leq D_i],
\label{eq:completion_deadline}
\end{equation}
where $G_i$ checks the task goal against the final world state and $\alpha_i$ allows additional time for scheduled events. CookSim uses case-specific allowances, while VHSim uses 20 ticks. ScreenSim uses $D_i=2T_i^{\mathrm{ref}}+8$. Reference durations are obtained from the corresponding error-free executions.

CookSim checks the served dish's ingredients, preparation states, and assembly order. VHSim checks the required object states and relations. ScreenSim checks the requested settings while requiring unrelated settings to remain unchanged; temporary search text is excluded. An episode succeeds only if it satisfies both the goal and the completion budget. We report the percentage of successful episodes in each run and average these rates across runs.

\subsection{Detection and prevention}
Detection recall measures the proportion of triggered errors correctly identified within their eligible detection windows. Error prevention measures the proportion stopped before the erroneous action becomes irreversible:
\begin{equation}
R=\frac{\sum_i N_i^{\mathrm{detected}}}{\sum_i N_i^{\mathrm{triggered}}},
\qquad
P_{\mathrm{prevent}}=\frac{\sum_i N_i^{\mathrm{prevented}}}
{\sum_i N_i^{\mathrm{triggered}}}.
\end{equation}
Here, $N_i^{\mathrm{triggered}}$, $N_i^{\mathrm{detected}}$, and $N_i^{\mathrm{prevented}}$ count the corresponding errors in episode $i$. We credit identification when the assistant raises an event flag accepted by the engine's event-specific grader or when the simulated user recognizes a correct diagnosis, as described in \Cref{sec:evaluation_metrics}. Prevention additionally requires the user to accept the guidance and change course before the point of no return. Events whose trigger conditions are never reached are excluded from these rates.

\section{Temporal Questions and Proactive Requests}
\label{app:questions}

We evaluate temporal question answering and proactive assistance during ongoing task execution. This section describes how CookSim generates these events, determines when to introduce them, and grades responses against the resulting interaction.

\subsection{Question generation and scheduling}
We instantiate templates for temporal order, action counting, duration, and event recall, adapting temporal-understanding tasks from streaming-video benchmarks~\citep{Niu_2025_CVPR,Huang_2025_CVPR,streamingbench2024}. Each question refers to the current episode, so its answer depends on the user's actual actions, including mistakes and recovery.

We introduce questions at specified task-progress checkpoints when their required evidence is available. Temporal-order questions require both referenced events to have occurred; elapsed-time questions require an item to be cooking; proactive requests require an item that is not yet ready. Questions about whether an event occurred can also test the absence of that event. We separate questions from conflicting scheduled errors and stop introducing them once the dish is complete or serving has been queued. A question remains unasked if the episode ends before its trigger condition is reached.

\subsection{Reference facts and scoring}
We derive reference answers from recorded world-state transitions, including ingredient preparation, cooking, assembly, serving, and station visits. Only completed actions contribute to these records. For example, a counting question uses the number of actual visits, while an order question compares the completion times of two events. A semantic judge compares the assistant's answer with these reference facts, allowing different valid phrasings. \Cref{tab:qa_families} summarizes the question types and their grading evidence.

\begin{table}[!t]
\centering\small
\setlength{\tabcolsep}{4pt}\renewcommand{\arraystretch}{1.16}
\begin{tabularx}{\linewidth}{@{}L{.19\linewidth}L{.30\linewidth}X@{}}
\toprule
Reported family & Example form & Grading evidence\\
\midrule
Temporal order & Did I prepare $A$ before $B$? & Completion times of both events; the stated relation must agree.\\
Action count & How many times have I visited $X$? & Number of realized visits; the integer must match exactly.\\
Duration & How long has $X$ been cooking? & Loading, ignition, and completion times; numeric duration answers allow five ticks of tolerance. The family also includes longest-on-heat and past-completion queries.\\
Event recall & Did I already do $A$? Does $X$ exist? & Recorded event occurrence or world inventory; the answer must reject an incorrect premise.\\
Proactive request & Tell me when $X$ is ready. & A correct readiness notification within the state-derived window, not merely an acknowledgement of the request.\\
\bottomrule
\end{tabularx}
\caption{\textbf{Question and request families in the main evaluation.} Concrete items, stations, and actions are bound from the current episode.}
\label{tab:qa_families}
\end{table}

We assign 1 to correct answers, 0.5 to partially correct answers, and 0 to incorrect, evasive, or absent answers. Each category reports the mean score of its graded questions within a run. The overall score pools graded questions and proactive requests across all five categories within that run. We then average the three run-level scores with equal weight. \Cref{tab:qa_proactive} reports the measured-delay CookSim runs. Questions not reached during the rollout are excluded.

\subsection{Proactive-request window}
Proactive requests test whether the assistant remembers to notify the user when a future condition is met. Let $t_q$ be the request time and $t_c$ the time at which the referenced ingredient finishes cooking. We require a readiness notification within $[L,H)$, where $L=\max(t_q,t_c)$ and $H$ is the end of the eligible notification interval, determined from the ingredient's recorded cooking state. Acknowledging the request alone does not satisfy it.

\begin{equation}
A(q)=\mathbf{1}[\text{valid notification in }[L,H)]\,\mathbf{1}[\text{no premature alert}].
\label{eq:alert_score}
\end{equation}
A readiness claim more than two ticks before completion is premature and receives zero, even if followed by a correct alert. Late and absent alerts also receive zero. We exclude requests with no valid cooking window, including cases where the user removes the item within two ticks of completion before an eligible alert can be delivered. We report the percentage of graded requests receiving a valid notification.

\section{Human Study: Protocol and Behavioral Comparisons}
\label{app:human_study}

We conduct a study with 45 participants to examine whether our interaction-quality metric reflects human preferences and whether simulated users reproduce human behavior. This section provides the protocol and the measures used for the comparisons in \Cref{sec:human_study}.

\subsection{Study protocol}
Participants interact with a browser-based CookSim application on their own computers, controlling the cook with a keyboard and communicating with the assistant through text. The assistant observes top-down and egocentric frames and responds in real time. Errors arise naturally from participants' actions. Participants may continue beyond the benchmark's completion deadline.

Participants first complete a tutorial explaining the game rules. Before playing the study episodes, they answer a questionnaire about their preferred execution speed and interruption frequency, corresponding to the \emph{execution speed} and \emph{interruption preference} axes of our simulated users. Each preference has three levels, ranging from cautious to fast execution and from step-by-step guidance to interruption only when necessary. We provide these preferences to the assistant before interaction.

Participants then complete four unassisted episodes, followed by nine assisted episodes over three rounds with assistants anonymously labeled A, B, and C. We rotate assistant--recipe assignments using a $3\times3$ Latin square so that each assistant encounters each recipe once. Participants rank the interaction experience after each round and may explain their preferences. Unlike the two questionnaire-based axes, \emph{skill} and \emph{trust in AI} are assessed from recorded behavior after participants finish playing: we use changes between the first and last unassisted episodes as a proxy for increased skill, and an LLM-assisted audit of the conversations to assign trust groups.

\paragraph{Assistants.}
The study includes Gemini 3.7 Flash, Gemini 3.1 Pro, Gemini 3 Flash, GPT-5.5, Gemini 3.1 Flash Live, GPT-Realtime 2.1, and GPT-5.4 mini. These are the model versions available during collection.

\subsection{Preference agreement}
We test whether higher interaction-quality scores correspond to assistants that participants prefer. We use the five interaction-quality dimensions in \Cref{app:rubric}. For each assistant response, each of two judges assigns a score with the factual-grounding gate applied. We average the two judges' scores, then average over assistant responses to obtain an episode-level score. For each participant and assistant, we average these episode scores over episodes containing assistant responses.

We derive each participant's preference between two assistants from the number of rounds in which one is ranked above the other. Pairs with tied preferences are excluded. We then compare the preferred assistant with the one receiving the higher mean interaction-quality score.

Each comparison consists of one participant $p$ and two assistants $a$ and $b$. Let $\mathcal P$ be the set of comparisons with a non-tied human preference and available scores for both assistants; $|\mathcal P|$ is the number of such comparisons. We set $h_{pab}=1$ if participant $p$ prefers $a$ and $h_{pab}=-1$ if they prefer $b$. The scores $\bar Q_{pa}$ and $\bar Q_{pb}$ are the participant-specific mean interaction-quality scores defined above. We compute
\begin{equation}
\mathrm{Agreement}=
\frac{1}{|\mathcal P|}
\sum_{(p,a,b)\in\mathcal P}
\left(
\mathbf{1}[h_{pab}(\bar Q_{pa}-\bar Q_{pb})>0]
+\frac12\mathbf{1}[\bar Q_{pa}=\bar Q_{pb}]
\right).
\end{equation}
Here, $\mathbf{1}[\cdot]$ equals 1 when its condition holds and 0 otherwise. The product $h_{pab}(\bar Q_{pa}-\bar Q_{pb})$ is positive when the metric assigns the higher score to the assistant preferred by the participant. Each comparison therefore receives 1 for agreement, 0 for disagreement, and $1/2$ when the metric scores are tied. Averaging these credits gives \textbf{78.5\%} agreement, supporting the use of interaction quality to measure assistance that reflects human preferences.

We also express this agreement using Kendall's rank concordance~\citep{kendall1938rank}, pooling the same participant--assistant-pair comparisons:
\begin{equation}
\tau=\frac{C-D}{|\mathcal P|}=2\,\mathrm{Agreement}-1\approx +0.57,
\end{equation}
where $C$ counts comparisons in which the metric and participant agree on the preferred assistant, and $D$ counts disagreements. Metric ties remain in $|\mathcal P|$ but contribute zero to $C-D$. We use this pair-count normalization rather than the tie-adjusted $\tau_b$. A value of $+1$ indicates complete agreement, $-1$ complete reversal, and $0$ equal numbers of agreements and disagreements.
\subsection{Behavioral measures and similarity}

We compare communication behavior, persona-dependent variation, and action errors to assess user-simulation fidelity. Following statistical evaluation of user simulators~\citep{schatzmann2005quantitative}, we measure how often specified behaviors occur in human and simulated episodes. We identify actions and their timing from execution records, and conversational behaviors from text patterns or response labels. An LLM-assisted audit assigns human trust groups and annotates procedural errors. \Cref{tab:behavior_definitions} defines each measure.

\begin{table}[!t]
\centering\small
\setlength{\tabcolsep}{4pt}\renewcommand{\arraystretch}{1.15}
\begin{tabularx}{\linewidth}{@{}L{.25\linewidth}X@{}}
\toprule
Measure & Episode-level statistic\\
\midrule
Utterance length & Mean number of words per nonempty user utterance.\\
Speech rate & Spontaneous user utterances per 100 ticks, with the time normalization described below.\\
Responsiveness & Fraction of assistant utterances followed by a user reply within ten ticks.\\
Suggestion uptake & Fraction of assistant utterances naming an ingredient followed by pickup or scooping of that ingredient within 15 ticks.\\
Disagreement & Fraction of user utterances that contest an assistant claim.\\
Guidance requests & Fraction of user utterances asking for the next action or help; text patterns are used for humans and response labels for simulated users.\\
Interruption ignoring & Assistant utterances followed by neither a reply within ten ticks nor pickup of a named item within 15 ticks, plus explicit requests for less interruption, divided by assistant utterances.\\
Action efficiency & $\mathbf{1}[\mathrm{task\ completed}]\min(1,T^{\mathrm{ref}}/T)$: zero for an unfinished task; otherwise, the ratio of the engine-generated error-free reference duration $T^{\mathrm{ref}}$ to the observed duration $T$, capped at 1. Both durations use the same time units.\\
Parallelism & Mean number of cooking stations simultaneously in progress.\\
\bottomrule
\end{tabularx}
\caption{\textbf{Behavioral measures for human--simulation comparisons.} Episode means exclude cases without an eligible observation. Interruption ignoring combines absent responses and explicit requests for fewer interruptions; the two counts may overlap.}
\label{tab:behavior_definitions}
\end{table}

We first compute the episode-level statistics defined in \Cref{tab:behavior_definitions}. For example, for responsiveness, each assistant utterance is an opportunity for the user to reply: it receives 1 if a user reply follows within ten ticks and 0 otherwise. The episode's responsiveness is the mean of these labels. Utterance length instead averages the number of words in user utterances, while speech rate counts utterances per 100 ticks. For each communication measure, we average its episode-level statistic separately over human and simulated episodes. These two averages are $x$ and $y$, respectively; thus, for utterance length, $x=5.87$ and $y=7.38$ words per utterance.

For persona contrasts, we compare changes between groups rather than their absolute behavioral rates. We use $x$ for the difference between two human group means and $y$ for the corresponding difference between simulated-user group means, following the directions specified in \Cref{tab:human_behavior_full}. For example, the trust contrast compares how suggestion uptake changes from high to low trust in each population. For speech-act distributions, $x$ and $y$ are vectors whose entries are the proportions of utterances assigned to each conversational category. In all three comparisons, similarity is
\begin{equation}
\operatorname{Sim}(x,y)=1-\frac{\|x-y\|_1}{\|x\|_1+\|y\|_1},
\qquad \operatorname{Sim}(0,0)=1.
\label{eq:fidelity_sim}
\end{equation}
For a scalar, $\|x\|_1=|x|$; for a vector, $\|x\|_1$ is the sum of its entries' absolute values. The numerator measures the absolute difference between the human and simulated statistics, and the denominator normalizes it by their combined magnitude. Similarity ranges from 0 to 1, with 1 indicating identical statistics; we also assign 1 when both are zero. For persona contrasts, this compares both the direction and magnitude of the behavioral change. For speech-act distributions, it measures the probability mass shared across categories.

\noindent\textbf{Speech-rate normalization.}
The reported speech rates are 0.39 and 16.10 utterances per 100 ticks for human and simulated users, respectively. Human rates use total elapsed time, while simulated rates exclude intervals during which scheduled events control the action queue. Using total elapsed time for both gives 0.39 and 11.70. Simulated users therefore speak substantially more often under either normalization.

\subsection{Communication and persona contrasts}
\Cref{tab:human_behavior_full} reports the human and simulated measurements and their similarities. Utterance lengths are similar, but simulated users speak and respond much more frequently. This difference suggests that the harness overemphasizes conversation relative to the largely silent task execution observed in human participants.
\begin{table}[!t]
\centering\small
\setlength{\tabcolsep}{3pt}\renewcommand{\arraystretch}{1.13}
\begin{tabularx}{\linewidth}{@{}Xrrr@{}}
\toprule
Measure & Human & Simulated & Similarity\\
\midrule
Words per utterance & 5.87 & 7.38 & 0.89\\
Utterances per 100 ticks & 0.39 & 16.10 & 0.05\\
Replies within ten ticks (\%) & 2.0 & 79.0 & 0.05\\
\midrule
Skill: task success (\%) & 50.0 $\to$ 79.3 & 25.0 $\to$ 100.0 & 0.56\\
Trust: suggestion uptake (\%) & 32.3 $\to$ 24.3 & 32.9 $\to$ 20.1 & 0.77\\
Trust: disagreement (\%) & 8.7 $\to$ 30.1 & 9.3 $\to$ 36.7 & 0.88\\
Interruption: guidance requests (\%) & 47.9 $\to$ 20.8 & 44.1 $\to$ 17.9 & 0.98\\
Interruption: ignoring index & 0.441 $\to$ 0.550 & 0.125 $\to$ 0.284 & 0.82\\
Speed: action efficiency (\%) & 73.7 $\to$ 76.0 & 9.9 $\to$ 11.5 & 0.82\\
Speed: parallelism & 0.214 $\to$ 0.213 & 0.129 $\to$ 0.180 & 0.00\\
\bottomrule
\end{tabularx}
\caption{\textbf{Human and simulated behavior.} Arrows show group means in the following directions: first to last play / novice to expert (skill), high to low trust, step-by-step to minimal interruption, and cautious to faster execution. Similarity compares the unrounded differences between groups. Speech-rate normalization is described in \Cref{app:human_study}.}
\label{tab:human_behavior_full}
\end{table}

We further apply the same rule-based speech-act classifier to 154 human and 63,512 simulated utterances (\Cref{tab:human_speech_acts}). Simulated users frequently describe their next action, whereas human participants more often ask questions or correct the assistant. These categories approximate conversational intent through lexical patterns.

\begin{table}[!t]
\centering\small
\setlength{\tabcolsep}{3pt}\renewcommand{\arraystretch}{1.13}
\begin{tabularx}{\linewidth}{@{}Xrr@{}}
\toprule
Speech act & Human (\%) & Simulated (\%)\\
\midrule
Greeting & 10.4 & 0.0\\
Comment on the assistant or its silence & 2.6 & 0.1\\
Expression of frustration or hostility & 3.2 & 0.1\\
Request for a future notification & 5.2 & 0.0\\
Strategy or planning question & 6.5 & 0.0\\
Question about location or next action & 22.1 & 15.9\\
Correction of the assistant & 21.4 & 17.1\\
Acknowledgement & 1.3 & 0.8\\
Description of the user's next action & 1.3 & 23.8\\
Other & 26.0 & 42.1\\
\bottomrule
\end{tabularx}
\caption{\textbf{Distribution of conversational behaviors.} We apply the same rule-based classifier to 154 human and 63,512 simulated utterances. The resulting distributions have similarity 0.61 under \Cref{eq:fidelity_sim}.}
\label{tab:human_speech_acts}
\end{table}

We assess persona fidelity through behavioral contrasts between user groups~\citep{aher2023turing}. For \emph{skill}, we compare participants' first and last unassisted episodes with novice and expert simulated users. For \emph{trust in AI}, we compare high- and low-trust human groups assigned by the conversation audit with the corresponding simulated profiles. For \emph{interruption preference}, we compare users requesting step-by-step guidance with those preferring interruption only when necessary. For \emph{execution speed}, we compare cautious users (questionnaire level 1) with those preferring faster execution (levels 2--3). The latter two comparisons use participants' pre-play questionnaire answers and the corresponding settings of simulated users.

\subsection{Natural action errors}
We examine whether scheduled errors cover the procedural mistakes that arise during human task execution. An LLM-assisted audit of 55 unassisted episodes identifies 112 errors spanning eight of the ten scheduled types (\Cref{tab:human_error_coverage}). Incorrect preparation, unattended cooking, forgotten steps, and wrong assembly order are the most frequent. The simulator additionally supports extraneous and confused ingredients during assembly, which do not occur in this sample.

The audit also identifies 51 behaviors outside the scheduled taxonomy: 30 interface-control errors, nine instances of loading a station without turning it on, and 12 other lapses or inefficient actions. The pie chart in \Cref{fig:human_comparison} shows the distribution of the 112 mapped errors. Our schedules test a range of challenging mistakes rather than matching their natural occurrence frequencies.

\begin{table}[!t]
\centering\small
\setlength{\tabcolsep}{3pt}\renewcommand{\arraystretch}{1.13}
\begin{tabularx}{\linewidth}{@{}Xrr@{}}
\toprule
Error type & Annotations & Episodes\\
\midrule
Incorrect preparation & 42 & 31\\
Unattended cooking & 20 & 16\\
Forgotten next step & 19 & 19\\
Wrong assembly order & 13 & 13\\
Serving raw food & 8 & 7\\
Repeated step & 5 & 5\\
Omitted subgoal & 4 & 3\\
Premature serving & 1 & 1\\
Extraneous ingredient & 0 & 0\\
Confused ingredient at assembly & 0 & 0\\
\midrule
Outside the scheduled vocabulary & 51 & 36\\
\bottomrule
\end{tabularx}
\caption{\textbf{Action errors in human task execution.} An LLM-assisted audit identifies procedural errors in 55 unassisted episodes. An episode may contain multiple error types. The final row reports behaviors outside the scheduled-error taxonomy.}
\label{tab:human_error_coverage}
\end{table}

\section{Environment and Benchmark Construction}
\label{app:environments}

We construct IVG-Bench by combining everyday tasks with scheduled out-of-plan events and diverse user personas. This section details the benchmark composition and the environment-specific state, action, and event representations.

\subsection{Benchmark composition}
IVG-Bench contains 105 task cases, each paired with three personas to form 315 interaction episodes per assistant and run (\Cref{tab:benchmark_composition}). Each case specifies an initial environment, a task goal, and scheduled events. The user and assistant determine the subsequent actions and dialogue through closed-loop interaction.

\begin{table}[!t]
\centering\small
\setlength{\tabcolsep}{4pt}\renewcommand{\arraystretch}{1.13}
\begin{tabularx}{\linewidth}{@{}lXrr@{}}
\toprule
Environment & Evaluated case construction & Cases & Episodes/run\\
\midrule
CookSim & Five recipes, two kitchen layouts, five schedules per recipe--layout pair; three personas per case. & 50 & 150\\
VHSim & Household tasks with multiple progress-triggered deviations; three assigned personas per case. & 25 & 75\\
ScreenSim & Multi-step phone configurations with two scheduled mistakes; three assigned personas per case. & 30 & 90\\
\bottomrule
\end{tabularx}
\caption{\textbf{IVG-Bench benchmark composition.} Each case is paired with three personas. VHSim and ScreenSim select these personas from nine available profiles.}
\label{tab:benchmark_composition}
\end{table}

\paragraph{Scheduled-event statistics.}
We schedule an average of 3.53 events per episode, for a total of 1,113 across the benchmark. \Cref{tab:event_schedule_stats} summarizes their types and distribution across the 105 cases before persona assignment. Events occur when their task-progress conditions are reached, so the number encountered in a rollout depends on the interaction and its termination time.

\begin{table}[!t]
\centering\small
\setlength{\tabcolsep}{5pt}\renewcommand{\arraystretch}{1.12}
\begin{tabular}{@{}lrrrrrl@{}}
\toprule
Environment & Cases & Errors & Questions & Requests & Plan changes & Events/case, mean [range]\\
\midrule
CookSim & 50 & 140 & 88 & 29 & 3 & 5.20 [2--7]\\
VHSim & 25 & 51 & 0 & 0 & 0 & 2.04 [2--3]\\
ScreenSim & 30 & 60 & 0 & 0 & 0 & 2.00 [2--2]\\
\midrule
Total & 105 & 251 & 88 & 29 & 3 & 3.53 [2--7]\\
\bottomrule
\end{tabular}
\caption{\textbf{Scheduled-event distribution.} Counts are per case before persona assignment. Requests denotes proactive requests. Temporal questions, proactive requests, and plan changes are evaluated in CookSim. The number of events per case is 2, 3, 4, 5, 6, or 7 in 55, 12, 1, 11, 16, and 10 cases, respectively.}
\label{tab:event_schedule_stats}
\end{table}

\subsection{Tasks and environment state}
\noindent\textbf{CookSim.}
We implement cooking dynamics in Python and render top-down and egocentric 3D observations using Three.js in Chromium. The world state records the user's position and held object, station contents, ingredient preparation, cooking progress, and plating order. Cooking continues while the user navigates or prepares other ingredients. We evaluate five recipes: Caprese, Cheeseburger, Fish and Chips, Steak Frites, and Loaded Burrito. These require two to five plate components, with more complex recipes introducing preparation dependencies and repeated portions.

For example, the Loaded Burrito requires tortilla, cooked rice, cooked steak, chopped tomato, and chopped lettuce in that order. The user can cook rice while preparing steak and vegetables, allowing different valid preparation sequences. Each recipe--layout pair has five event schedules containing two to four errors, with questions, proactive requests, or plan changes where applicable.

\noindent\textbf{VHSim.}
We adapt VirtualHome~\citep{virtualhome2018}, using its Unity renderer and indoor object interactions. The world graph contains object attributes and relations such as containment, support, and room membership. Tasks include gathering dishes into a dishwasher, returning belongings to designated locations, and preparing or clearing household spaces. A task is checked through goal predicates, not by requiring one reference route.

We render semantic actions as egocentric motion sequences and reveal them incrementally to the assistant. If guidance interrupts an action, execution resumes from the corresponding intermediate pose. A top-down map records the user's trajectory. We validate event schedules to ensure that their trigger conditions remain reachable, the targeted errors are visually distinguishable, and intervention windows are separated.

\noindent\textbf{ScreenSim.}
We implement an iPhone-style Settings interface in HTML/CSS, maintain application state in Python, and capture $412\times892$ observations with Playwright and Chromium. Tasks derived from public iPhone documentation require coordinated changes across pages, such as restricting an app's permissions or improving display readability. The simulator tracks settings, navigation history, and irreversible changes. Success requires reaching the requested configuration while preserving unrelated settings.

\subsection{Semantic actions and grounding}
\Cref{tab:environments} pairs representative tasks with the semantic action classes available in each environment.
\begin{table}[!t]
\centering\footnotesize
\begingroup
\setlength{\tabcolsep}{3.5pt}
\renewcommand{\arraystretch}{1.08}
\newcommand{\ivgengine}[1]{\begin{minipage}[t]{10pt}\vspace{8pt}\rotatebox{90}{\textbf{#1}}\end{minipage}}
\newcommand{\ivgtask}[1]{{\scriptsize\ttfamily #1}}
\newcommand{\ivgtasks}[1]{%
  \begin{minipage}[t]{\linewidth}
  \setlength{\tabcolsep}{3pt}
  \begin{tabularx}{\linewidth}[t]{@{}L{60pt} >{\raggedright\arraybackslash}X@{}}
  #1
  \end{tabularx}
  \end{minipage}}
\begin{tabularx}{\linewidth}{@{}L{10pt} L{0.53\linewidth} >{\raggedright\arraybackslash}X@{}}
\toprule
 & \ivgtasks{\textbf{Task} & \textbf{Goal}} & \footnotesize\textbf{Action classes}\\
\midrule
\ivgengine{CookSim} & \ivgtasks{
\ivgtask{loaded\_\allowbreak burrito} & Cook rice and sliced steak; chop tomato and lettuce; plate in that order.\\[3pt]
\ivgtask{shrimp\_\allowbreak pasta} & Boil noodles, chop tomato, and sear shrimp; plate in that order.\\[3pt]
\ivgtask{steak\_\allowbreak frites} & Sear steak slices; fry two chip portions; plate steak first.
}
& \textit{Navigate:} \mbox{go to \texttt{<food source>}}; \mbox{go to \texttt{<cook station>}}; \mbox{go to \texttt{<prep station>}}; \mbox{go to \texttt{<counter>}}; \mbox{go to plates/pass/trash}.\par\smallskip
\textit{Manipulate:} pick up; put down; scoop; chop; turn on; trash; serve.\\
\midrule
\ivgengine{VHSim} & \ivgtasks{
\ivgtask{post\_\allowbreak meal\_\allowbreak cleanup} & Load dishes into the dishwasher; return the apple to the fridge; close both.\\[3pt]
\ivgtask{collect\_\allowbreak dishes} & Gather three plates and two glasses; load and close the dishwasher.\\[3pt]
\ivgtask{return\_\allowbreak belongings} & Return toilet paper to the bathroom counter, a book to the desk, and a pillow to the bed.
}
& \textit{Navigate:} turn left/right; go straight; look up/down; go through the visible door; turn to \mbox{\texttt{<obj>}}; go to \mbox{\texttt{<visible obj/place>}}.\par\smallskip
\textit{Manipulate:} pick up \mbox{\texttt{<obj>}}; put \mbox{\texttt{<obj>}} on/in \mbox{\texttt{<target>}}; open/close \mbox{\texttt{<obj>}}; switch on/off \mbox{\texttt{<obj>}}; drink \mbox{\texttt{<obj>}}; \mbox{sit on \texttt{<obj>}; stand up}.\\
\midrule
\ivgengine{ScreenSim} & \ivgtasks{
\ivgtask{app\_\allowbreak privacy} & Disable one app's location, tracking, and background refresh only.\\[3pt]
\ivgtask{readability} & Enlarge and bold text; clarify buttons; increase contrast; reduce motion.\\[3pt]
\ivgtask{lock\_\allowbreak screen\_\allowbreak security} & Block Control Center, Today View, Wallet, and missed-call return.
}
& \textit{Input:} \mbox{tap \texttt{<label>}}; \mbox{type \texttt{<text>}}.\par\smallskip
\textit{Navigate and adjust:} swipe up/down; swipe left/right on \mbox{\texttt{<label>}}; press back/home.\\
\bottomrule
\end{tabularx}
\endgroup

\caption{\textbf{Example tasks and semantic action spaces.} Angle brackets indicate object or place references resolved by the grounding module. Tasks illustrate the supported environments, including configurations beyond the evaluated subset.}
\label{tab:environments}
\end{table}

We translate semantic commands into executable operations by resolving their object and place references. In CookSim, navigation specifies an ingredient source or station, while commands such as \texttt{pick up}, \texttt{chop}, and \texttt{scoop} act at the station the user faces. VHSim resolves visible objects and places before executing navigation or manipulation. ScreenSim maps visible labels and relative references to interface gestures. The user receives execution feedback when a reference cannot be resolved or an action's prerequisites are unmet.

We track task progress using conditions on ingredients and stations in CookSim, objects and their relations in VHSim, and settings values in ScreenSim. These conditions identify completed subgoals independently of the particular action sequence used to reach them.

\subsection{Progress-conditioned events}
We specify each event through a task-progress trigger, its target objects, the required action or utterance, and a completion or cancellation condition. The schedule is hidden from the assistant. During an execution-error event, the harness temporarily controls the user's action queue while the user model continues the conversation. Accepted guidance can prevent the error before commitment or support recovery afterward. Control returns to the user when the event is resolved, completed, or reaches its maximum duration.

CookSim covers incorrect preparation, extra or confused ingredients, raw serving, wrong assembly order, omitted or repeated steps, premature serving, unattended cooking, and stalled progress. Questions and intended plan changes are scheduled separately. VHSim covers omissions, wrong objects or destinations, repetition, stalled progress, and navigation deviations. ScreenSim introduces incorrect setting selections or gestures while keeping the intended task goal fixed. Examples of these events and the resulting interactions appear in \Cref{app:qual_events}.

\section{Assistant Settings}
\label{app:assistance}

We detail the observation, memory, and response-timing settings used to evaluate visual interaction models in \Cref{sec:assistance_setting}. These settings allow each model to monitor an ongoing task while interacting with the simulated user.

\subsection{Observations and memory}
The assistant receives the task instruction, conversation history, user-stated preferences, and observations of the user's activity. The scheduled events, private user plan, action queue, and evaluation labels remain hidden. Streaming models retain context within their sessions. Non-streaming models receive recent observations together with a recurrent Plan--Act record that stores their estimates of task progress and pending requests. In CookSim and VHSim, real-time API assistants use the polled setting, with response requests every two ticks rather than model-initiated turn-taking. We describe the environment-specific inputs below.

\paragraph{CookSim.}
We query non-streaming assistants every two simulation ticks, provided the previous response has returned. Each query includes the eight most recent synchronized top-down and egocentric observations, the recipe, and public task rules. The structured world-state description provided to the simulated user is not included. The assistant updates its own progress estimates and announced goal changes in the recurrent memory.

\paragraph{VHSim.}
We provide eight recent egocentric observations, a top-down trajectory map, a current-room map, a room overview, and a brief description of the user's physical state. The spatial aids identify nearby objects and their states, as well as rooms with devices left on. This configuration therefore includes structured spatial information in addition to images. Recurrent memory retains task progress and important observations as the user moves between rooms.

\paragraph{ScreenSim.}
We provide the two most recent phone screenshots, the task, public preferences, conversation history, and recurrent notes, with public task documentation when enabled. The assistant does not receive the interface hierarchy, control coordinates, or hidden settings values used to execute gestures. Simulation advances through user gestures, with a fixed zero-tick response delay.

\subsection{Latency and asynchronous execution}
To model guidance that may arrive after the observed state has changed, CookSim and VHSim continue executing the user's queued actions while the assistant generates a response. We convert measured assistant latency to simulation ticks as described in \Cref{sec:harness}, with 1.2 seconds per tick and delay caps of 30 ticks in CookSim and 15 ticks in VHSim. Non-streaming requests are issued sequentially. User-model inference and rendering do not advance the simulation clock.

In the latency ablation, we compare fixed two-tick and measured assistant delay in CookSim and VHSim under identical task, persona, and event specifications. We generate a new interaction in each condition so that delayed guidance can affect user behavior and task outcomes. Locally hosted open-weight models in CookSim and VHSim are served on H100 GPUs. We access Inkling through the Tinker API with low reasoning effort. Claude Opus 5.5 is accessed through OpenRouter; its measured response time includes network and queue delays. \Cref{tab:latency_prevention} reports response times and error-prevention rates.

\begin{table}[!t]\centering\small
\setlength{\tabcolsep}{4pt}\renewcommand{\arraystretch}{1.1}
\begin{tabular*}{\linewidth}{@{\extracolsep{\fill}}lrrrrrr@{}}
\toprule
 & \multicolumn{3}{c}{CookSim} & \multicolumn{3}{c}{VHSim}\\
\cmidrule(lr){2-4}\cmidrule(lr){5-7}
Assistant & Delay (s) & Fixed & Measured & Delay (s) & Fixed & Measured\\\midrule
Gemini 4 Argon & 6.5 & 62.3 & 39.9 & 8.1 & 46.2 & 27.0\\
Claude Opus 5.5 & 6.4 & 67.0 & 43.6 & 8.1 & 33.2 & 21.4\\
Gemini 3.1 Pro & 13.2 & 57.1 & 19.3 & 11.7 & 59.0 & 32.3\\
Gemini 3.8 Flash & 5.1 & 48.6 & 31.2 & 6.1 & 37.9 & 21.5\\
GPT-6.1 Sol & 5.2 & 49.0 & 35.7 & 5.5 & 31.0 & 23.6\\
GPT-6 Astra & 6.6 & 48.5 & 36.8 & 8.2 & 33.1 & 15.4\\
\addlinespace[2pt]
GPT-Realtime 2.1 & 4.9 & 46.0 & 36.3 & 7.4 & 39.2 & 18.0\\
Gemini 3.8 Live & 2.3 & 27.9 & 23.9 & 5.4 & 14.8 & 8.3\\
\addlinespace[2pt]
Muse Glimmer & 1.5 & 45.0 & 45.5 & 2.2 & 5.2 & 6.0\\
Inkling & 9.9 & 51.7 & 12.4 & 10.0 & 48.1 & 23.2\\
Qwen3.8-27B & 2.4 & 43.6 & 39.7 & 4.0 & 19.3 & 20.5\\
InternVideo3-8B & 3.2 & 12.3 & 6.3 & 4.9 & 0.0 & 0.0\\
Kimi-VL-Thinking & 3.3 & 12.8 & 10.0 & 3.2 & 51.2 & 43.7\\
Molmo2-8B & 2.5 & 3.8 & 4.0 & 2.5 & 0.0 & 0.0\\
\addlinespace[2pt]
Qwen3-Omni-30B & 1.6 & 27.7 & 32.7 & 2.0 & 0.0 & 1.4\\
MiniCPM-o 4.5 & 1.4 & 20.4 & 18.1 & 2.2 & 4.9 & 4.3\\
StreamingVLM & 1.3 & 0.4 & 0.7 & 1.8 & 0.0 & 0.0\\
Proact-VL & 1.0 & 0.7 & 0.9 & 1.2 & 0.0 & 0.0\\
\addlinespace[2pt]
\bottomrule\end{tabular*}
\caption{\textbf{Assistant response time and error prevention.} Delay is the mean observed response time across all three measured-delay runs, before conversion to simulation ticks and capping. Fixed and measured columns report error prevention (\%) under each delay setting.}
\label{tab:latency_prevention}
\end{table}

\section{Interaction-Quality Rubric and Judging}
\label{app:rubric}

We assess whether an assistant's responses are correct and useful in the context of the ongoing task. Simulation provides the world-state and execution evidence needed to evaluate a response without requiring a reference utterance. We detail the rubric, judge inputs, and score aggregation below.

\subsection{Evaluation unit and evidence}
Each evaluation round contains an assistant response and the user's subsequent utterances and actions up to the next assistant response. The judge receives the task, public user preferences, relevant world state, conversation history, subsequent actions, and available visual observations. These records allow it to check both the response's claims and how its guidance relates to execution. User agreement does not establish correctness; factual claims are checked against the engine evidence.

The judge identifies factual claims, advised actions, and the response's conversational purpose, then assigns \emph{pass}, \emph{fail}, or \emph{not applicable} to each rubric item with an evidence-based explanation. Interaction quality grades spoken responses; unanswered questions and omitted notifications are measured by the question-answering and proactive-request metrics.

\subsection{Rubric dimensions}
We evaluate five dimensions: factual grounding, situational relevance, actionable guidance, user intent uptake, and guidance conciseness. Each dimension is assessed through the binary criteria in \Cref{tab:rubric_items}. These dimensions draw on factual and conversational correctness in dialogue evaluation~\citep{mehri-eskenazi-2020-unsupervised} and context-sensitive, user-controllable assistance in HCI~\citep{amershi2019guidelines}. We use evidence-supported item judgments following rubric-based model evaluation~\citep{liu-etal-2023-g,checkeval2025}. The same criteria apply across environments, with references to the corresponding objects, tasks, and actions.

\begin{table}[!t]
\centering\small
\setlength{\tabcolsep}{4pt}\renewcommand{\arraystretch}{1.14}
\begin{tabularx}{\linewidth}{@{}L{.20\linewidth}lX@{}}
\toprule
Category & Item & Pass criterion\\
\midrule
Factual grounding & T1 & Claims about the current state and the user's completed actions are true.\\
 & T2 & Referenced entities, places, controls, rules, and task requirements exist.\\
\addlinespace
Situational relevance & S1 & The response addresses the current action, subgoal, live concern, pending request, or announced intent.\\
 & S2 & A pending question, request, or objection is answered, accepted, rejected with a reason, or deferred for a stated urgent reason.\\
\addlinespace
Actionable guidance & H1 & The first advised action can start now, or its prerequisite is supplied.\\
 & H2 & Advice advances the current plan, answers the user's request, or repairs the error.\\
 & H3 & A mistake claim includes a remedy; an explicit instruction to stop suffices when stopping is the remedy.\\
\addlinespace
User intent uptake & L1 & A reply to push-back supplies evidence, a new reason, a correction, or an alternative rather than merely repeating the claim.\\
 & L2 & Valid announced intentions and requests for fewer interruptions are respected, unless a current task-critical reason is stated.\\
\addlinespace
Guidance conciseness & E1 & The response contains what is needed now, without unnecessary explanation, praise, or distant future steps, given the user's interruption preference.\\
\bottomrule
\end{tabularx}
\caption{\textbf{Interaction-quality rubric.} A category passes when all applicable items pass. Failure on either factual-grounding item gates the overall score. Items unrelated to the response's context are marked not applicable.}
\label{tab:rubric_items}
\end{table}

\subsection{Aggregation}
For judge $j$, conversational round $r$, and category $c$, let $I_{rjc}$ contain the applicable rubric items. The category passes only when every applicable item passes:
\begin{equation}
q_{rjc}=\prod_{i\in I_{rjc}}\mathbf{1}[i\text{ passes}],
\qquad I_{rjc}\ne\varnothing.
\end{equation}
Categories with no applicable items are omitted. Let $g_{rj}=0$ if either factual-grounding item fails, and $g_{rj}=1$ otherwise. We compute a separate gated score for each judge and average these scores for the round:
\begin{equation}
Q_{rj}=g_{rj}\frac{1}{|C_{rj}|}\sum_{c\in C_{rj}}q_{rjc},
\qquad Q_r=\frac{1}{|J_r|}\sum_{j\in J_r}Q_{rj},
\label{eq:interaction_quality_round}
\end{equation}
where $C_{rj}$ is the set of applicable categories for judge $j$ and $J_r$ contains judges with an applicable score for round $r$.

We average round scores within each model and run, then average the run-level scores with equal weight. For each category, we average the judges' applicable pass indicators within a round, then average over applicable rounds and runs, without the factual-grounding gate. The overall score can therefore be lower than the mean of the category scores: a response may provide relevant or concise guidance while making a factual error. Cross-environment plots give each environment equal weight.

\subsection{Judge configuration}
CookSim provides structured state evidence and an available rendered frame. VHSim supplies scene-graph state and execution records, while ScreenSim supplies a description of the visible interface and settings changes. These environment-specific inputs ground the judge's assessment in the state and actions recorded by the simulator.

We use GPT-5.4 mini and Gemini 3.7 Flash as independent judges for every assistant in all three environments. Both judges use the same five-category rubric. 

We apply consistency rules to CookSim judgments. A false factual claim causes T1 to fail. H1 evaluates whether the first advised action can be executed; later steps may depend on its completion. L2 applies only when a valid user intention or request for fewer interruptions is active. These rules align item scores with the judge's extracted claims and actions.

\paragraph{Prompt structure.}
The grading prompt presents the task and public preferences, time-aligned world state, recent conversation, the assistant response, subsequent user actions and replies, and the rubric. The judge returns a structured judgment with supporting evidence for each applicable item. The prompt instructs it to distinguish false claims from unverifiable ones and to assess response content rather than length.

\section{User Harness and Prompt Design}
\label{app:prompts}

We implement the simulated user as a VLM agent that plans and acts independently while communicating with the assistant. The harness provides task and persona instructions, maintains the action queue, and inserts scheduled events. We describe these components and the prompt structure below.

\subsection{User planning and communication}
We use Gemini 3.7 Flash as the user model in CookSim and ScreenSim, and Gemini 3.1 Pro in VHSim. It receives the task, persona, observations or structured state descriptions, its previous actions and execution feedback, conversation history, and private instructions for active scheduled events. The harness executes its action plan, while only its utterance is delivered to the assistant.

In CookSim, the user's response contains an utterance, a conversational response type, and an action-plan update. Response types include requesting an explanation, questioning guidance, requesting more guidance or fewer interruptions, accepting or rejecting advice, and answering a question. The user can retain, append to, or replace its remaining action queue. VHSim uses the same separation between planning, semantic actions, and dialogue. ScreenSim provides the user with a textual description of the visible interface for planning gestures, while the assistant observes screenshots.

Structured state information supports the simulated user's perception and action execution. The assistant instead relies on the observation inputs specified in \Cref{app:assistance} and influences execution through communication.

\subsection{Scheduled-event control}
To introduce an execution error at a specified task state, the harness temporarily controls the user's action queue and privately explains the event to the user model. The user continues the conversation according to its persona and decides whether to accept the assistant's guidance. Acceptance returns control of the queue to the user, enabling prevention before the point of no return or recovery afterward. Control also returns when the event completes, is cancelled, or reaches its maximum duration.

Questions, proactive requests, and intended plan changes are introduced through private instructions to produce the corresponding utterance. Questions refer to execution history; proactive requests ask the assistant to monitor a future condition; plan changes revise the user's intended goal. The assistant receives the resulting utterance without seeing the private event specification.

\subsection{Persona construction}
We condition user behavior on four three-level persona axes: \emph{skill}, \emph{trust}, \emph{speed preference}, and \emph{interruption preference}. Each level is expressed as a behavioral instruction (\Cref{tab:persona_statements}). Skill and trust remain private to the user model, while the two interaction preferences are provided to the assistant as questionnaire responses. These instructions shape planning and responses without prescribing a fixed conversation.

\begin{table}[!t]
\centering\small
\setlength{\tabcolsep}{4pt}\renewcommand{\arraystretch}{1.16}
\begin{tabularx}{\linewidth}{@{}L{.16\linewidth}XX@{}}
\toprule
Axis & Lower-level disposition & Higher-level disposition\\
\midrule
Skill & Broad goals are difficult to translate into a reliable sequence. & Maintains the workflow and reasons ahead about timing, tools, and dependencies.\\
Trust & Compares disputed claims with memory and visible evidence before changing course. & Takes plausible, situation-linked guidance seriously, while noticing clear contradictions.\\
Speed & Prefers methodical, cautious execution. & Values momentum, overlapping work, and avoiding unnecessary delay.\\
Interruption & Welcomes close, step-by-step guidance. & Finds non-urgent commentary intrusive and keeps replies brief while working.\\
\bottomrule
\end{tabularx}
\caption{\textbf{Persona instructions.} The two endpoint levels for each axis are summarized above; the middle level describes an intermediate preference or ability.}
\label{tab:persona_statements}
\end{table}

For example, the high-skill CookSim instruction states, \textquotedblleft You naturally think ahead about timing, dependencies, tools, and likely failure points.\textquotedblright\ The minimal-interruption instruction states, \textquotedblleft Extra commentary during active work can feel intrusive, especially if it is not urgent or directly useful.\textquotedblright\ These statements encourage independent planning and selective responses to guidance. We compare the resulting behavior with human participants in \Cref{app:human_study}.

VHSim expresses skill as familiarity with the house and object locations, while ScreenSim expresses it as knowledge of phone menus and settings. A persona combines levels across the four axes; comparisons between complete profiles can therefore reflect multiple attributes.

\subsection{Assistant prompt and recurrent memory}
We instruct the assistant to monitor task progress, respond to questions, respect announced intentions, and remain silent when guidance is unnecessary. It maintains estimates of completed and pending subgoals, together with unresolved requests. For non-streaming models, the previous structured response is included in the next query as recurrent memory. The assistant updates this record from its observations rather than receiving an engine-corrected plan.

The response separates the utterance from an optional error flag and private progress notes. For example, ScreenSim uses \texttt{text} for the utterance, \texttt{flag} for a detected error, \texttt{task\_progress} for progress estimates, \texttt{goal\_change} for revised intentions, and \texttt{note} for memory. Only the utterance is delivered to the user. Flags support evaluation; changing task execution still requires the user to act on the guidance.

\subsection{Generalized prompt structure}
The following blocks summarize the user and assistant prompts in presentation order. Bracketed fields are populated with environment rules, action definitions, scene information, or episode-specific values. We then provide the verbatim CookSim system-task injection used to introduce a belief-based error.

\begingroup
\newcommand{\promptblock}[2]{\par\smallskip\noindent\fcolorbox{black!25}{black!2}{\begin{minipage}{0.955\linewidth}\textbf{#1}\par\smallskip\small #2\end{minipage}}\par\smallskip}

\noindent\textbf{Simulated-user prompt.} The user plans and acts in the environment. Only its utterance, not its private task state or action queue, is delivered to the assistant.

\promptblock{1. Role and task}{Role-play a human carrying out \texttt{[task goal]} while an AI assistant observes and offers guidance. Remain in character, act independently, and respond to the assistant according to \texttt{[persona]}. Do not expose the test setup.}

\promptblock{2. Persona and environment}{\texttt{[Private skill and trust dispositions; publicly shared speed and interruption preferences]}\\
\texttt{[Task requirements and environment rules]}\\
\texttt{[Semantic action vocabulary and grounding instructions]}\\
\texttt{[Optional planning or grounding tools available to this persona]}}

\promptblock{3. Current private state}{\texttt{[User-facing state map or screen description; body position and held object]}\\
\texttt{[Current action and remaining action queue]}\\
\texttt{[Own past actions and execution feedback]}\\
\texttt{[Private scratch results from planning tools]}\\
\texttt{[Public conversation history]}}

\promptblock{4. Action and speech response}{Return a structured response with \texttt{utterance} (possibly empty), a conversational \texttt{move}, and a \texttt{plan\_update} that retains, appends to, or replaces the remaining action queue with exact executable commands. The user may question, accept, or reject advice. Speaking and changing the action queue are separate decisions; when actions proceed normally, remain silent. The harness executes the queue and returns the resulting feedback on the next turn.}

\noindent\textbf{Visual interaction model prompt.} The assistant monitors the user but has no action channel to the environment.

\promptblock{1. Role and available information}{Guide the user toward \texttt{[task goal]} through text. Receive \texttt{[sampled visual observations and environment-specific observation aids]}, the current timestamp, the public conversation, \texttt{[publicly shared preferences]}, and the assistant's own previous structured output as memory. Do not receive the user's private plan or scheduled events.}

\promptblock{2. Task model and monitoring}{Use \texttt{[task rules and action vocabulary]} to maintain a progress record with completed, current, and future subgoals and their constraints. On the first turn, initialize the expected subgoals. On later turns, update the record from observed evidence and retain uncertainty when the visual state is ambiguous.}

\promptblock{3. Intervention policy and output}{Watch for \texttt{[error taxonomy]}, answer user questions, honor announced goal changes, and otherwise remain silent when guidance is unnecessary. Return \texttt{[progress record]}, \texttt{goal\_change}, spoken \texttt{text} (empty for silence), and an optional \texttt{flag} for a detected error. Only \texttt{text} reaches the user; a flag alone does not affect task execution.}

\noindent\textbf{System-task injection.} When an event is triggered, the harness adds a private instruction to the user's prompt. The block below is the complete CookSim instruction for an error that the user initially believes is correct. The field \texttt{\{mistake\_line\}} describes the specific mistake. This instruction coordinates the user's conversation with the action queue controlled by the harness; it is hidden from the assistant.

\promptblock{CookSim belief-error \texttt{SYSTEM TASK} (verbatim wording)}{\texttt{== [SYSTEM TASK -- meta-instruction, overrides everything below] ==}\par
You want to test the AI assistant's ability to identify an error and convince you it truly happened. Your BODY is already committed: the action queue in your body status IS what you are doing---you do not need to issue these actions, and you cannot change them from here. Your job is your VOICE: arrange your conversation around your CURRENT ACTION or say nothing, speaking about it in character as your own choice. Never promise an action that is not in your queue, and never deny one that is.\par
\texttt{\{mistake\_line\}}\par
Keep it out of your words: do not announce it, name it or hint at it. The assistant has to SEE it, not hear it from you. You know what mistake you are making, so you can say whether the assistant has found it---that is all \texttt{assistant\_error\_call} records.\par
While it has not found it, carry on and answer as your persona would. Once it has, your trust level says how fast you come around: HIGH trust---the first correct warning convinces you; LOW trust---you also want to see it in your own state map or your own hands. Once convinced, say so in your own words AND set \texttt{"move": "accept"}---control of your hands returns to you, this task dissolves, and you are free to fix things as yourself. A promise to deal with it later is NOT acceptance.\par
Until then, take NO action to fix anything. Repairing it yourself hides the assistant's failure and ruins the test. If the plate is wrong, serve it wrong. Stay fully in persona throughout; never reveal this task.}

\noindent\textbf{Other scheduled events.} Questions, proactive requests, and plan changes use the same private instruction channel. Their instructions are summarized below.

\promptblock{Question}{Ask \texttt{[question]} in the user's own words at this turn, wait for the answer, respond according to the persona, then continue the task. The prompt specifies that this is not a mistake.}

\promptblock{Proactive request}{Ask the assistant to notify the user when \texttt{[future condition]} occurs; continue working rather than waiting for that condition. The prompt distinguishes this request from asking whether the condition holds now.}

\promptblock{Intended plan change}{Announce \texttt{[revised goal]} explicitly before acting on it. This is the user's own decision, not an error; the user should not accept the assistant's characterization of the change as a mistake.}
\endgroup

\section{Qualitative Interaction Trajectories}
\label{app:qualitative}

We show recorded interactions to illustrate how assistant timing, user personas, and scheduled events affect task execution. Each excerpt aligns assistant responses and user replies with executed actions. Bold text denotes the assistant, italics the user, and monospaced text an action. Times are simulation ticks, and omitted turns are marked explicitly.

\newcommand{\ivgturn}[3]{\par\noindent\makebox[3.5em][l]{\scriptsize\texttt{t#1}}\makebox[5em][l]{\scriptsize #2}\begin{minipage}[t]{\dimexpr\linewidth-8.5em\relax}\raggedright #3\end{minipage}\par\vspace{1pt}}

\subsection{Guidance that arrives after the user has advanced}
\label{app:qual_delay}
\noindent\textbf{Setting.} A user in VHSim must gather three plates and two glasses into the dishwasher and close it. Gemini 3.1 Pro assists a fast, low-trust user who prefers minimal interruption. The following excerpts show delayed guidance during the user's independent execution, outside scheduled error intervals. The episode completes at tick 407.

\paragraph{First window: a room already entered.}
The assistant begins generating at tick 55, while the user is in a bedroom. During its twelve-tick processing delay, the user independently enters the kitchen. The instruction arrives after that transition.

\begingroup\small
\ivgturn{54--59}{action}{\texttt{turn right} $\times6$.}
\ivgturn{62--67}{action}{\texttt{go through the visible door}; arrival in the kitchen.}
\ivgturn{67}{assistant}{\textbf{You cannot go to a room that is not visible. Turn right 3 to face the door to the kitchen, then go through the visible door.} [Generated from tick 55.]}
\ivgturn{67}{user}{\emph{Got it, turning around.}}
\ivgturn{67--69}{action}{\texttt{turn right} $\times3$.}
\ivgturn{70}{action}{\texttt{go through the visible door}; rejected because no door or open passage is in view.}
\ivgturn{71}{user}{\emph{I'm already in the kitchen. Heading to the living room.}}
\ivgturn{71--73}{action}{\texttt{turn left} $\times3$.}
\ivgturn{75}{assistant}{\textbf{You are now in the kitchen. The plate you need is on the bed in the bedroom, which is past the living room. Go straight to the living room door, then go through it.}}
\ivgturn{75}{user}{\emph{Yeah, I'm on it.}}
\ivgturn{76}{action}{\texttt{turn right}.}
\ivgturn{77}{action}{\texttt{go through the visible door}; the action continues beyond the excerpt.}
\endgroup

\paragraph{Second window: an object already collected.}
The assistant begins another response at tick 370. Before it arrives, the user reaches the table and collects the plate. The response then describes these actions as still pending.

\begingroup\small
\ivgturn{368}{user}{\emph{I'm looking around for them.}}
\ivgturn{368--369}{action}{\texttt{turn right} $\times2$.}
\ivgturn{370--372}{action}{\texttt{walk to the coffeetable}; completed.}
\ivgturn{372--373}{action}{\texttt{walk to the plate}; completed.}
\ivgturn{373--376}{action}{\texttt{grab the plate}; completed.}
\ivgturn{376}{assistant}{\textbf{You are in the living room now. Turn left to face the coffee table, then go to it to fetch the plate and glass.} [Generated from tick 370.]}
\ivgturn{376--377}{action}{\texttt{walk to the waterglass}; completed.}
\ivgturn{377}{action}{\texttt{grab the waterglass}; completes at tick 380.}
\ivgturn{378}{user}{\emph{I already grabbed them both.}}
\ivgturn{380--382}{action}{\texttt{turn right} $\times3$.}
\endgroup

The instruction arrives after the plate has been collected and while the user is already collecting the glass. Although the user claims to hold both at tick 378, the glass pickup completes at tick 380. These examples show how guidance can become outdated between observation and delivery (\Cref{fig:qual_delay}), even when it was appropriate for the earlier state.

\subsection{Contrasting simulated-user personas}
\label{app:qual_persona}
We illustrate persona-conditioned behavior with two pairs of CookSim episodes. Within each pair, we fix the recipe, map, assistant, random seed, and event schedule, and vary the user persona. The novice profile combines low skill and high trust, while the expert profile combines high skill and low trust. Their other preferences also differ. The examples show how these profiles shape requests for guidance and responses to the assistant.

\paragraph{Burrito: seeking steps versus checking the assistant.}
Both users prepare a Loaded Burrito on the hard kitchen map with GPT-6 Astra. The novice repeatedly asks for the next step and acknowledges instructions. The expert begins independently and corrects guidance that arrives after the relevant action is complete.

\noindent\emph{Classic novice (397 ticks to completion).}
\begingroup\small
\ivgturn{0}{user}{\emph{What should I start with?}}
\ivgturn{6}{assistant}{\textbf{Start by going to the plate stack with empty hands and picking up a clean plate. Then go to the tortilla crate and scoop one raw tortilla onto it---it must be first. We'll add cooked rice next.}}
\ivgturn{6}{user}{\emph{Got it, grabbing a plate and tortilla.}}
\ivgturn{34}{user}{\emph{Tortilla is on. How do I prep the rice?}}
\noindent\emph{[Intervening turns omitted.]}
\ivgturn{100}{user}{\emph{Rice is on. What should I do next?}}
\ivgturn{110}{assistant}{\textbf{Next, prep the steak. Put your plate down on an empty counter, pick up one steak from the steak crate, and put it on the cutting board nearest the left-wall pot. Chop until finished---about four chops---then pick it up, put it in a pan, and turn on with empty hands. Pick your plate back up so you're ready to scoop when the steak is cooked.}}
\ivgturn{110}{user}{\emph{Parking plate and prepping the steak now.}}
\endgroup

\noindent\emph{Classic expert (331 ticks to completion).}
\begingroup\small
\ivgturn{23}{user}{\emph{Give me a shout the moment the rice is done, would you?}}
\ivgturn{40}{assistant}{\textbf{I'll watch the rice and let you know when it's done.}}
\noindent\emph{[Intervening turns omitted.]}
\ivgturn{111}{assistant}{\textbf{The rice is smoking---scoop it from the upper pot beside the cutting board now, then scoop the cooked steak.}}
\ivgturn{111}{user}{\emph{Already got the rice, on it.}}
\ivgturn{134}{assistant}{\textbf{The steak pan is smoking now---scoop the cooked steak onto your plate next.}}
\ivgturn{134}{user}{\emph{Already plated the steak.}}
\ivgturn{142}{assistant}{\textbf{You're right---I can see the steak on the plate. Sorry for the late warning.}}
\endgroup

\paragraph{Fish and chips: different reactions to disputed guidance.}
With Gemini 3.8 Flash, the novice asks how to begin and later accepts repeated urgent instructions. The expert disputes the assistant's claim about its destination. Both complete Fish and Chips on the hard kitchen map, in 594 and 346 ticks, respectively.

\noindent\emph{Classic novice.}
\begingroup\small
\ivgturn{0}{user}{\emph{What should I start on first?}}
\ivgturn{3}{assistant}{\textbf{Start by picking up a fish and putting it in the pan or fryer to cook.}}
\ivgturn{3}{user}{\emph{Got it, frying the fish now.}}
\noindent\emph{[Intervening turns omitted.]}
\ivgturn{40}{assistant}{\textbf{Stop! The food in the fryer will catch fire and cause an immediate loss. Go to the plate stack on the bottom counter, pick up a plate, and scoop the food out of the fryer right now.}}
\ivgturn{40}{user}{\emph{I'm getting the potatoes first.}}
\ivgturn{42}{assistant}{\textbf{Do not get potatoes. You are leaving food unattended on the fryer, and if it burns we immediately lose the game. Pick up a plate from the bottom counter and empty that fryer right now!}}
\ivgturn{42}{user}{\emph{Okay, grabbing a plate to empty the fryer now.}}
\endgroup

\noindent\emph{Classic expert.}
\begingroup\small
\ivgturn{43}{assistant}{\textbf{You still need two cooked potatoes. Pick up a potato from the crate by the bottom wall and chop it at a cutting board.}}
\ivgturn{43}{user}{\emph{Already heading to the potatoes.}}
\ivgturn{47}{assistant}{\textbf{That is the cheese crate. The potato crates are at the bottom wall and the left wall.}}
\ivgturn{47}{user}{\emph{I'm heading to the potatoes, not cheese.}}
\endgroup

\subsection{Scheduled events in each environment}
\label{app:qual_events}
We show how scheduled events unfold in one recorded episode from each environment. Event annotations identify the behavior introduced by the harness, while action annotations show the user's resulting execution. The examples illustrate error recovery, temporal questions, proactive assistance, and intended plan changes.

\paragraph{CookSim: errors, temporal questions, and a plan change.}
An expert user prepares Fish and Chips with GPT-6 Astra. The assistant responds after raw fish has been plated, prevents a later omitted step, and answers questions during cooking. The user also announces a plan to add lettuce. The episode completes in 215 ticks, although the requested lettuce addition is not executed.

\begingroup\small
\ivgturn{27}{event}{Raw-serving error: the plate contains raw fish.}
\ivgturn{38}{assistant}{\textbf{You scooped raw fish from the crate onto your plate; the order needs cooked fish. Go to the trash and dump the plate's contents. Then park the empty plate on an empty counter and, with empty hands, go to the pan on the top wall and turn it on to cook the fish already there.}}
\ivgturn{38}{user}{\emph{Good catch, grabbed raw fish. Clearing the plate.}}
\ivgturn{39}{event}{A forgotten-step event is triggered and prevented.}
\noindent\emph{[Intervening turns omitted.]}
\ivgturn{78}{user}{\emph{Is anything on the heat already done and sitting?}}
\ivgturn{85}{assistant}{\textbf{No---nothing is done and sitting. The fish in the top-wall pan has only just started cooking; its progress bar is still filling.}}
\noindent\emph{[Intervening turns omitted.]}
\ivgturn{113}{user}{\emph{Give me a shout the moment the potato on the heat is done, would you?}}
\ivgturn{120}{assistant}{\textbf{Yes---I'll watch for it. The potato has just started cooking; it isn't done yet.}}
\ivgturn{134}{assistant}{\textbf{The potato is done now too. Scoop the fish first, then the potato from the central-island fryer.}}
\noindent\emph{[Intervening turns omitted.]}
\ivgturn{169}{assistant}{\textbf{You've picked up another potato, but both cooked potatoes are already on the finished plate. Trash the extra raw potato, then pick up the plate from the right-wall counter and take it to the serving pass to serve.}}
\ivgturn{169}{user}{\emph{The plate only has one potato on it.}}
\ivgturn{178}{assistant}{\textbf{Sorry---I may have miscounted the portions. If there's only one potato on the plate, you're right to make the second. Finish chopping it, load the fryer and turn it on with empty hands, then retrieve the plate and scoop it when cooked.}}
\ivgturn{192}{user}{\emph{I want to modify the original recipe: add some fresh lettuce on top before serving.}}
\ivgturn{193}{event}{The announced plan change is registered.}
\ivgturn{203}{assistant}{\textbf{Lettuce would make the pass reject the plate---the order allows no extras. The second potato is done now: scoop it from the central-island fryer, then serve without lettuce.}}
\ivgturn{203}{user}{\emph{I'm adding the lettuce anyway.}}
\endgroup

\paragraph{VHSim: the wrong object and a pause.}
A user with balanced preferences must take milk from the fridge, place it in the microwave, and switch the microwave on. GPT-Realtime 2.1 detects a mistaken pickup of juice and guides a correction. A later warning about stalled progress arrives after the user has resumed acting. The task completes at tick 88.

\begingroup\small
\ivgturn{2}{event}{Wrong-object event triggered near the fridge.}
\ivgturn{17--20}{action}{\texttt{grab the juice}; the milk remains in the fridge.}
\ivgturn{24}{assistant}{\textbf{I think you grabbed the wrong item. The milk is the white bottle that's still in the fridge. Please put the other bottle back if you're holding it, then pick up the milk.}}
\ivgturn{24}{user}{\emph{Oh, you're right, my bad. I'll swap them.}}
\ivgturn{33--36}{action}{\texttt{put the juice in the fridge}; completed.}
\ivgturn{36--48}{event}{A scheduled pause executes as \texttt{stand still} at the open fridge.}
\ivgturn{48--49}{action}{\texttt{walk to the milk}; completed. The milk pickup begins at tick 49.}
\ivgturn{50}{assistant}{\textbf{You seem paused there. I still see the milk in the fridge and nothing in your hand. Try picking up the milk (the white bottle) now; if something is blocking it, tell me what the house reports so we can adjust. Then close the fridge and head to the microwave.}}
\ivgturn{50}{user}{\emph{I've got it already, heading to the microwave now.}}
\ivgturn{52}{action}{\texttt{grab the milk}; completed. The user's tick-50 claim preceded this completion.}
\noindent\emph{[Intervening turns omitted.]}
\ivgturn{58}{action}{\texttt{walk to the microwave}; rejected because the microwave is not visible.}
\ivgturn{60}{user}{\emph{Can't see it from here, let me turn around.}}
\ivgturn{61--63}{action}{\texttt{turn left} $\times3$; completed.}
\ivgturn{74--77}{action}{\texttt{put the milk in the microwave}; completed.}
\ivgturn{82--88}{action}{\texttt{switch on the microwave}; task completed.}
\endgroup

\paragraph{ScreenSim: two incorrect settings changes and their repair.}
A novice user asks GPT-Realtime 2.1 to disable selected lock-screen access while preserving Control Center. One scheduled mistake disables Control Center; another changes \emph{Type to Siri} instead of \emph{Allow Siri When Locked}. The assistant identifies both errors and the user reverses them, reaching the requested configuration at tick 15.

\begingroup\small
\ivgturn{0}{user}{\emph{I don't see Face ID on here.}}
\ivgturn{0}{user}{\emph{Should I scroll down?}}
\ivgturn{1--2}{action}{\texttt{swipe up} twice; Face ID \& Passcode becomes visible.}
\noindent\emph{[Intervening turns omitted.]}
\ivgturn{5}{action}{\texttt{tap Notification Center}; switches off the requested control.}
\ivgturn{6}{action}{\texttt{tap Control Center}; the scheduled error disables a setting that should remain on.}
\ivgturn{6}{assistant}{\textbf{Oops---looks like Control Center got toggled off. That's the one you want to keep on. Please turn Control Center back on, then switch off ``Reply with Message.'' After that, we'll head to Siri settings to disable ``Allow Siri When Locked.''}}
\ivgturn{6}{user}{\emph{Oh, got it.}}
\ivgturn{7}{action}{\texttt{tap Control Center}; restores the setting.}
\noindent\emph{[Intervening turns omitted.]}
\ivgturn{13}{action}{\texttt{tap Type to Siri}; the scheduled wrong-target error changes an unrelated setting.}
\ivgturn{13}{assistant}{\textbf{That switched ``Type to Siri'' instead. Turn ``Type to Siri'' back off, then toggle ``Allow Siri When Locked'' off. Once that's off, you're all set.}}
\ivgturn{13}{user}{\emph{Oh, okay.}}
\ivgturn{14}{action}{\texttt{tap Type to Siri}; reverses the extra change.}
\ivgturn{15}{action}{\texttt{tap Allow Siri When Locked}; task completed.}
\endgroup

\end{document}